\documentclass[9.9pt,logo,copyright]{nvidiatechreport}
\usepackage[authoryear,sort&compress,round]{natbib}

\usepackage[utf8]{inputenc} 
\usepackage[T1]{fontenc}    

\usepackage{parskip}        
\usepackage{url}            
\usepackage{booktabs}       
\usepackage{amsfonts}       
\usepackage{nicefrac}       
\usepackage{microtype}      
\usepackage{xcolor}         
\usepackage{graphicx}
\usepackage{animate}        
\usepackage{subcaption}
\usepackage{tabularx}
\usepackage{makecell}
\usepackage{adjustbox}
\usepackage{setspace}
\newcolumntype{M}[1]{>{\centering\arraybackslash}m{#1}}
\usepackage{float}
\usepackage{tikz}

\usepackage{amsmath,amsfonts,bm, bbm,leftindex}

\newcommand{\diff}{\mathrm{d}}

\def\eqref#1{equation~\ref{#1}}

\def\1{\bm{1}}

\def\rvepsilon{{\mathbf{\epsilon}}}

\def\rvx{{\mathbf{x}}}

\def\mI{{\bm{I}}}

\DeclareMathAlphabet{\mathsfit}{\encodingdefault}{\sfdefault}{m}{sl}
\SetMathAlphabet{\mathsfit}{bold}{\encodingdefault}{\sfdefault}{bx}{n}

\def\gN{{\mathcal{N}}}

\def\gX{{\mathcal{X}}}

\newcommand{\E}{\mathbb{E}}

\newcommand\norm[1]{\left\lVert#1\right\rVert}

\definecolor{darkred}{rgb}{0.7, 0.0, 0.0}

\usepackage[nameinlink]{cleveref}
\crefname{equation}{Eq.}{Eqs.}
\crefname{figure}{Fig.}{Figs.}
\crefname{section}{Sec.}{Sec.}
\crefname{appendix}{App.}{App.}
\crefname{table}{Tab.}{Tabs.}
\crefname{algorithm}{Algo}{Algo}
\crefname{thm}{Thm}{Thm}
\Crefname{thm}{Thm}{Thm}
\crefname{prop}{Prop}{Prop}

\newcommand{\crefnames}[3]{%
  \@for\next:=#1\do{%
    \expandafter\crefname\expandafter{\next}{#2}{#3}%
  }%
}

\title{Edify Image: High-Quality Image Generation with Pixel Space Laplacian Diffusion Models}

\author{NVIDIA\footnote{A detailed list of contributors and acknowledgments can be found in~\cref{sec:contributors} of this paper.}}

\begin{abstract}
We introduce Edify Image, a family of diffusion models capable of generating photorealistic image content with pixel-perfect accuracy. Edify Image utilizes cascaded pixel-space diffusion models trained using a novel Laplacian diffusion process, in which image signals at different frequency bands are attenuated at varying rates. Edify Image supports a wide range of applications, including text-to-image synthesis, $4K$ upsampling, ControlNets, $360^{\circ}$ HDR panorama generation, and finetuning for image customization. 

\end{abstract}

\begin{document}

\maketitle

\vspace{-15pt}
\begin{figure}[H]
    \label{fig:teaser}
    \centering
    \begin{subfigure}{0.62\linewidth}
        \centering        
        \includegraphics[width=\textwidth]{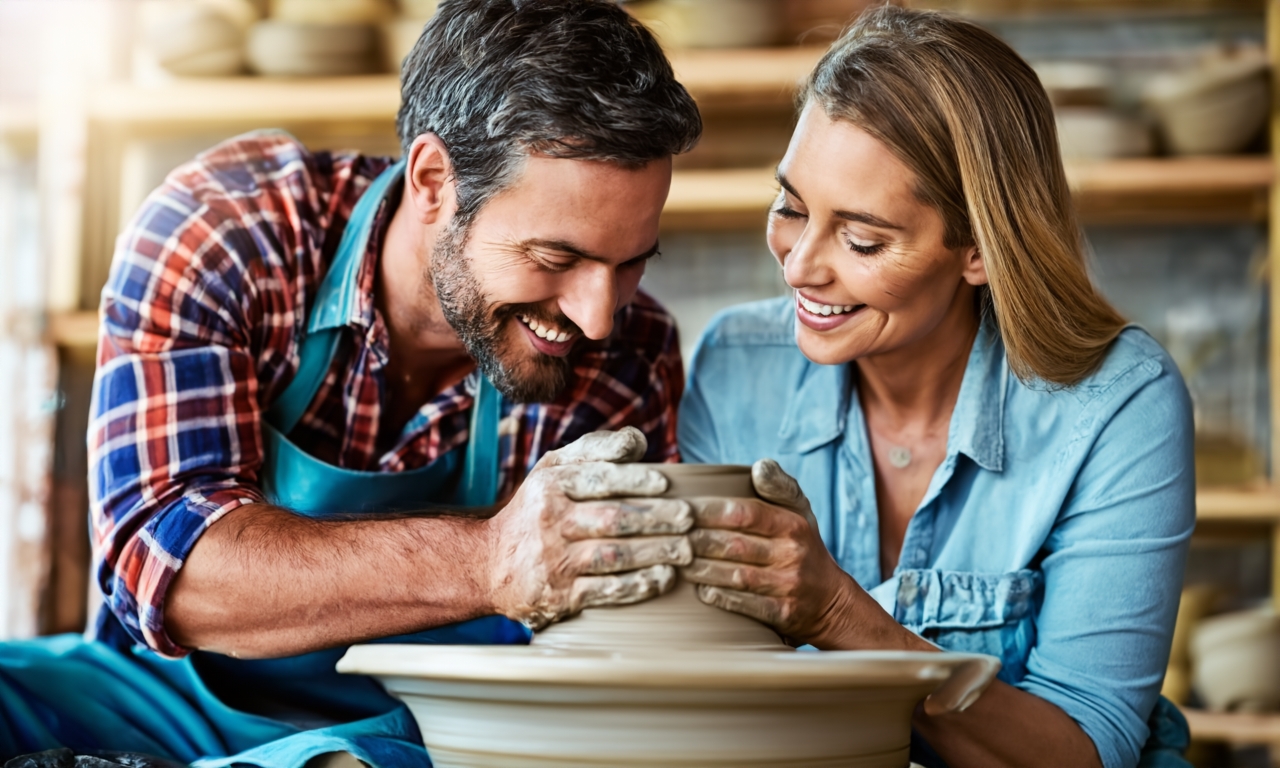}
        \emph{A photo of a couple doing pottery together in a well-lit room}
    \end{subfigure}
    \begin{subfigure}{0.372\linewidth}
        \centering
        \includegraphics[width=\textwidth]
        {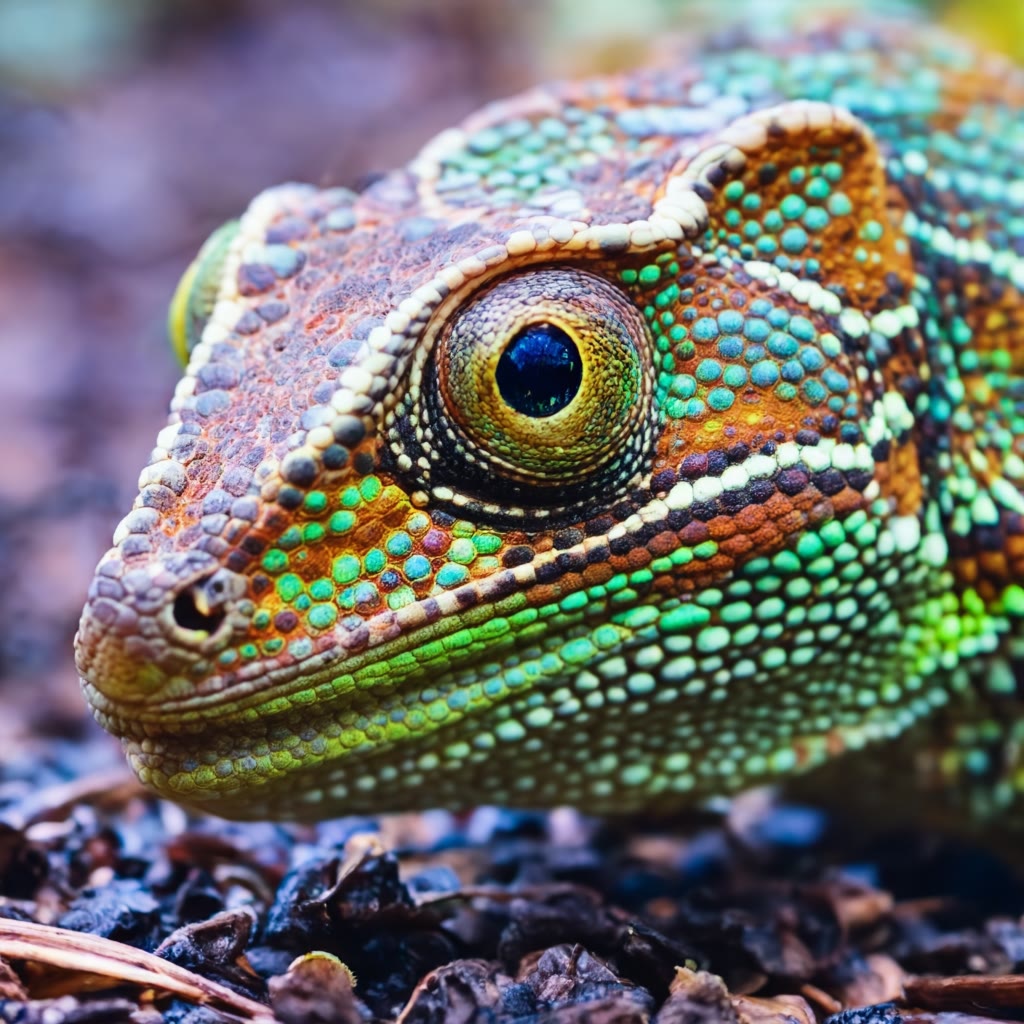}
        \emph{A chameleon showing colorful scales}
    \end{subfigure}
    \\
    \begin{subfigure}{0.99\linewidth}
    \caption{Text-to-image generation}
    \end{subfigure}
    \vspace{3mm}
    \\    
    \begin{subfigure}{0.3\linewidth}
        \centering
        \includegraphics[width=\textwidth]{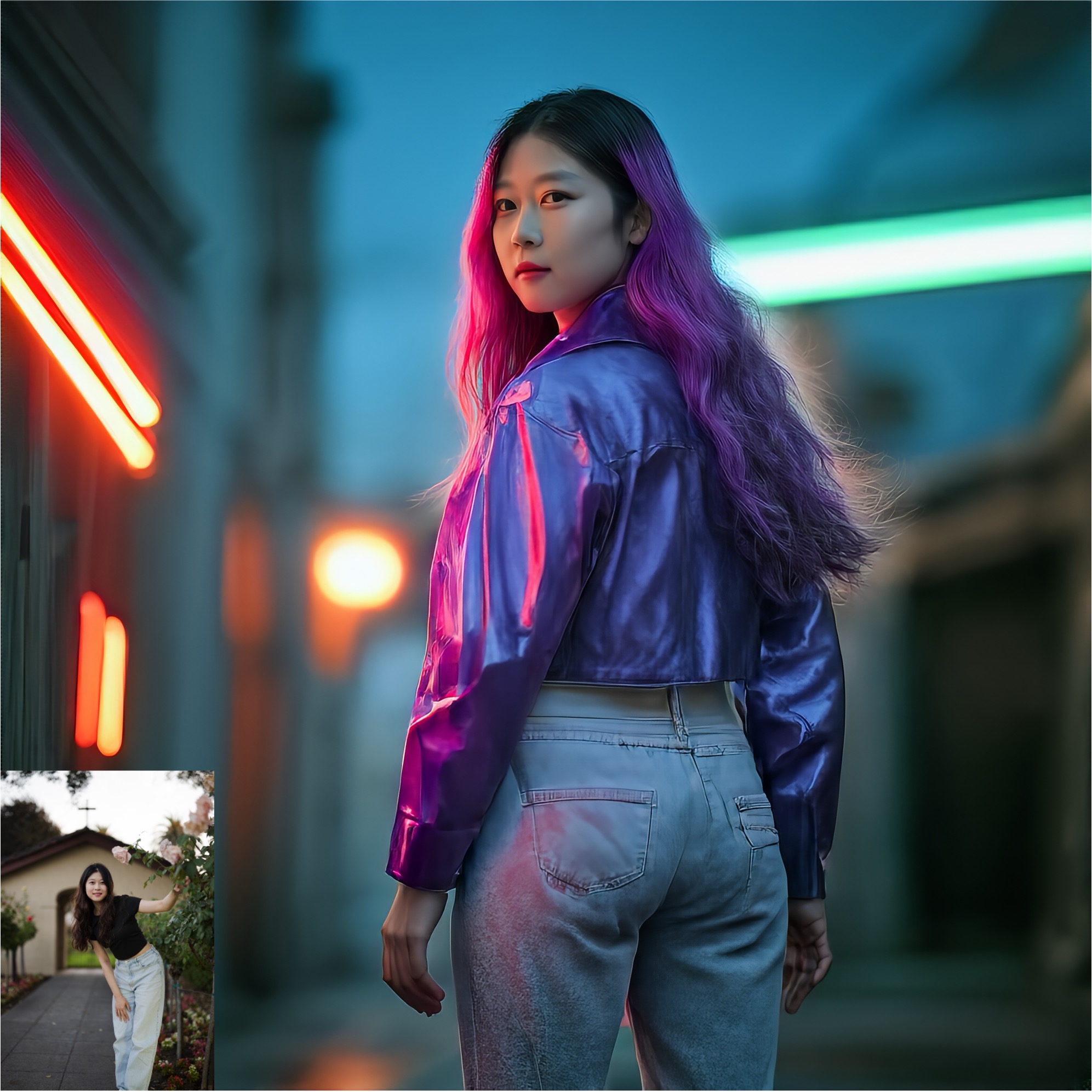}
        \caption{Finetuning}
    \end{subfigure}
    \begin{subfigure}{0.3\linewidth}
        \centering
        \includegraphics[width=\textwidth]{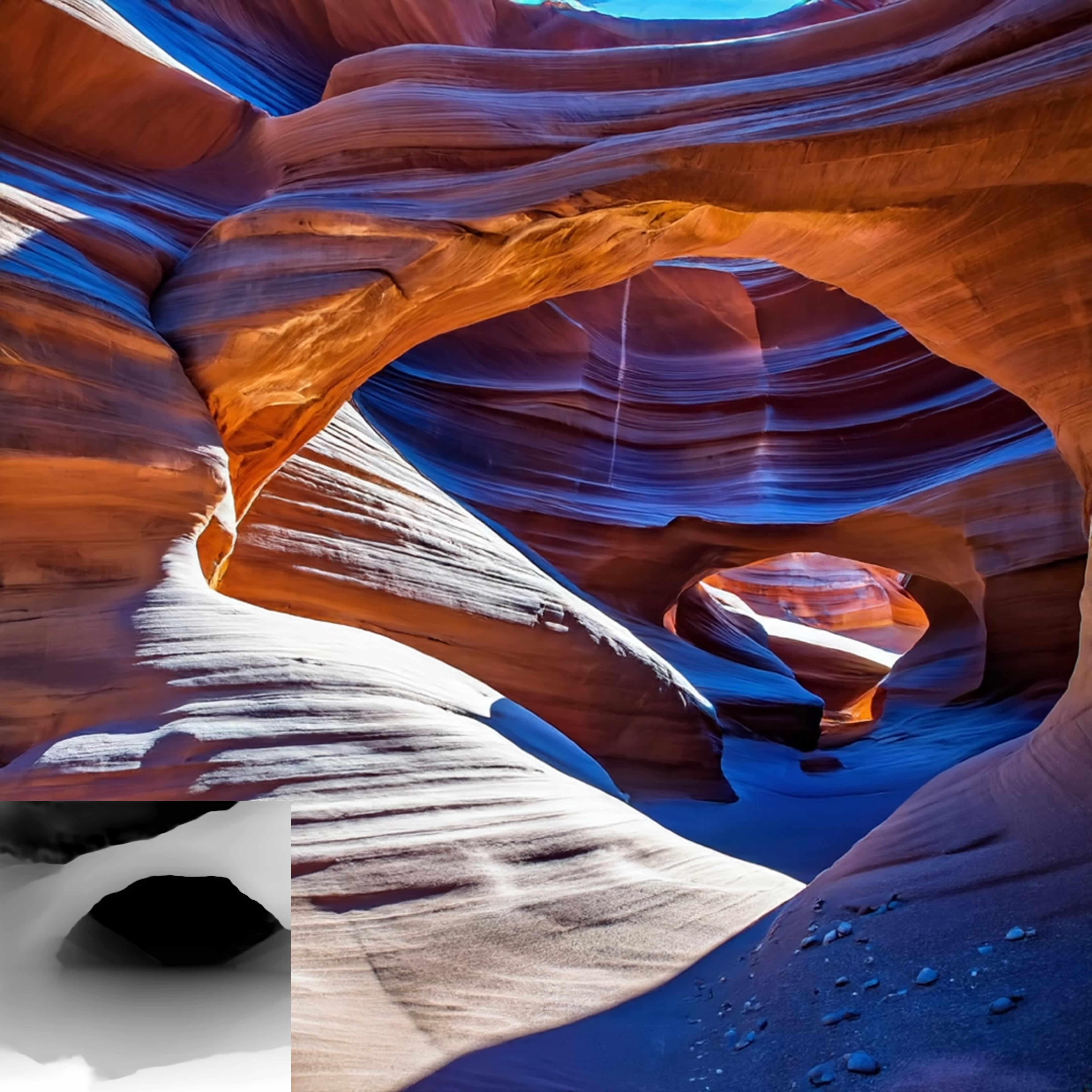}
        \caption{Additional control}
    \end{subfigure}
    \begin{subfigure}{0.39\linewidth}
        \centering
            \animategraphics[loop,autoplay, trim = 25mm 1.7mm 25mm 1.7mm, clip, width=\textwidth, poster=28, interpolate]{15}{images/360/360video/360video-15-}{051}{150}
        \vfill
        \caption{Panorama}
    \end{subfigure}
    \caption{\textbf{Edify Image can generate photorealistic high-resolution images from text prompts.} Our models support a range of capabilities, including (a) Text-to-image generation, (b) Finetuning, (c) Generation with additional control, and (d) Panorama generation. For (b) and (c), an example of a finetuning image and the control input are provided in the bottom left corner, respectively. Best viewed with Acrobat Reader. Click the panorama image to play the video clip.}
\end{figure}

\abscontent

\begin{center}\noindent\href{https://research.nvidia.com/labs/dir/edify-image}{\textbf{https://research.nvidia.com/labs/dir/edify-image}}
\end{center}

\section{Introduction}
\label{sec:intro}

The field of text-to-image synthesis has witnessed remarkable progress in recent years, with state-of-the-art models~\citep{betker2023improving, esser2024scaling, baldridge2024imagen, podell2023sdxl} generating increasingly realistic and diverse images from natural language descriptions. These models typically leverage large-scale diffusion-based architectures trained on billions of image-text pairs. The ability to generate high-resolution, photorealistic images has far-reaching implications across domains such as content creation, gaming, synthetic data generation, and the design of digital avatars. 

In this technical report, we present Edify Image, a family of pixel-space diffusion models capable of generating high-resolution images with exceptional controllability. We train our models in a cascaded fashion, where a base model generates low-resolution images, and subsequent models progressively upsample the images from the previous stage. Our models are trained using a novel multi-scale Laplacian diffusion process, in which image signals at different frequency bands are attenuated at varying rates. This enables our models to capture and refine details with precision across multiple scales, resulting in photorealistic, pixel-perfect generations.

Using the Laplacian Diffusion Model formulation, we train a suite of diffusion models capable of generating images from various input signals.
\begin{itemize}
    \item \textbf{Text-to-image models.} We train a two-stage cascaded text-to-image diffusion model that can generate $1K$ resolution images from input text. Our model handles long text prompts, generates images with varying aspect ratios, exhibits improved fairness and diversity when generating human subjects, and can support the use of camera controls such as pitch and depth of field. 
    \item \textbf{$4K$ Upsampler.} We train an upsampler model that takes a $1K$ resolution image as input and upsamples it to $4K$ resolution. The upsampler involves fine-tuning the $1K$ resolution generator on $4K$ images with an additional low-resolution input conditioning. Our model is capable of synthesizing high-frequency details while remaining faithful to the low-resolution input image.
    \item \textbf{ControlNets.} We train ControlNets on the $256$ resolution base models for various modalities, including depth, sketch, and inpainting mask. The $1K$ and $4K$ base models can be reused for upsampling. Our model can generate high-quality images while enabling flexible structural controls. 
\end{itemize}

In addition, we also support the following two capabilities:
\begin{itemize}
    \item \textbf{$360^{\circ}$ HDR Panorama  Generation.} We design an algorithm for generating $4K$, $8K$ and $16K$ resolution HDR panorama from the input text. We utilize the base text-to-image models to perform sequential inpainting in which images from different perspectives are generated in an overlapping manner and stitched together with consistency.
    \item \textbf{Finetuning.} We propose an algorithm for finetuning the base text-to-image models on a small subset of reference images. Our model is especially capable of generating various hyper-realistic humans with identities consistent with the reference set.
\end{itemize}

\section{Dimension-Varying EDM}
\label{sec:dim_varying_edm}

Edify Image models are diffusion-based generators operating in the pixel space. Existing pixel-space generators employ a series of cascaded diffusion models in which subsequent stages upsample the low-resolution images produced in the previous stage, often leading to notorious artifact accumulation. To mitigate this issue, we introduce a new diffusion model that synthesizes large contexts in a single diffusion process.
The key innovation is the introduction of a multi-scale diffusion process, termed the Laplacian Diffusion Model. This model simulates a resolution-varying diffusion process in the time domain by simultaneously decaying different image frequency bands at different rates.

\subsection{Preliminary}

\subsubsection{Diffusion Model}
Given an image data distribution $p_0(\rvx_0)$, where $\rvx_0 \in \gX$, a diffusion model derives a family of distributions $p_t(\rvx_t)$ by injecting~\iid Gaussian noise into data samples during the diffusion forward process, such that $\rvx_t = \rvx_0 + \sigma_t \rvepsilon$ with $\rvepsilon \sim \gN(0, \mI)$ and $\sigma_t$ monotonically increasing with respect to time $t \in [0, T]$. 
To simulate the diffusion backward sampling process, which generates samples by iteratively removing noise starting from Gaussian noise, diffusion models obtain the score function $\nabla_{\rvx_t} \log p_t(\rvx_t)$ (\ie, the gradient of log-probability) via a denoising score matching objective~\citep{vincent2011a,Karras2022edm,song2021scorebasedgenerativemodelingstochastic,ho2020denoisingdiffusionprobabilisticmodels}:
\begin{equation}\label{eq:loss}
    L_t(\theta) = \E_{\rvx_0, \rvx_t} [\norm{D_\theta(\rvx_t, t) - \rvx_0}_2^2],
\end{equation}
where $D_\theta: \gX \times [0, T] \to \gX$ is a time-conditioned neural network that tries to denoise the noisy sample $\rvx_t$. Assuming an infinite capacity of $D_\theta$, the predictions of the optimal model are related to the score function via Tweedie's formula~\citep{efron2011tweedie}:
\begin{align}
    \hat{\rvx}_t := & \ D_\theta(\rvx_t, t) 
     = \rvx_t + \sigma_t^2 \nabla_{\rvx_t} \log p_t(\rvx_t), \label{eq:x0-pred}
\end{align} 
which represents the minimum mean squared error (MMSE) estimator of $\rvx_0$ given $\rvx_t$ and $\sigma_t$.
We follow the precondition design for $D_\theta(\rvx_t, t)$ and log normal distribution $\sigma$ during training introduced in~\citet{Karras2022edm}.

\begin{figure}
    \centering
    \adjincludegraphics[width=0.95\linewidth]{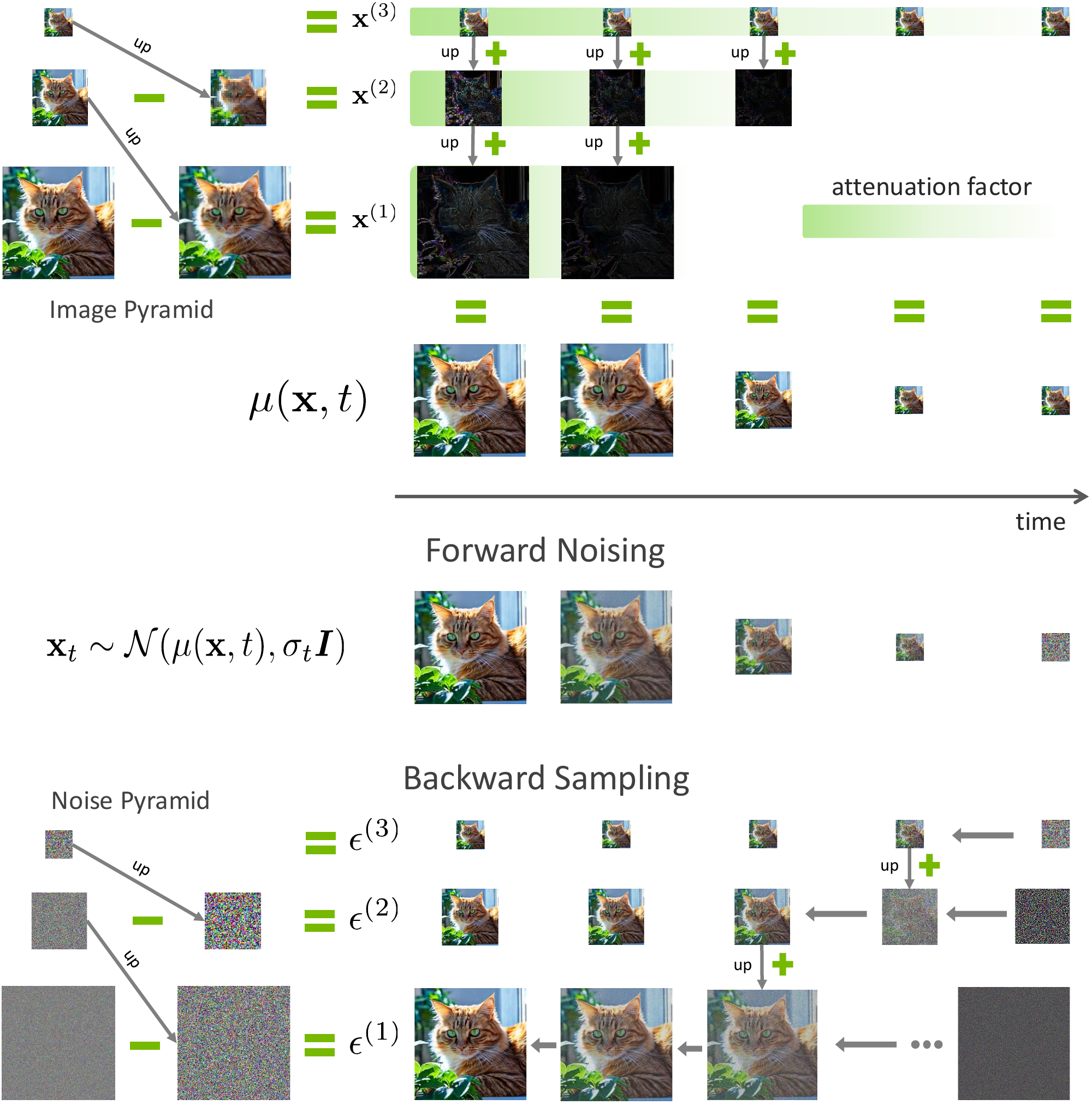}
    \captionof{figure}{
\textbf{Laplacian diffusion for multi-resolution image generation.} 
(Top) \textbf{Image Laplacian Decomposition}. Each image sample $\rvx$ can be decomposed into a set of components. The example shows three components, $\rvx = \rvx^{(1)} + \text{up}(\rvx^{(2)}) + \text{up}(\text{up}(\rvx^{(3)}))$. This decomposition is implemented using basic upsampling and downsampling operations, where each component corresponds to different frequency bands. The function $\mu(\rvx, t)$ represents a weighted sum of these components across different frequency spaces.
(Middle) \textbf{Forward Noising Process.} Components are attenuated at different rates, with higher frequencies attenuated more rapidly than lower ones. We use the decaying background color in the top part of the figure to illustrate the attenuation factors. As a result, the signal-to-noise ratio (SNR) diminishes faster in the high-frequency components, allowing them to be discarded without significant loss of information once their attenuation coefficients approach zero.
(Bottom) \textbf{Backward Sampling Process.} Denoisers are trained at multiple stages to generate images at various resolutions. We decompose the noise into a noise Laplacian pyramid. The Laplacian Diffusion process synthesizes higher-resolution images by first upsampling a lower-resolution noisy sample and then denoising it, with random noise injected into the corresponding components during upsampling. When operating solely at the lowest resolution, the process reduces to standard EDM.}
\label{fig:formulation}
\end{figure}

\subsubsection{Image Laplacian Decomposition}

The Image Laplacian Decomposition is a multi-scale representation technique that decomposes an image into a series of progressively lower-resolution images, capturing different frequency bands at each level. This hierarchical structure consists of a sequence of band-pass filtered images, where each level represents the difference between two successive versions of the original image. 
Specifically, we consider a simple image downsampling operation as a way to obtain the low-frequency component, where high-frequency details from the original image are effectively removed. 
We denote upsampling and downsampling operations as $\text{up}(.)$ and $\text{down}(.)$, respectively.
We illustrate this decomposition in~\cref{fig:formulation}.
Through this decomposition, for simplicity, we assume there are three resolution stages, i.e. $\rvx = \rvx^{(1)} + \text{up}(\rvx^{(2)}) + \text{up}(\text{up}(\rvx^{(3)}))$, where:
\begin{subequations}
\label{eq:laplacian-decomp}
\begin{align}
\label{eq:decompose1}
\rvx^{(3)} &= \text{down}(\text{down}(\rvx)), \\
\rvx^{(2)} &= \text{down}(\rvx) - \text{up}(\rvx^{(3)}), \\
\rvx^{(1)} &= \rvx - \text{up}(\text{down}(\rvx)).
\end{align}
\end{subequations}
Note that even though we use a $d$ dimensional vector to present $\rvx^{(i)}$, their internal representation can be more compact. For example, we can use a downsampled $d/16$ dimensional vector to represent $\rvx^{(3)}$ to tackle high-resolution image synthesis.

\subsection{Laplacian Diffusion Model}
We introduce our diffusion process, which is built upon image Laplacian decomposition using an intuitive approach. 
At its core, we explicitly control how image signals at different frequency bands are attenuated and synthesized at varying rates rather than entangling them together and allowing them to be corrupted through an implicit approach.
A rigorous treatment can be derived with stochastic differential equations.
We start with a 3-stage image Laplacian decomposition in~\cref{eq:laplacian-decomp}. The formulation can be extended to more stages easily.

\subsubsection{Forward Noising Process} 
We generalize the isotropic forward process utilized in standard diffusion models, where $\rvx_t \sim \gN(\rvx_0, \sigma_t \mI)$, to a more flexible formulation: $\rvx_t \sim \gN(\mu(\rvx_0, t), \sigma_t \mI)$. In this context, $\mu$ is defined as:
\begin{equation}\label{eq:mean_mu}
\mu(\rvx_0, t) = \sum_{i=1}^3 \alpha^{(i)}_t \rvx_0^{(i)},
\end{equation}
where the coefficients $\alpha^{(i)}_t$ are attenuation factors. We define attenuation factors to be monotonically non-increasing with respect to the diffusion time $t$. 
This forward process can be expressed as the summation of three diffusion models operating in different subspaces:
\begin{equation}\label{eq:forward-diffusion}
\rvx_t = \sum_{i=1}^3 \alpha^{(i)}_t \rvx_0^{(i)} + \sigma_t \epsilon
= \sum_{i=1}^3 \alpha^{(i)}_t  \rvx_0^{(i)} + \sigma_t \epsilon^{(i)},
\end{equation}
where $\epsilon^{(i)}$ can be obtained via the Laplacian decomposition as in~\cref{eq:laplacian-decomp}. We also visualize this process at the bottom of~\cref{fig:formulation}. Most existing diffusion models choose $\alpha^{(i)}_t$ that are invariant to subspace, thereby entangling the three components at any given time $t$. 
Consequently, the denoising network is required to operate across all three subspaces to reconstruct the original signals for all diffusion processes.

In our study, we employ distinct rates for the set ${ \alpha_t^{(i)} }$, such that the components in the high-frequency branch decay more rapidly than those in the lower-frequency branch, as illustrated in~\cref{fig:formulation}. 
We identify two critical time points, $t^{(1*)}$ and $t^{(2*)}$, at which $\alpha_t^{(1)}$ and $\alpha_t^{(2)}$ respectively diminish to zero. Consequently, beyond these timestamps, a more compact, low-resolution representation suffices for the signal, as the high-frequency components no longer contribute to $\rvx_t$.

\subsubsection{Training} 
We utilize the same loss function, as defined in~\cref{eq:loss}, to train the denoising network $D_\theta(\rvx_t, t)$. However, the Laplacian forward process introduces greater flexibility in network design, allowing us to operate across different resolution ranges. Moreover, this approach greatly improves the training efficiency by separating the low-frequency and high-frequency components of the image, allowing the model to adapt more quickly. As illustrated in~\cref{fig:formulation}, we can train a large network for the whole time interval: $[0, \infty)$. 
Alternatively, we can employ a mixture of experts approach, where a low-resolution denoiser $D_\theta^{(3)}$ is trained on $\gX^{(3)}$ for the entire time range $[0, \infty)$, a mid-resolution denoiser $D_\theta^{(2)}$ is trained on $\gX^{(2)} \cup \gX^{(3)}$ for the interval $[0, t^{(2*)})$, and a high-resolution denoiser $D_\theta^{(1)}$ is trained on $\gX$ for the interval $[0, t^{(1*)})$. 

\subsubsection{Backward Sampling Process}
Laplacian Diffusion Models offer a flexible approach to synthesizing samples at various resolutions, thanks to the Laplacian decomposition and the utilization of a mixture of denoiser experts trained across different denoising ranges. We illustrate the different sampling modes in~\cref{fig:formulation}. 

\begin{itemize}
    \item To synthesize the lowest resolution images in $\gX^{(3)}$, the backward sampling process simplifies to that of regular diffusion models, as it involves only a single stage based on $D_\theta^{(3)}$.
    \item For generating mid-resolution images, we can combine the outputs of the denoisers $D_\theta^{(3)}$ and $D_\theta^{(2)}$. Specifically, we perform backward sampling in $\gX^{(3)}$ up to $t^{(2)*}$, then transition to using $D_\theta^{(2)}$ to complete the remaining sampling trajectory.
    \item To synthesize the highest resolution images, we switch the sampling trajectory from $D_\theta^{(2)}$ at the sampling timestamp $t^{(1)*}$, and rely on $D_\theta^{(1)}$ to generate the remaining high-resolution details.
\end{itemize}
We include more discussions and derivations in~\cref{app:formulation}. We extend high-order sampling algorithms~\citep{zhang2023improved,zhang2022fast} from standard diffusion models to the Laplacian Diffusion Model following the similar spirit introduced by~\citet{zhang2022gddim}.

\subsubsection{Switching Between Different Resolutions} 
When synthesizing low-resolution images, we completely disregard the signals from the high-frequency band to reduce computational costs. This approach is justified by the fact that the signal-to-noise ratio is zero during the corresponding time interval. However, to synthesize high-resolution images, it is necessary to switch the sampling trajectory by upsampling the noisy image $\rvx_t$ and reintroducing the high-frequency noise components.
We illustrate this concept using a low-resolution image ($r$) and assume that we are at a noise level $\sigma$ (under resolution $r$). Transitioning to a high-resolution ($R$) image with a noise level $R/r \cdot \sigma$ involves two steps: first, upscale the low-resolution image to high resolution, and second, add the corresponding high-resolution Gaussian noise component, multiplied by $(\sigma \cdot R / r)$.

We justify the approach using a concrete example. Consider that a noisy state $\rvx_t$ at resolution $(r)$ can be decomposed as:
\begin{align}
    \rvx^{(r)} + \sigma \epsilon^{(r)},
\end{align}
where $\epsilon^{(r)}$ is the resolution-$r$ standard Gaussian noise. Let us define $\epsilon^{(R)}$ to be the standard Gaussian noise of resolution $R$, such that:
\begin{align}
    \epsilon^{(r)} = \text{down}(\epsilon^{(R)}, R/r) \cdot R / r, \label{eq:downsample-variance}
\end{align}
where the coefficient is due to the averaging of Gaussian noise. Thus, we have that:
\begin{align}
   & \underbrace{\text{up}(\rvx^{(r)} + \sigma \epsilon^{(r)})}_{\text{upscale}} + \underbrace{\sigma R/r \cdot (\epsilon^{R} - \text{up}(\text{down}(\epsilon^{(R)}, R/r))}_{\text{add noise}} \\
   = \ & \text{up}(\rvx^{(r)}) + \sigma R/r \cdot \epsilon^{R} + \sigma \cdot \text{up}(\epsilon^{(r)} - \text{down}(\epsilon^{(R)}, R/r) \cdot R/r) \\
   = \ & \text{up}(\rvx^{(r)}) + \sigma R/r \cdot \epsilon^{R},
\end{align}
where the last equality is from \cref{eq:downsample-variance}. Here, we have translated the low-resolution Gaussian noise to high-resolution Gaussian noise. 

\section{$1K$ Generation Using Two-Stage Laplacian Diffusion Models}\label{sec:1K_generation}

\begin{figure*}[t]
    \centering
    \includegraphics[width=0.95\textwidth]{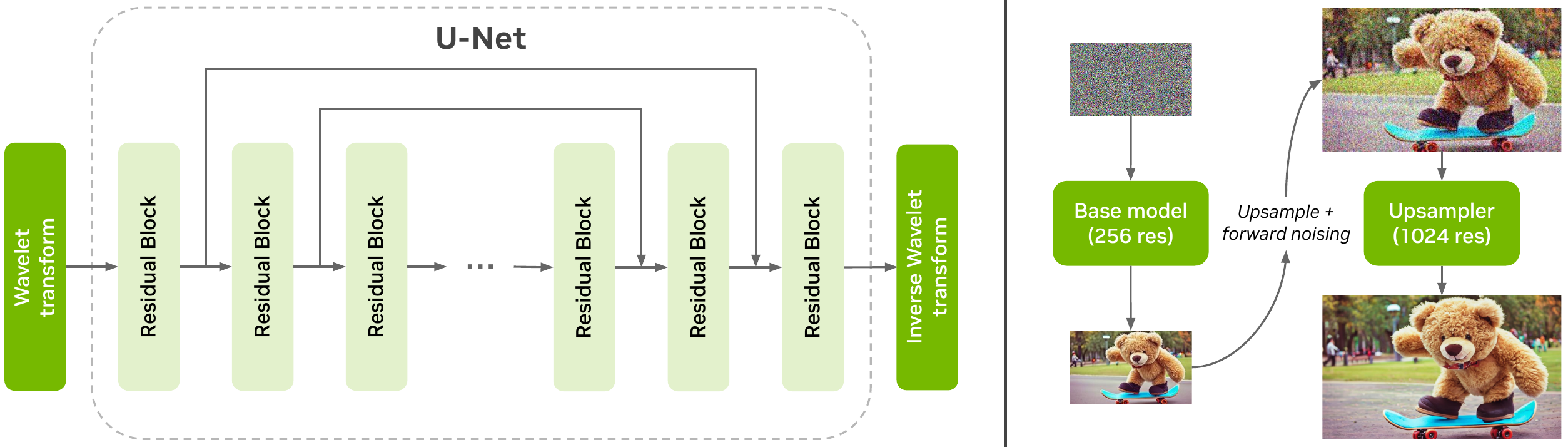}
    \caption{\textbf{Model architecture.} As shown in the left panel, our diffusion models use a U-Net based architecture with a sequence of residual blocks with skip connections. We use wavelet and Inverse wavelet transform at the beginning and end of the network to bring down the spatial resolution of the images. In the right panel, we show how the $256$ and $1K$-resolution models are combined in a 2-stage cascade to generate the $1024$-resolution image.}
    \label{fig:model_arch}
\end{figure*}
\setlength{\tabcolsep}{0.5pt}
\renewcommand{\arraystretch}{0.5}
\begin{table*}[ht!]
\centering
\begin{tabularx}{\textwidth}{M{0.25\textwidth} M{0.25\textwidth} M{0.25\textwidth} M{0.25\textwidth}}
    \multicolumn{2}{c}{\includegraphics[width=0.497\textwidth]{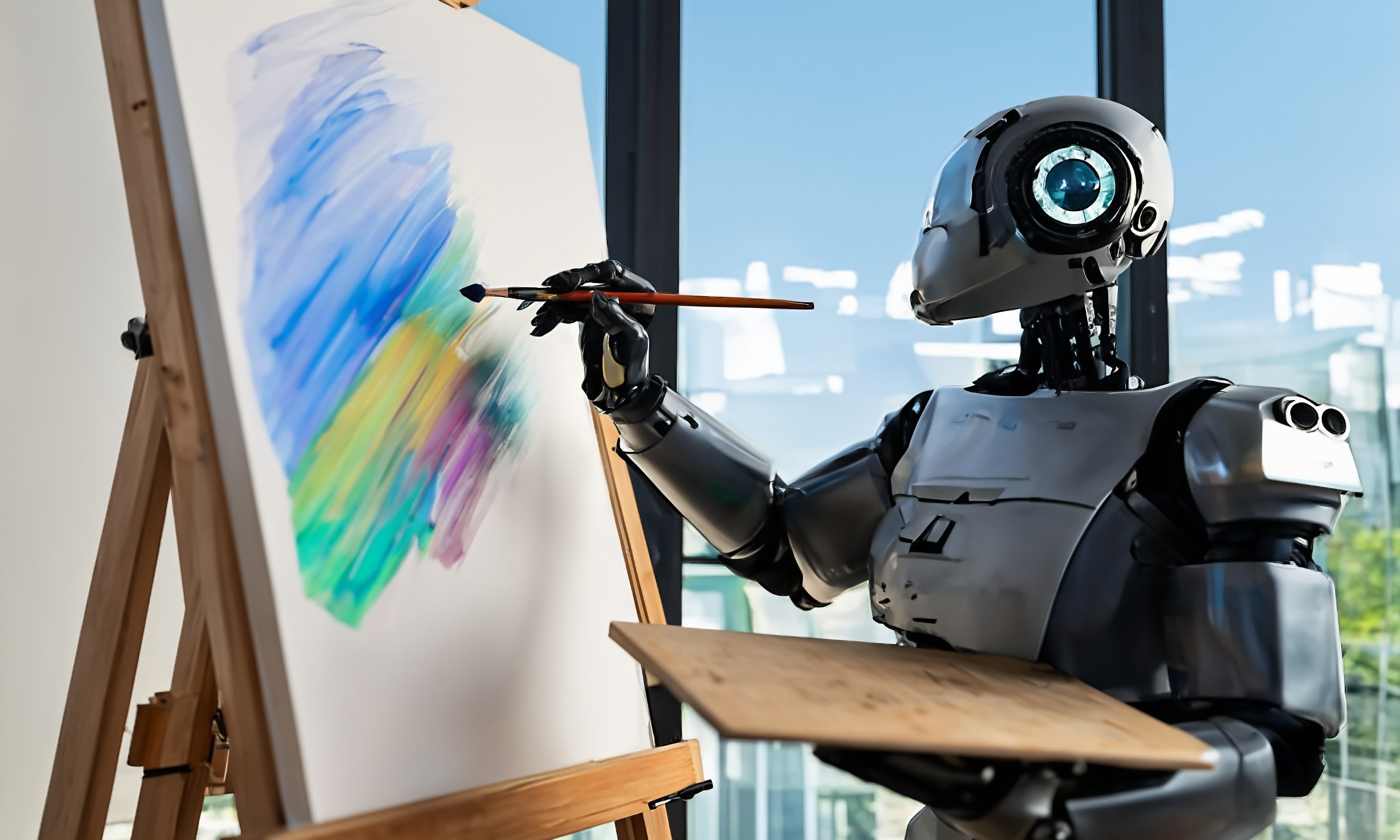}} &
    \multicolumn{2}{c}{\includegraphics[width=0.497\textwidth]{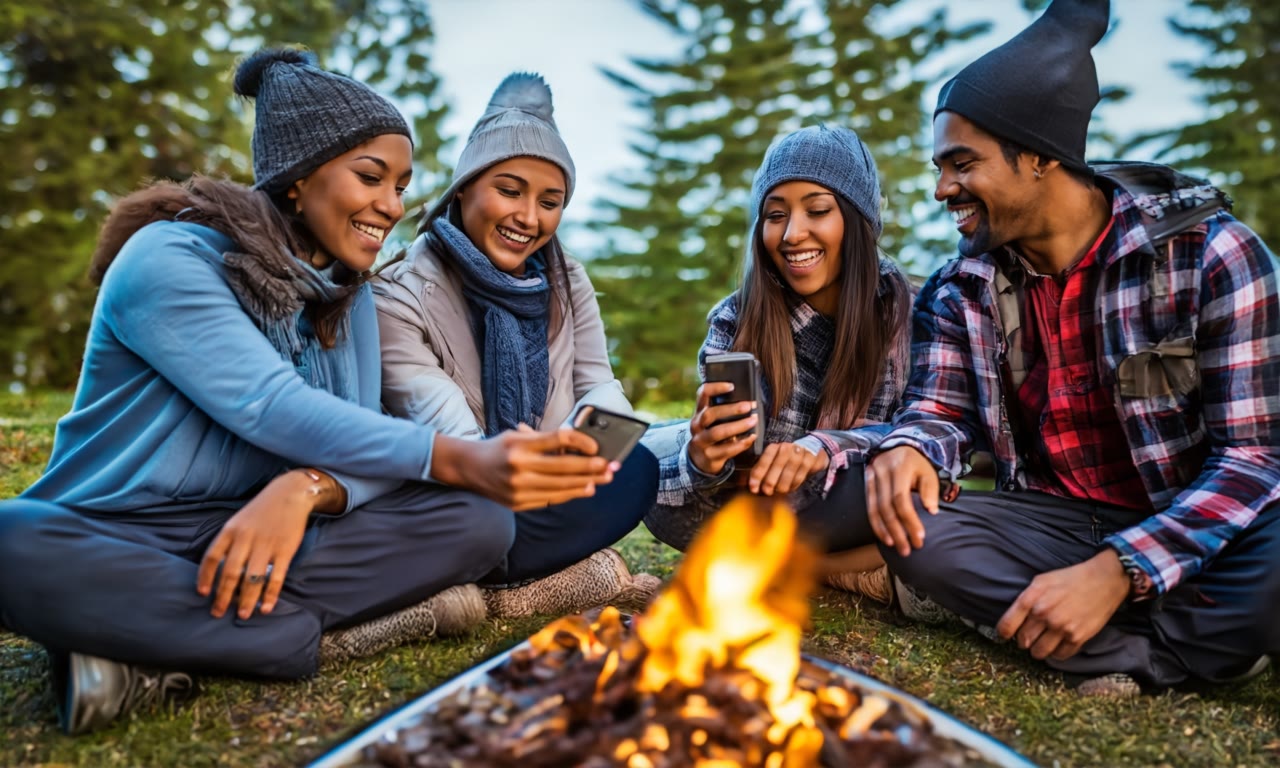}} \\[-0.5mm]
    \multicolumn{2}{c}{\makecell[{{p{0.48\textwidth}}}]{
            \emph{\small{A photo of a robot holding a brush and painting a picture.}}
    }} & 
    \multicolumn{2}{c}{\makecell[{{p{0.48\textwidth}}}]{
            \emph{\small{A group of friends sitting around a campfire.}}
    }} \\[-0.5mm]
    
    \multicolumn{2}{c}{\includegraphics[width=0.497\textwidth]{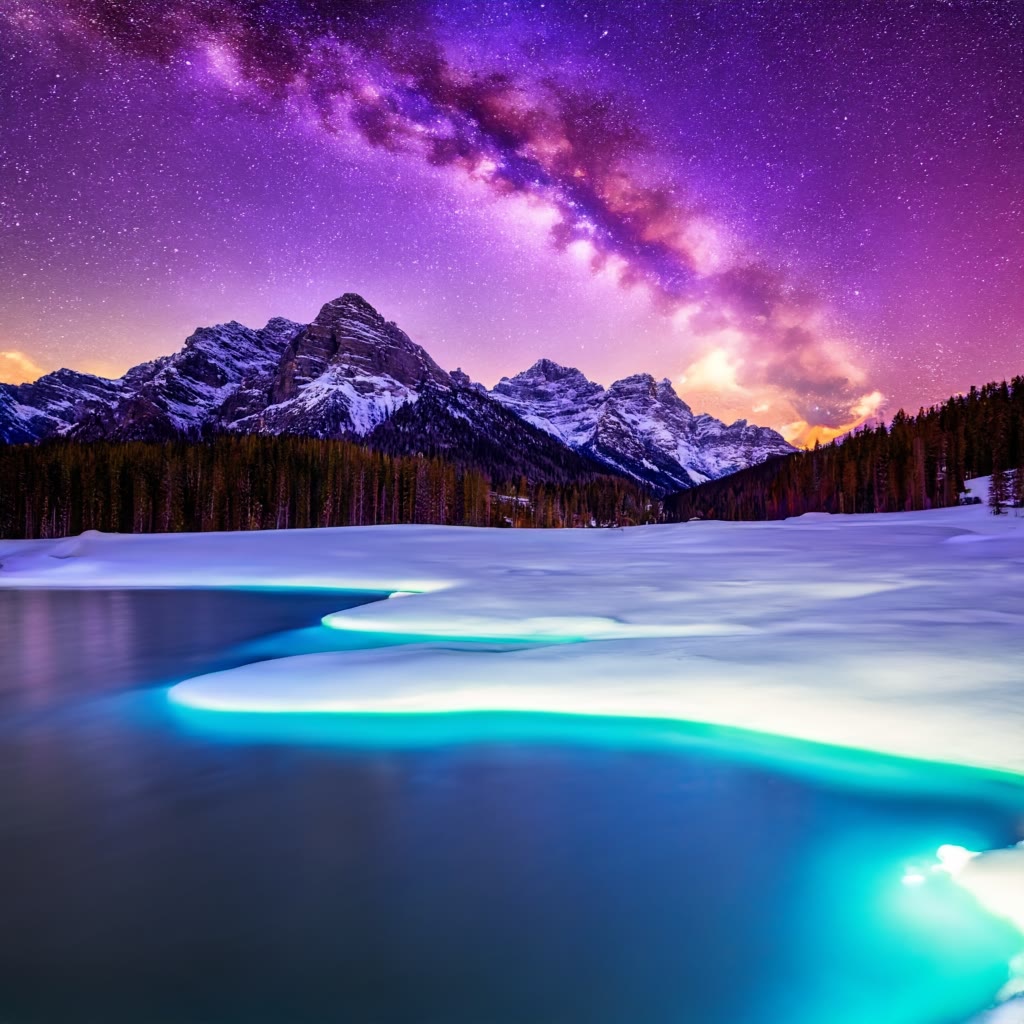}} &
    \multicolumn{2}{c}{\includegraphics[width=0.497\textwidth]{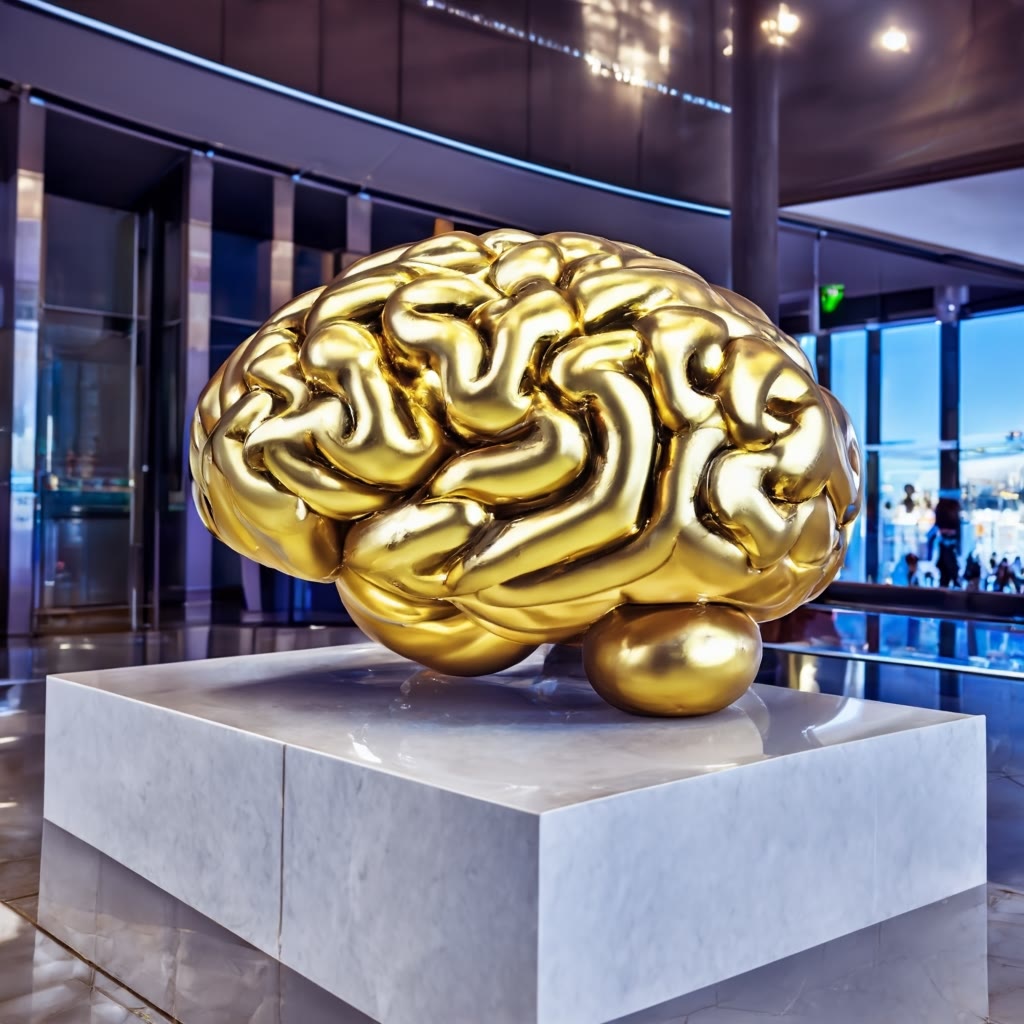}} \\[-0.5mm]
    
    \multicolumn{2}{c}{\makecell[{{p{0.48\textwidth}}}]{
            \emph{\small{A beautiful nature scene with snowy mountains, purple sky and bioluminescent blue icy lake}}
    }} & 
    \multicolumn{2}{c}{\makecell[{{p{0.48\textwidth}}}]{
            \emph{\small{Large golden human brain sculpture on a marble pedestal in a modern museum}}
    }} \\[-0.5mm]

    \includegraphics[width=0.245\textwidth]{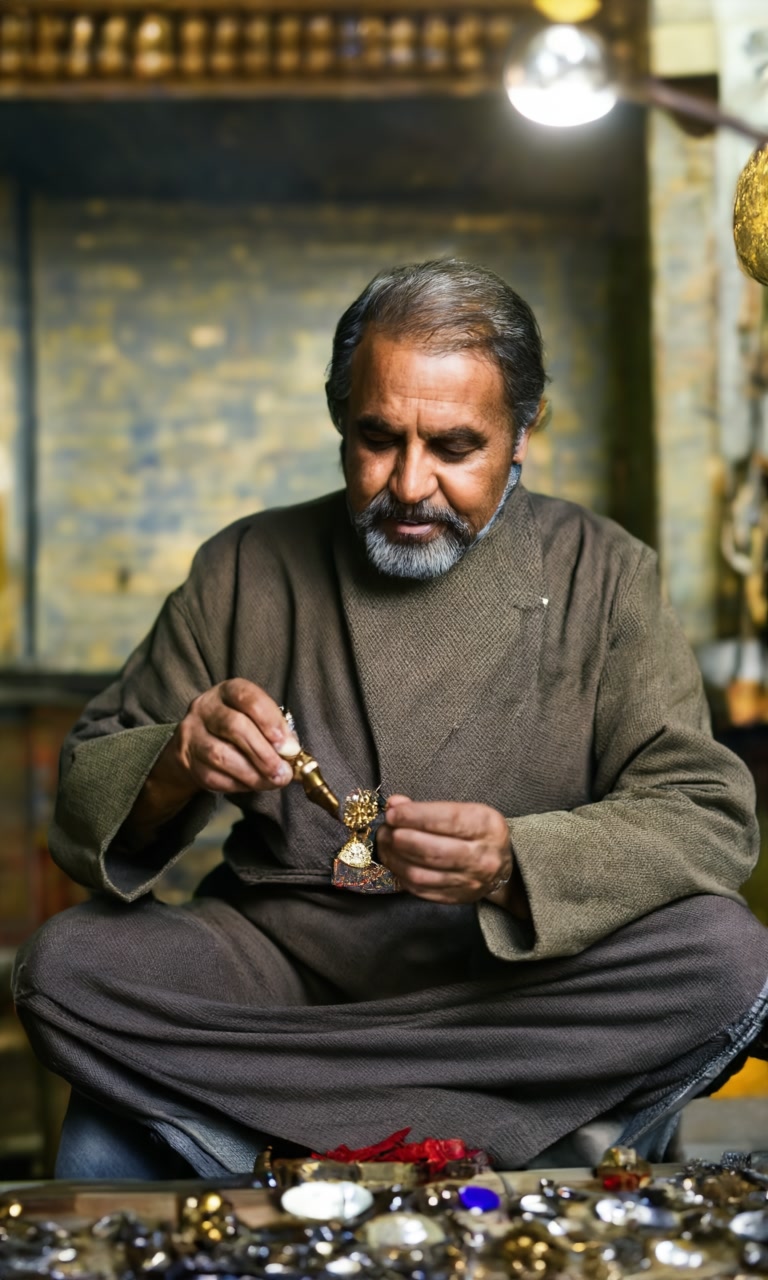} & \includegraphics[width=0.245\textwidth]{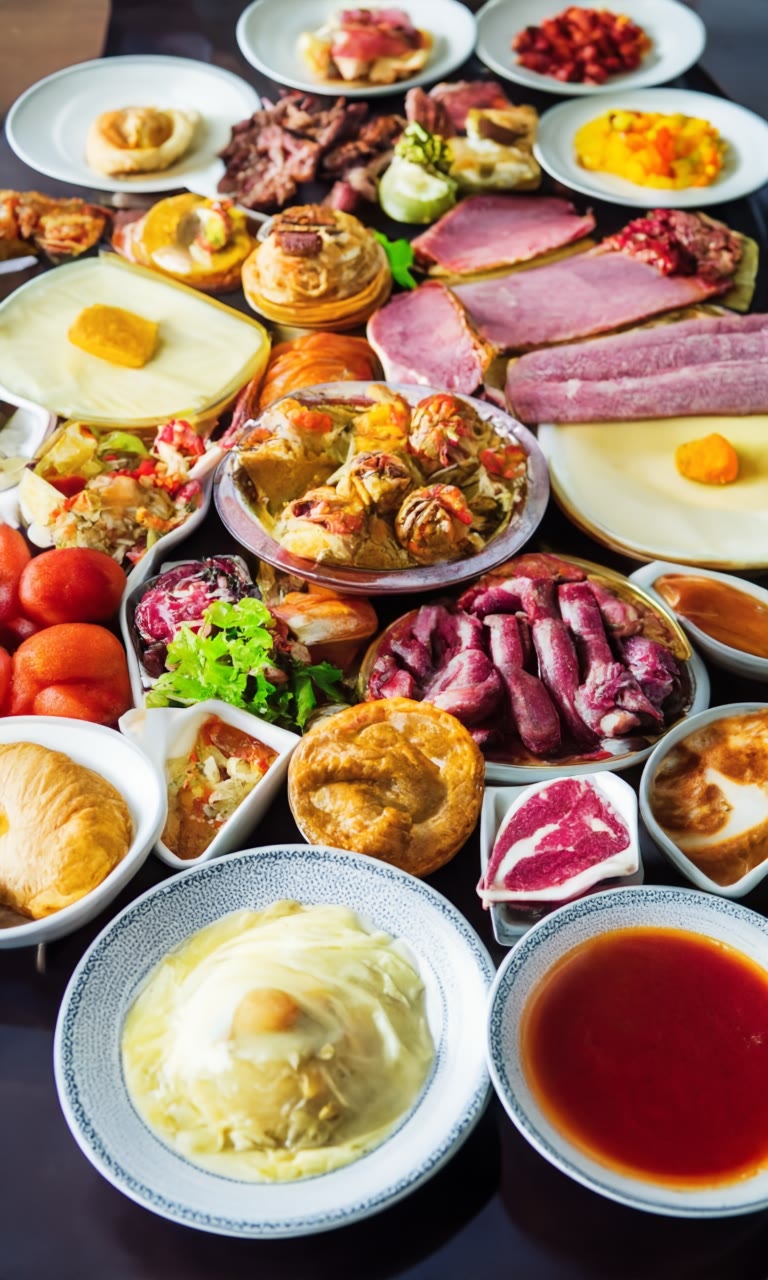} & \includegraphics[width=0.245\textwidth]{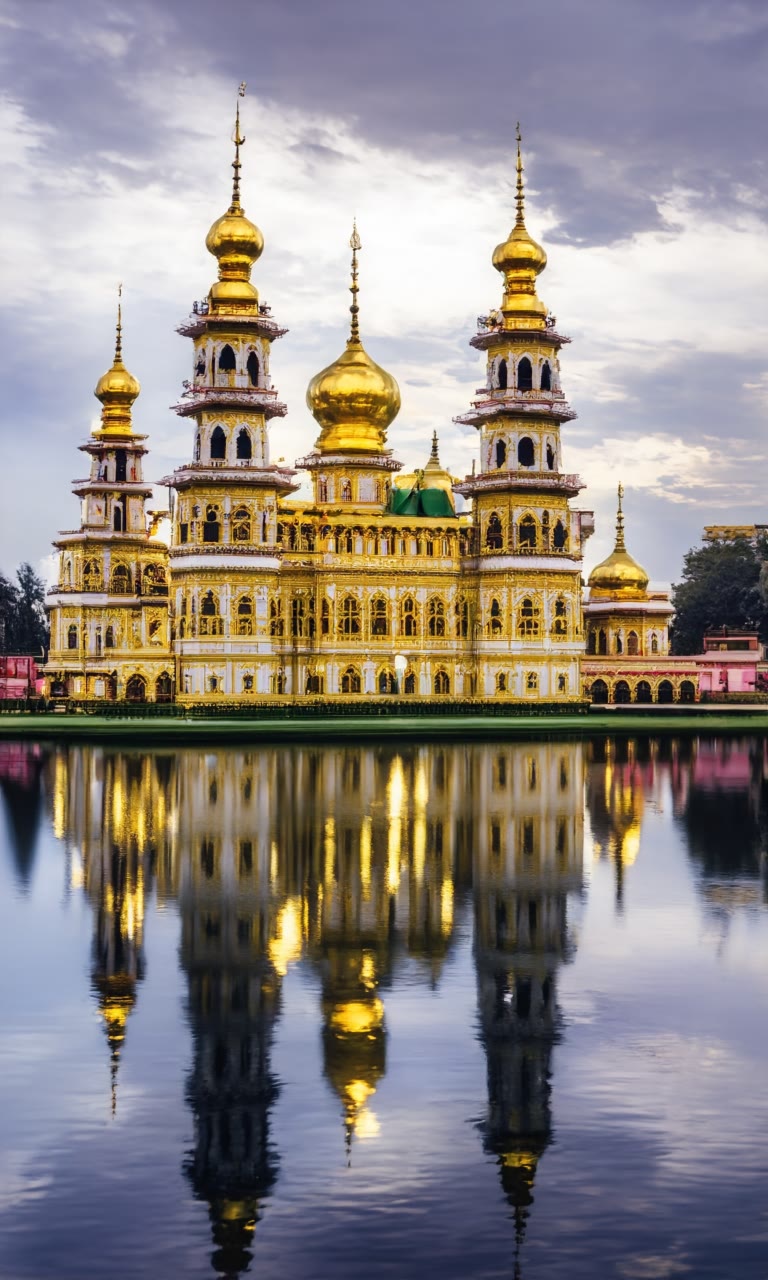} & \includegraphics[width=0.245\textwidth]{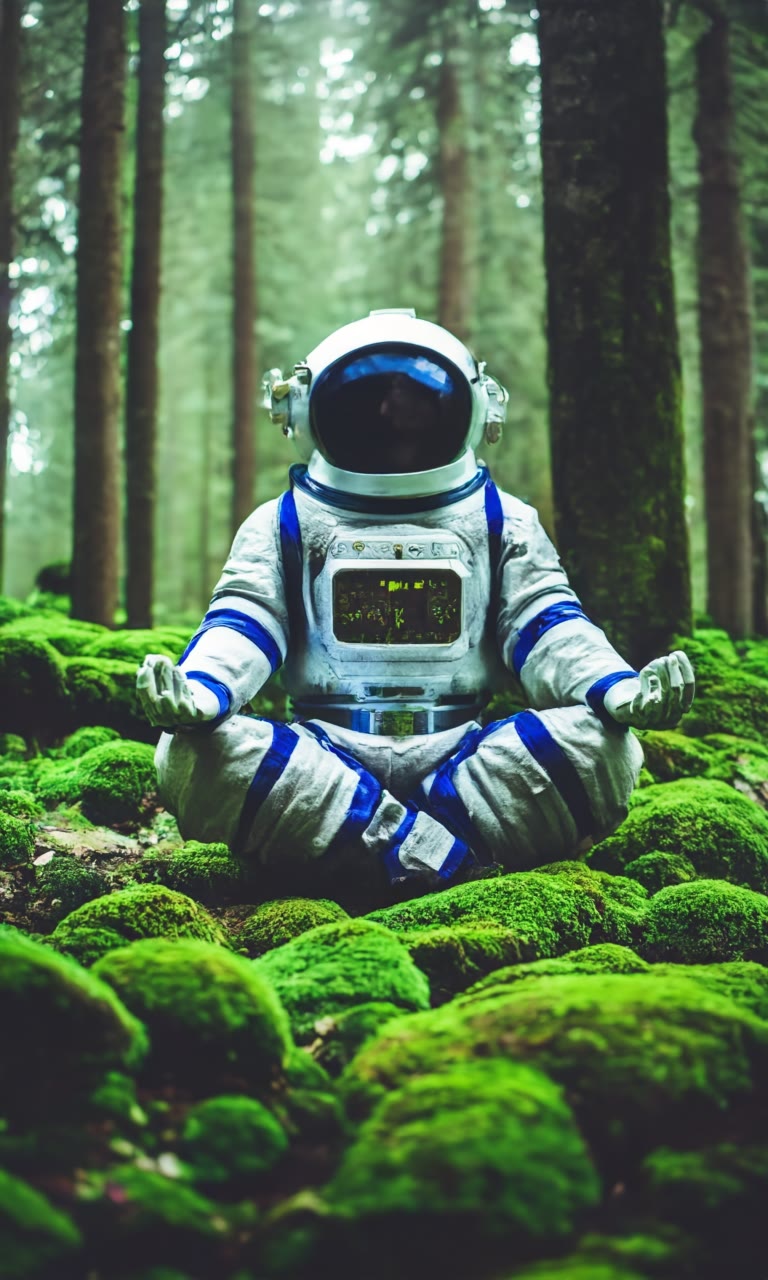}\\[-0.5mm]

    \makecell[{{p{0.22\textwidth}}}]{
            \emph{\small{A photo of an old goldsmith making jewelry.
            }}
    } & 
    \makecell[{{p{0.22\textwidth}}}]{
            \emph{\small{A dining table with lots of dishes and pastries.}}
    } &
    \makecell[{{p{0.22\textwidth}}}]{
            \emph{\small{A beautiful palace made of gold by a lake.
            }}
    } & 
    \makecell[{{p{0.22\textwidth}}}]{
            \emph{\small{An astronaut meditating in a lush green forest.
            }}
    } \\[-0.5mm]
\end{tabularx}
\vspace{-5pt}
\captionof{figure}{\textbf{Samples generated by our text-to-image model with 16:9, 1:1 and 9:16 aspect ratios.}}
\label{fig:generations_2d_main}
\end{table*}

\setlength{\tabcolsep}{0.5pt}
\renewcommand{\arraystretch}{0.5}

\begin{table}[htbp]
\centering
\begin{tabularx}{\textwidth}{M{0.49\textwidth} M{0.49\textwidth}}
    \includegraphics[width=0.485\textwidth]{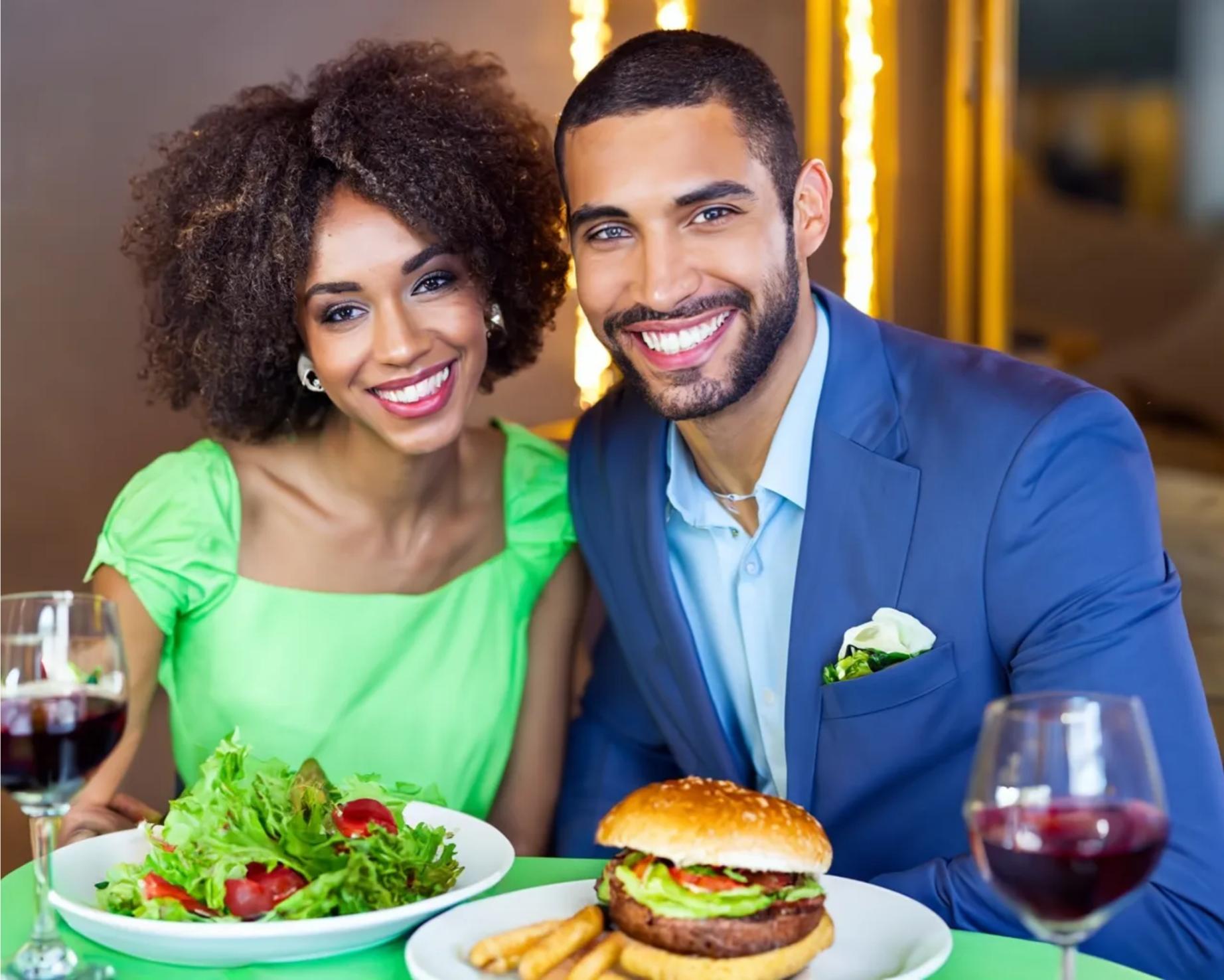} & \includegraphics[width=0.485\textwidth]{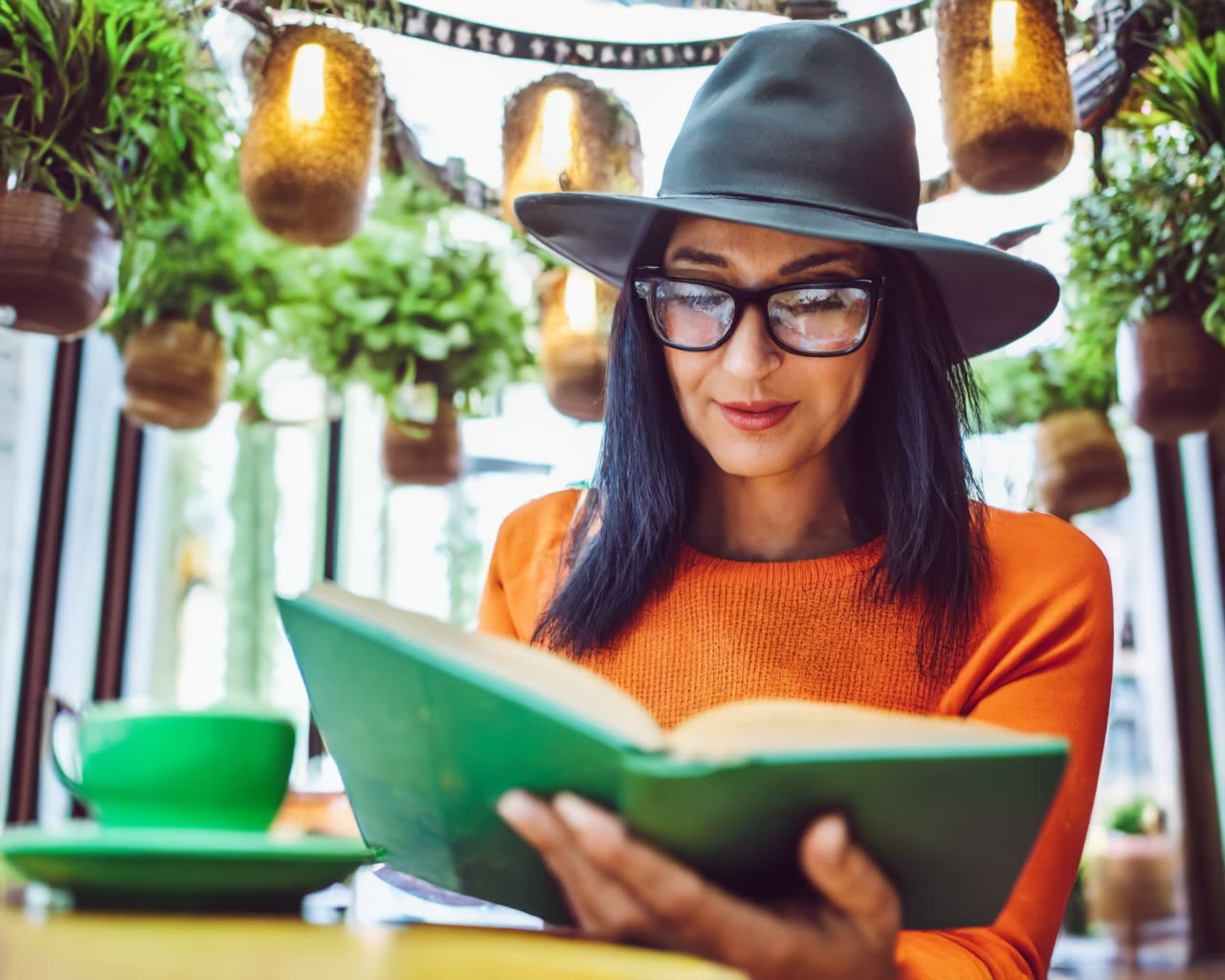} \\
    
    \makecell[{{p{0.47\textwidth}}}]{
            \emph{\small{A happy couple is having a romantic dinner at a restaurant. In the table, there is a burger, fries, two glasses of red wine and a salad. The man is wearing a blue suit and the woman is wearing a green dress.}}
    } & 
    \makecell[{{p{0.47\textwidth}}}]{
            \emph{\small{A photo of a woman in a coffee shop reading a book. She is wearing glasses. Her hair is black in color. She is wearing a fancy hat and an orange shirt. There is a green coffee cup on a saucer placed next to the book. There are many plants in the background with string lights on the ceiling.}}
    } \\[4mm]
    \includegraphics[width=0.485\textwidth]{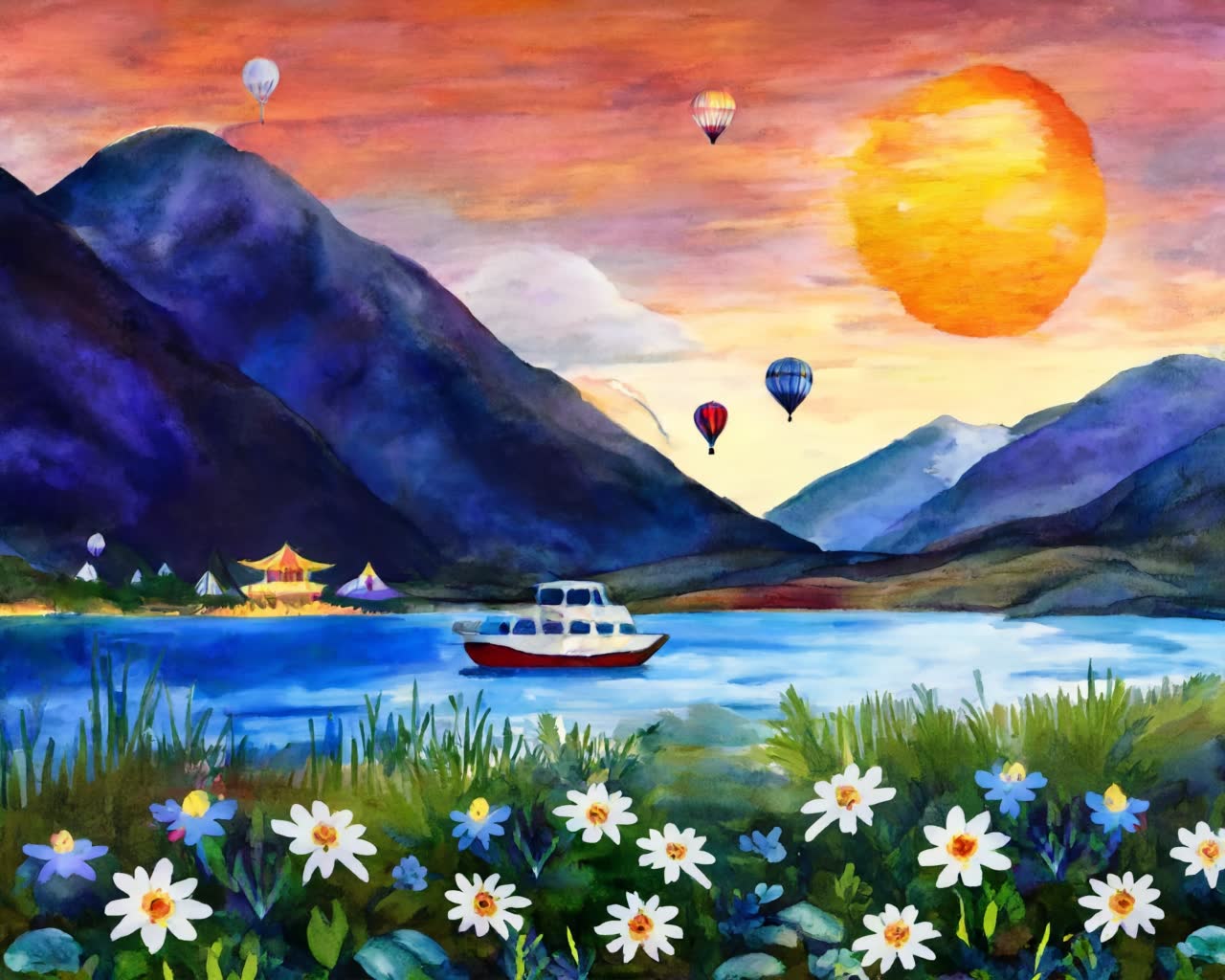} & \includegraphics[width=0.485\textwidth]{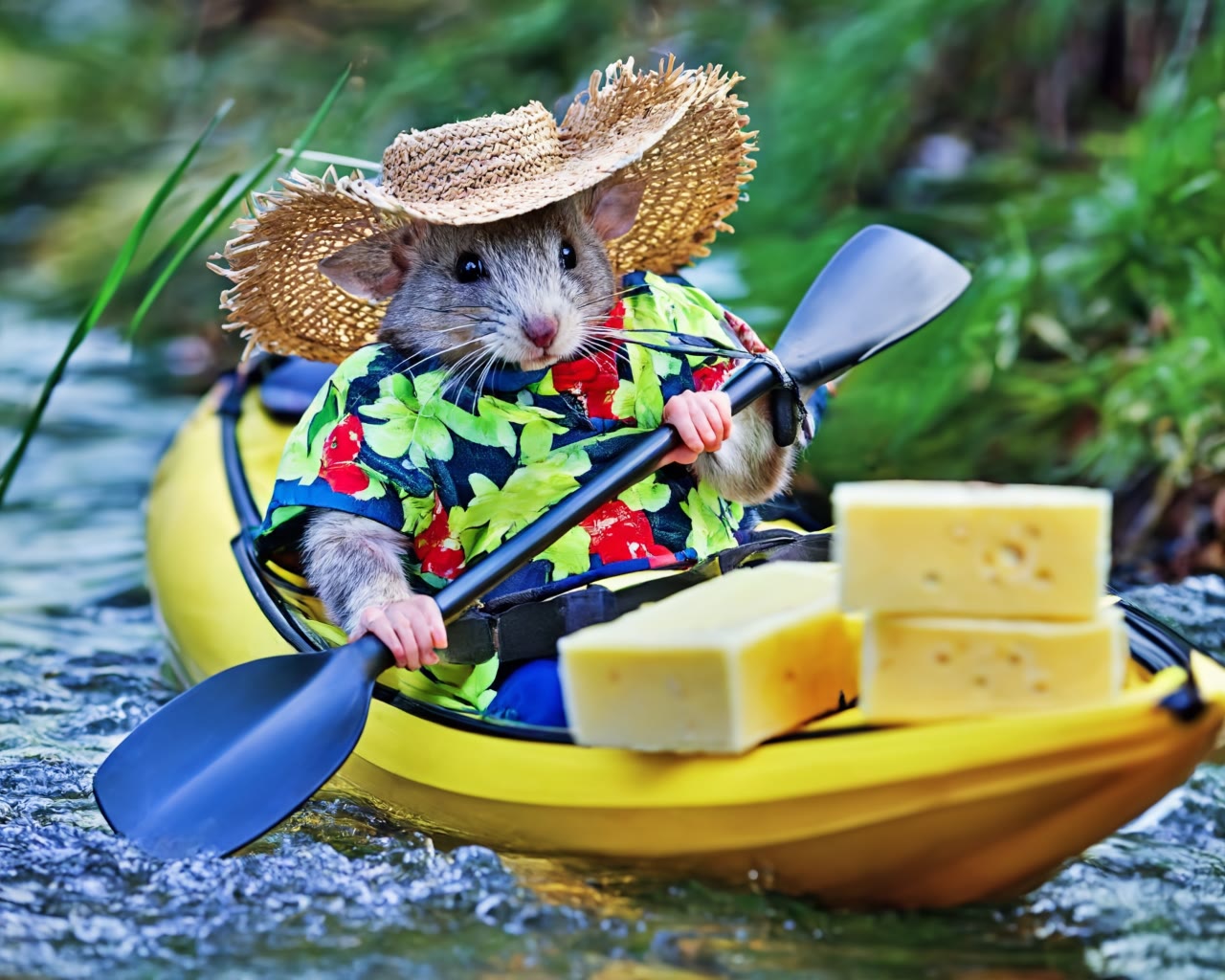} \\
    
    \makecell[{{p{0.47\textwidth}}}]{
            \emph{\small{Watercolor painting of a nature scene. There are mountains in the background. There is a field with white and blue flowers in the foreground. There is a lake under the mountain where a boat is sailing. The sun is setting with orange sky. There are many hot air balloons floating in the sky.}}
    } & 
    \makecell[{{p{0.47\textwidth}}}]{
            \emph{\small{A 4K dslr photo of a mouse kayaking in a stream of water set against the backdrop of a lush green forest. The mouse is wearing a Hawaiian shirt and a straw hat. There are several blocks of cheese stacked in the kayak. }}
    } \\
\end{tabularx}
\captionof{figure}{\textbf{Long prompt generation.} Edify Image can faithfully generate images from long descriptive prompts.}
\label{fig:generations_long_prompts}
\end{table}

To generate images of $1024$ resolution, we train a two-stage cascaded pixel-space diffusion model where the first model generates an image of $256$ resolution while the second model upscales the image to $1024$ resolution. Our training pipeline is provided in~\cref{fig:model_arch}. The $256$ resolution model is trained on the full noise range $[0, \sigma^{max}_{256}]$, while the $1K$ model operates on a smaller noise range $[0, \sigma^{max}_{1024}]$ ($\sigma^{max}_{1024} < \sigma^{max}_{256}$). During inference, we first generate the $256$ resolution sample by running the full sampling loop on the base model. Then, we apply forward diffusion on the generated sample with $\sigma = \sigma^{max}_{1024}$ and denoise the image using the upsampler. All our models are trained with the objective discussed in Section~\ref{sec:dim_varying_edm}.

\subsection{Model Architecture}

We use U-Net-based architecture for the base and upsampling models following \citet{ho2020denoisingdiffusionprobabilisticmodels, DALLE2, saharia2022photorealistic}. The U-Net model typically consists of a sequence of residual and attention blocks that progressively downsample (or upsample) feature maps with skip connections. For high-resolution synthesis, the spatial resolution of feature maps increases, which makes the computation of attention maps expensive. To address this issue, we propose to operate on the smaller spatial resolution by using invertible wavelet transforms at the beginning and the end of the network, inspired by \cite{hoogeboom2023simple}. This is illustrated in~\cref{fig:model_arch}.  We use 2-level Haar wavelets to downsample the images in the pixel space from resolution $(3 \times H \times W)$ to $(48 \times (H / 4) \times (W / 4))$. This reduces the number of spatial tokens in the attention layers by a factor of 16, dramatically improving the training efficiency. Our base model consists of $2.7B$ parameters, while the $1K$ upsampler consists of $1.6B$ parameters.

\subsection{Conditioning Inputs}\label{sec:conditioning_inputs}

Prior text-to-image generators \cite{DALLE2, podell2023sdxl} mainly use text embeddings from CLIP~\cite{CLIP} and T5~\cite{raffel2020exploring} models as conditional inputs. To provide better controllability to our image generators, we use the following conditioning inputs.

\begin{itemize}
    \item \textbf{T5 embeddings.} We use text embeddings from the T5-XXL model. To enable support for long prompt generation, we use a sequence length of $512$.
    \item \textbf{Camera embeddings.} To provide better camera control while generating images, we additionally condition the synthesis using camera attributes. For each image, we obtain integer-valued pitch and depth of field annotations. These annotations are then passed through an embedding layer and used as a conditional signal during training.
    \item \textbf{Media type.} Each image in the dataset is assigned a media type label such as `Photography' and `Illustration', which is then used as a conditional attribute during training.
\end{itemize}
All conditional embeddings are then concatenated along the sequence dimension and used in the cross-attention layer in the U-Nets. During training, we apply random embedding dropout to each of the conditional embeddings. This ensures that the model can generate using any combination of conditional signals. When all embeddings are dropped out, we obtain the unconditional score.

\subsection{Data}
We train various Edify Images models for our AI foundry partners, who are responsible for sourcing the image dataset. To achieve better prompt alignment, having detailed and descriptive captions is critical, as shown in ~\cite{betker2023improving}. So, in addition to the ground truth captions, we use LLM based captioners to obtain long descriptive captions. During training, we randomly sample captions from ground truth and AI generations. This allows our model to generate images from both long and short text prompts.

\setlength{\tabcolsep}{1pt}
\renewcommand{\arraystretch}{0.5}
\begin{figure*}[tb!]
    \centering
    \begin{tabular}{cccc}
        \includegraphics[width=0.23\textwidth]{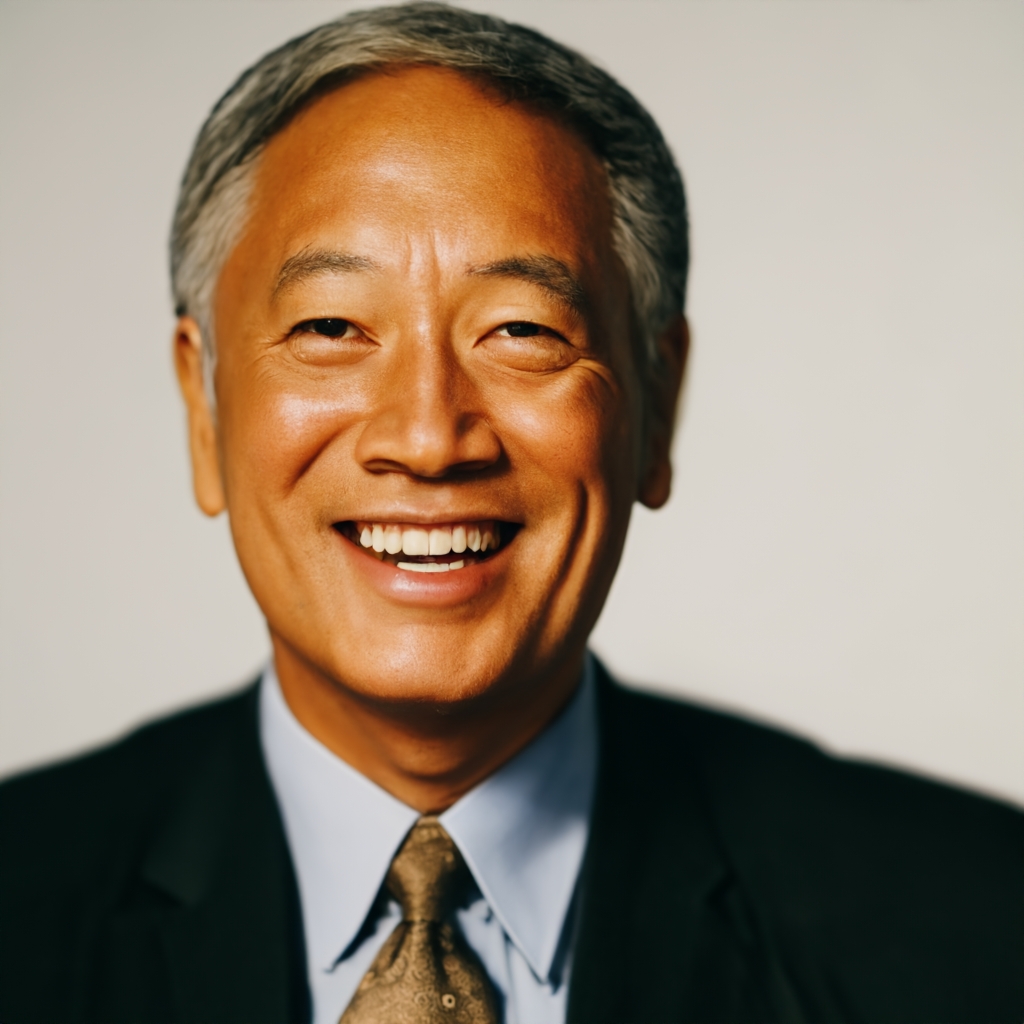} &
        \includegraphics[width=0.23\textwidth]{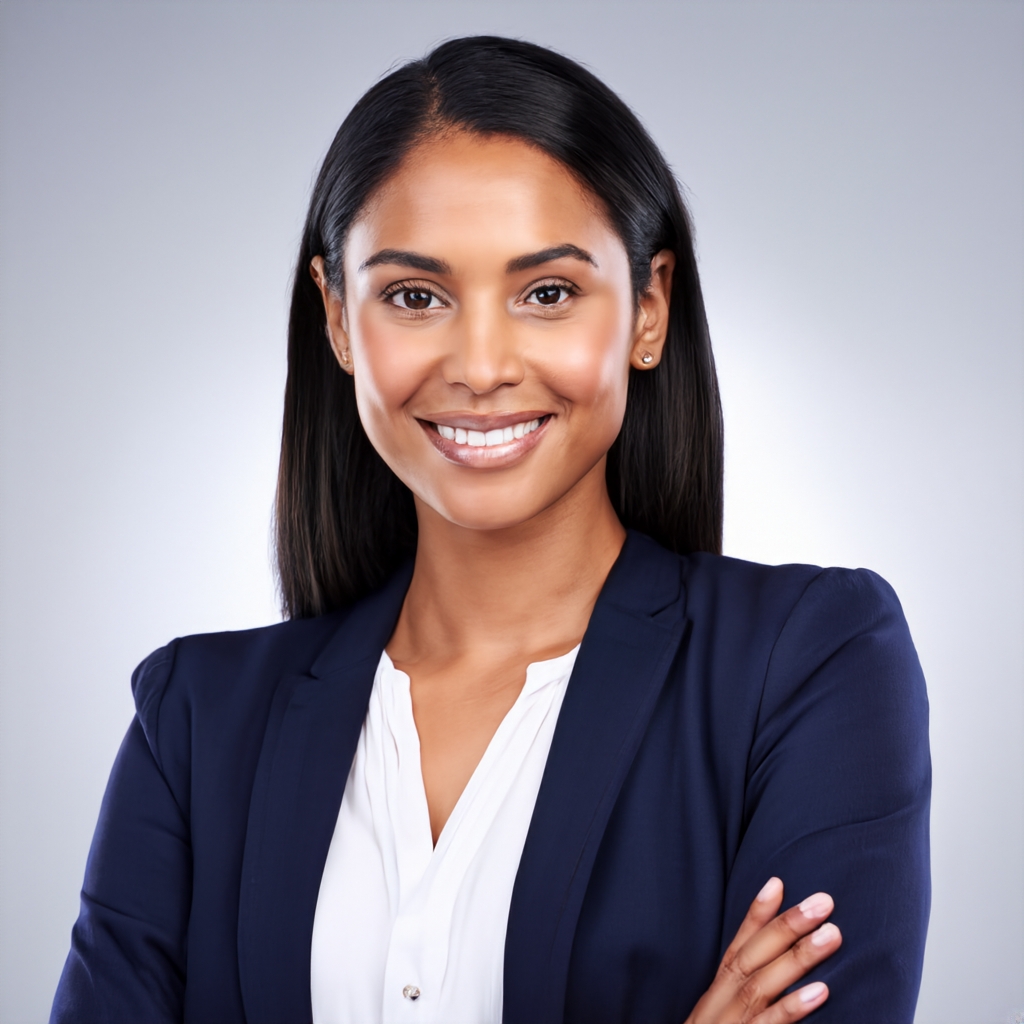} &
        \includegraphics[width=0.23\textwidth]{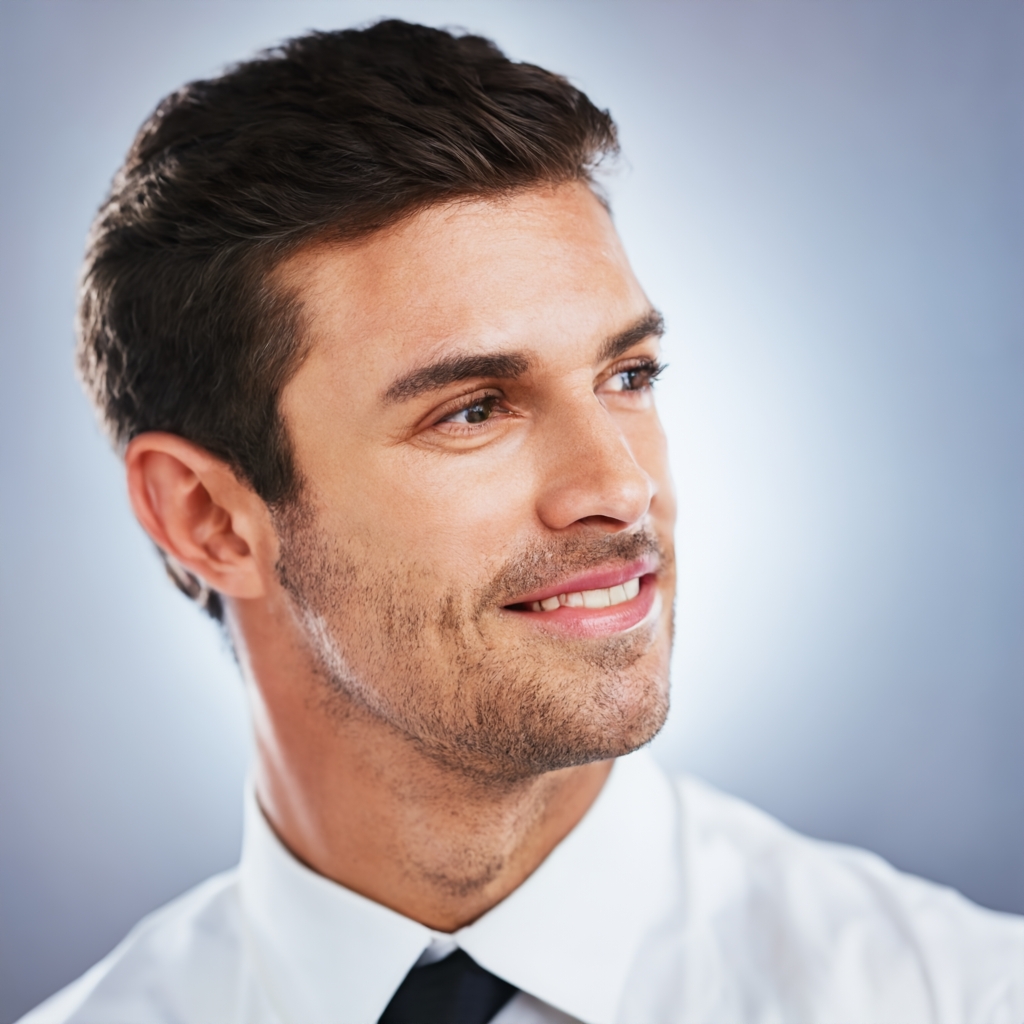} &
        \includegraphics[width=0.23\textwidth]{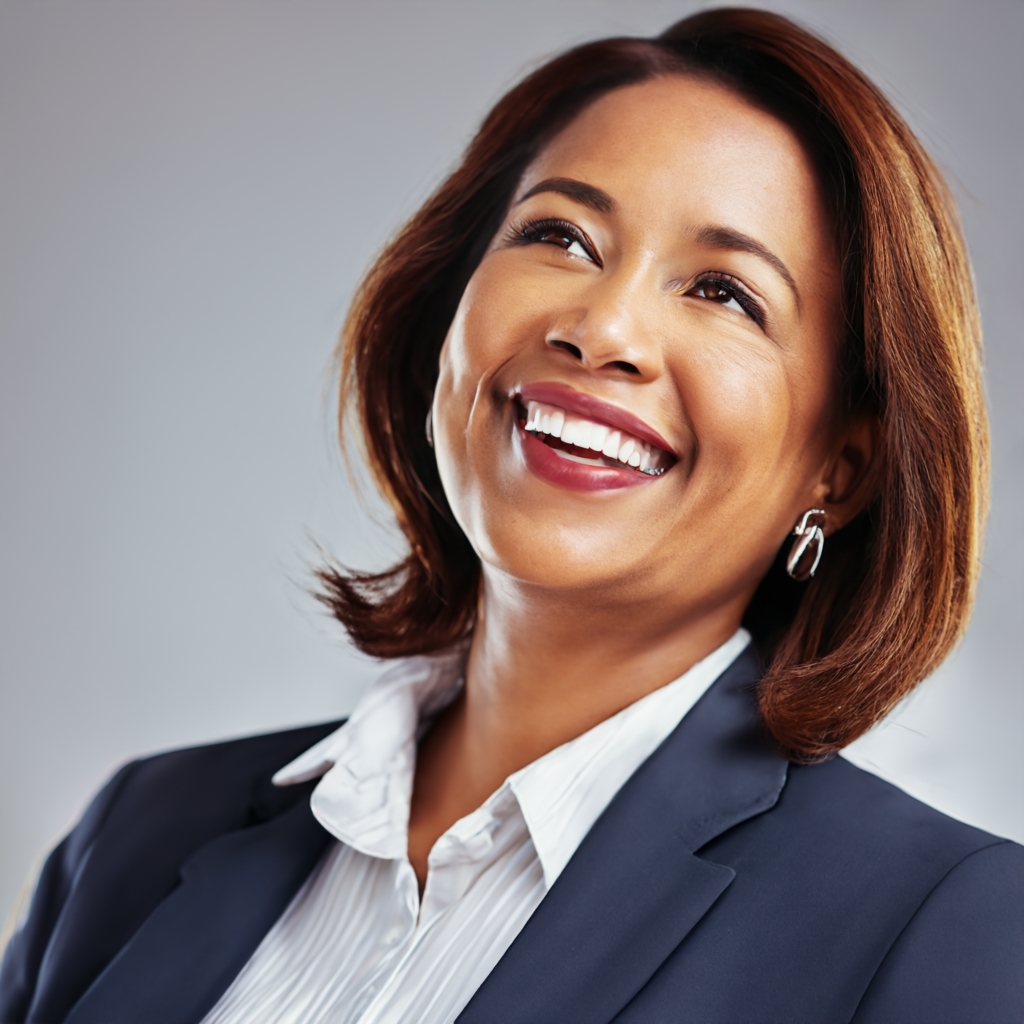} \\
        \includegraphics[width=0.23\textwidth]{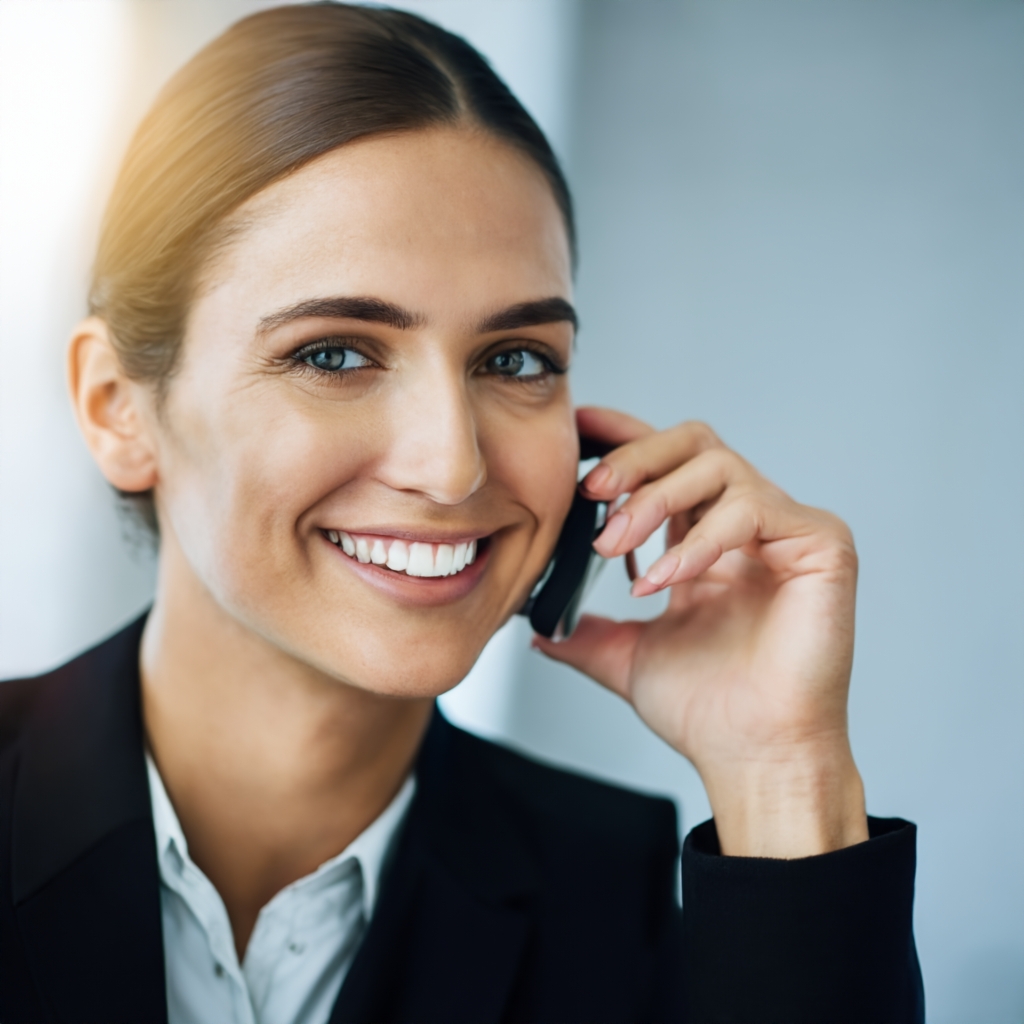} &
        \includegraphics[width=0.23\textwidth]{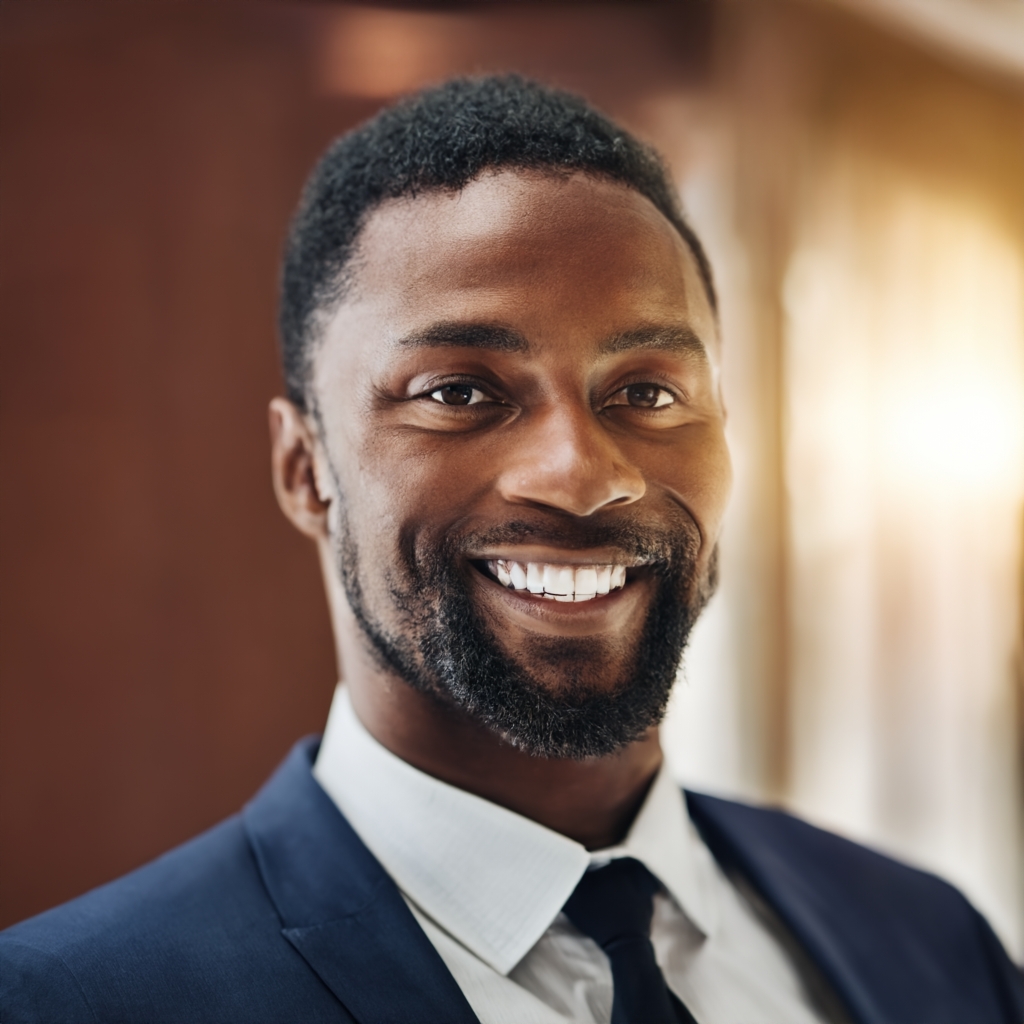} &
        \includegraphics[width=0.23\textwidth]{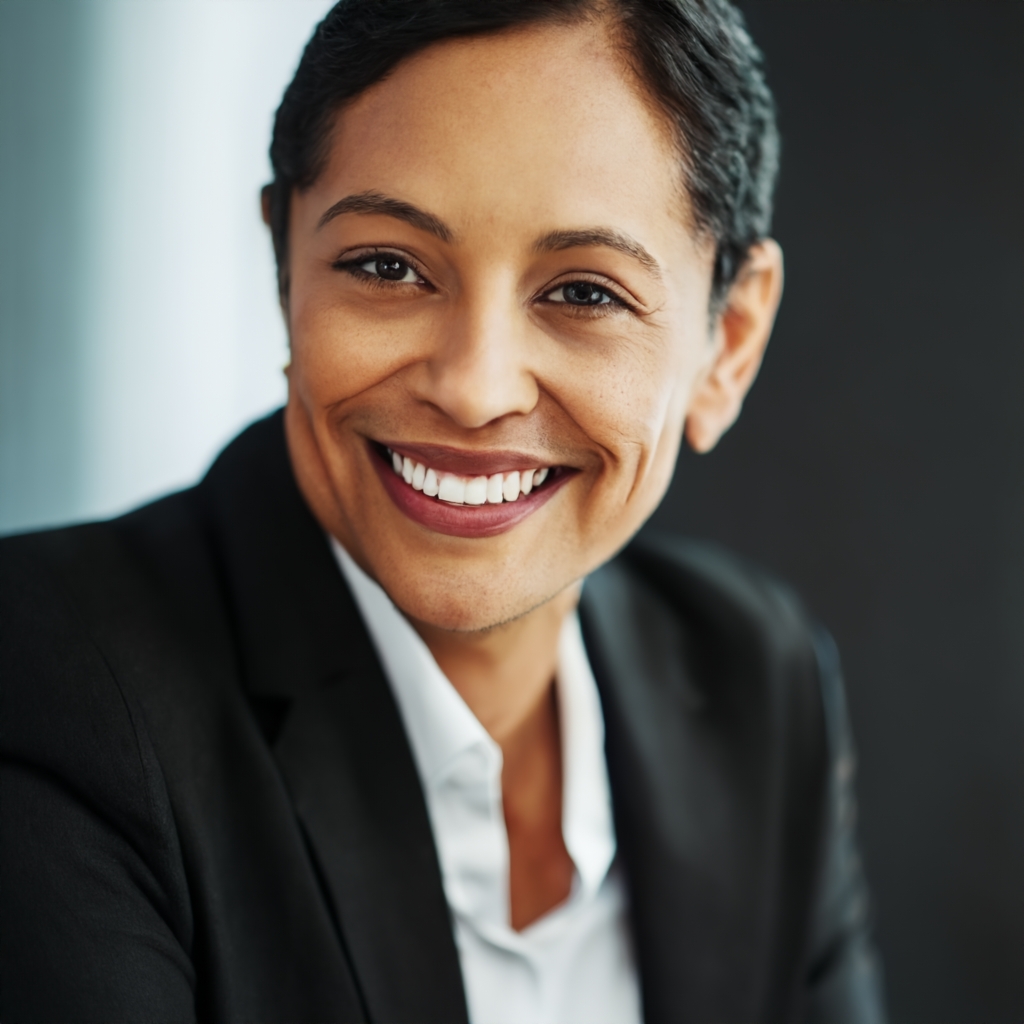} &
        \includegraphics[width=0.23\textwidth]{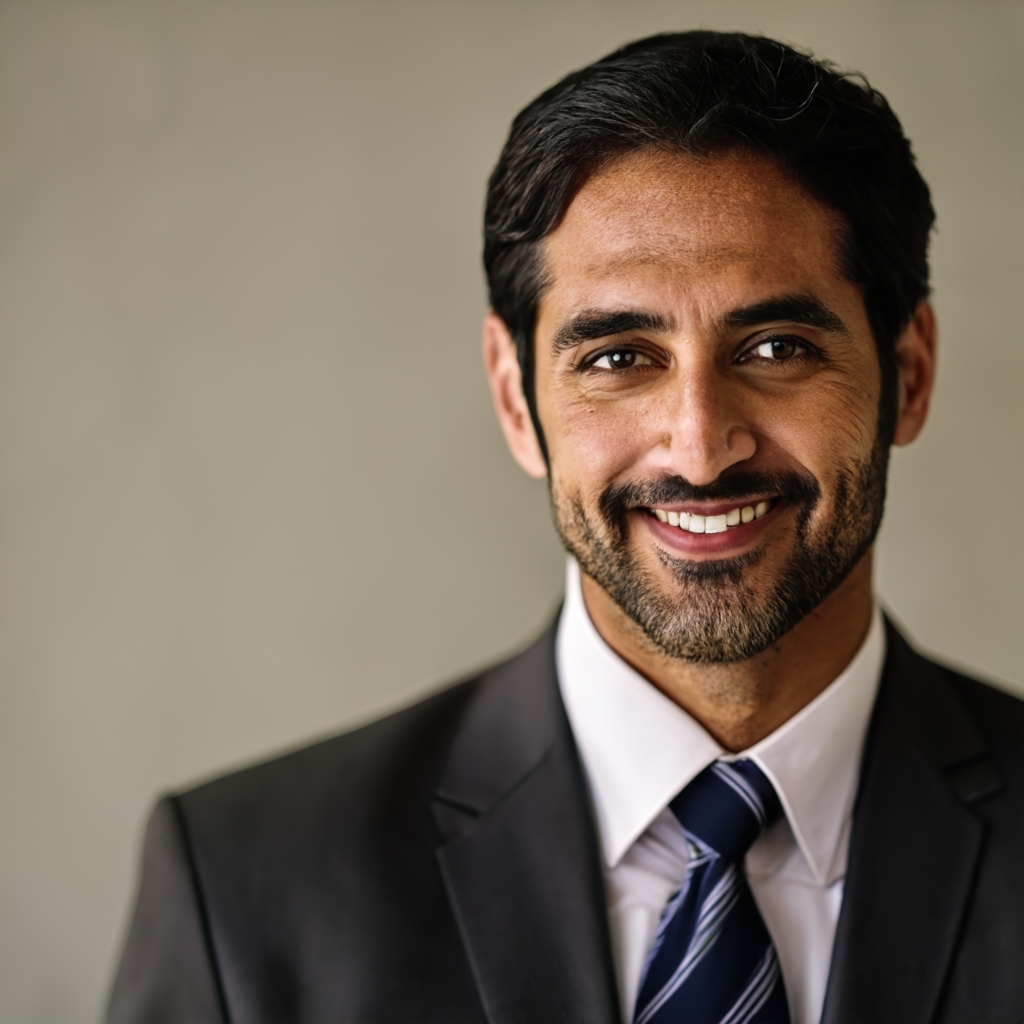} \\[2mm]
\end{tabular}
\caption{\textbf{Human diversity.} Our model is able to generate images with good gender and race diversity. The prompt used is \emph{"A studio portrait of a smart CEO"}.}
\label{fig:race_diversity}
\end{figure*}
\setlength{\tabcolsep}{1pt}
\renewcommand{\arraystretch}{0.5}
\begin{figure*}[tb!]
    \centering
    \begin{tabular}{ccc}
        {\centering Descending view} & {\centering Eye level view} & {\centering Ascending view} \\[2mm] 
        \includegraphics[width=0.32\textwidth]{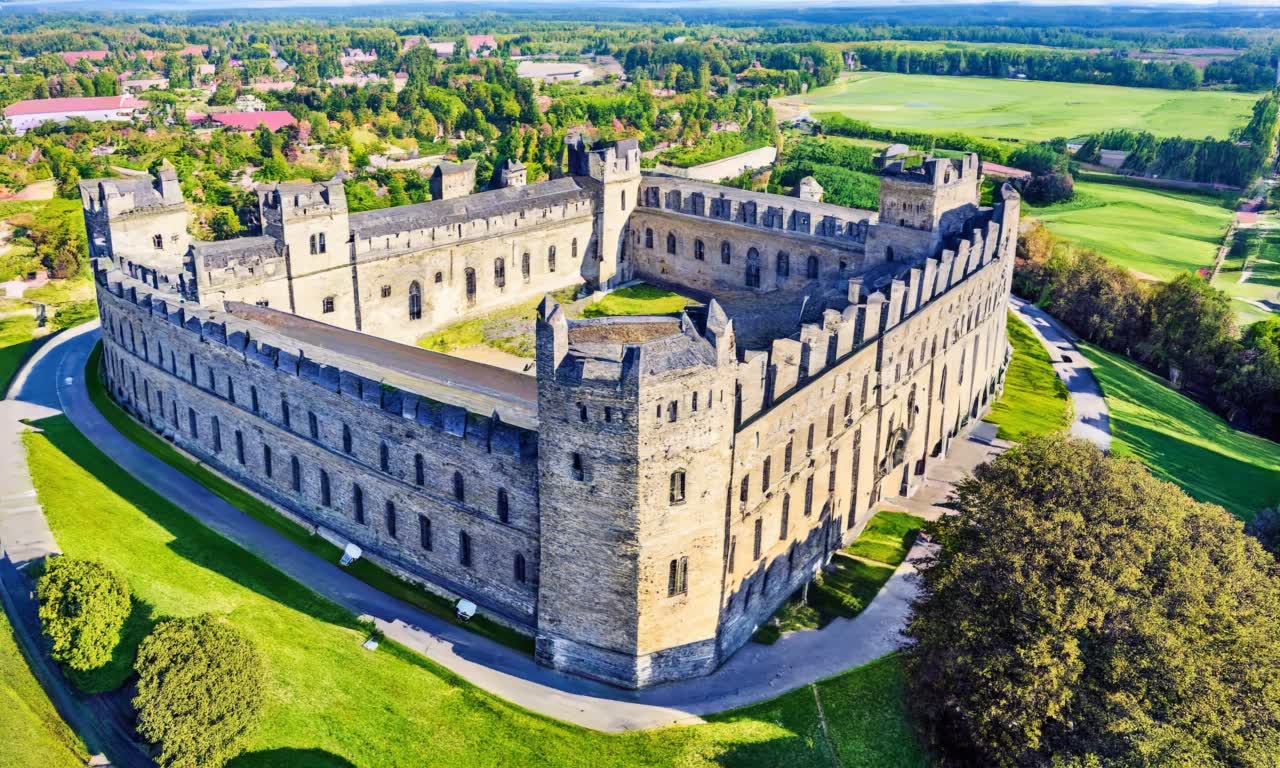} &
        \includegraphics[width=0.32\textwidth]{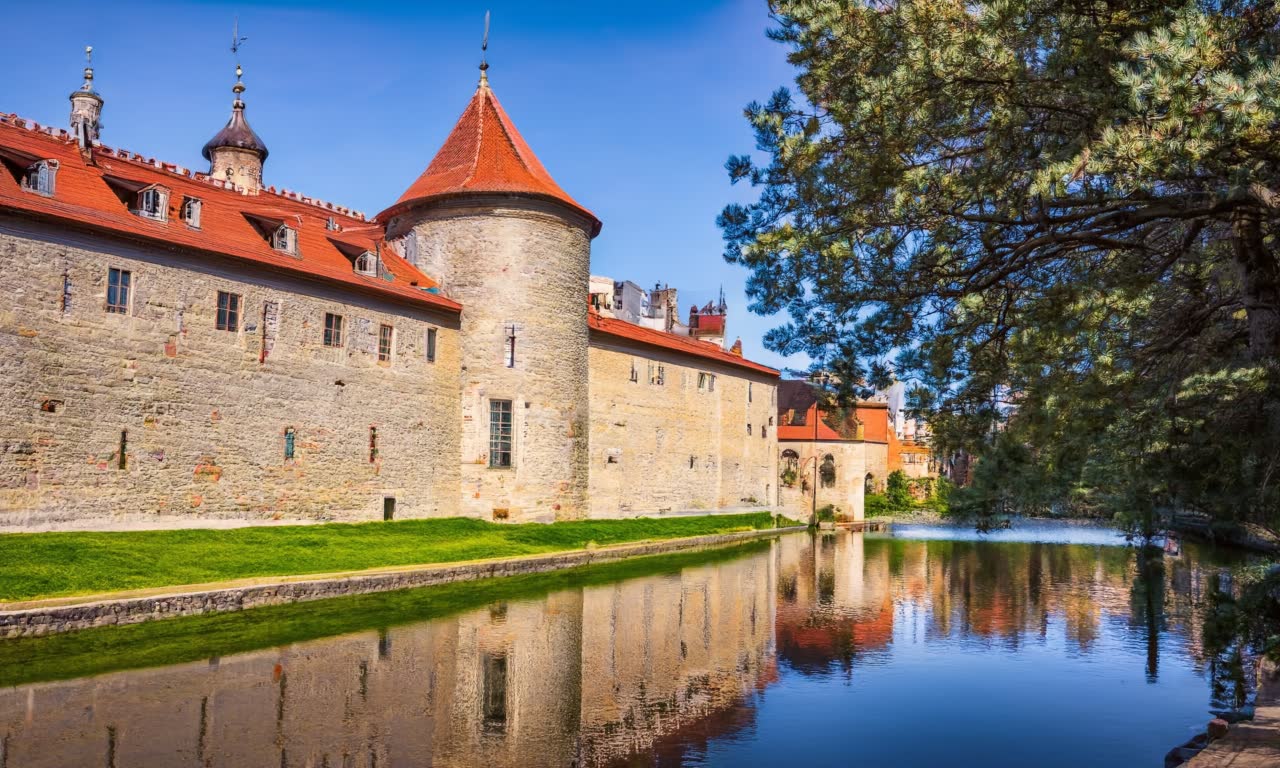} &
        \includegraphics[width=0.32\textwidth]{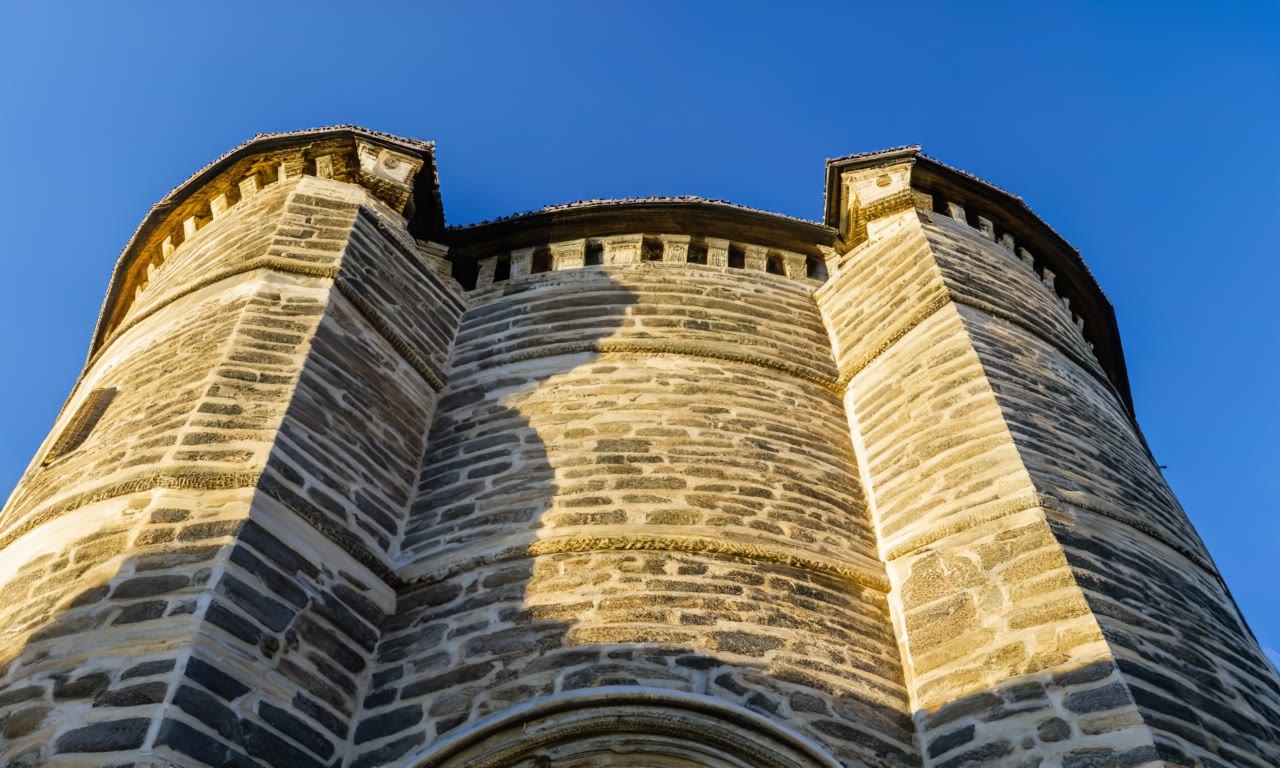} \\[2mm]
        \multicolumn{3}{c}{\emph{A photo of a beautiful ancient castle from medieval times}} \\[2mm] 
        \includegraphics[width=0.32\textwidth]{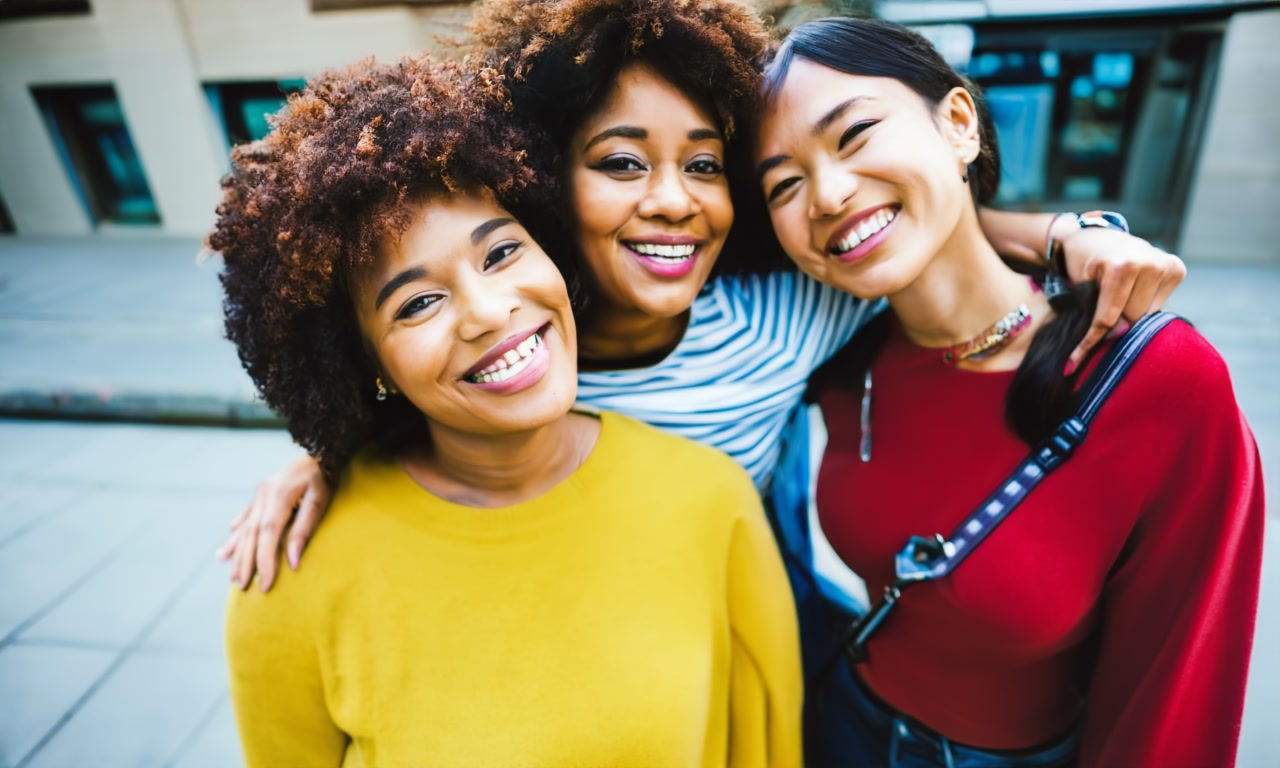} &
        \includegraphics[width=0.32\textwidth]{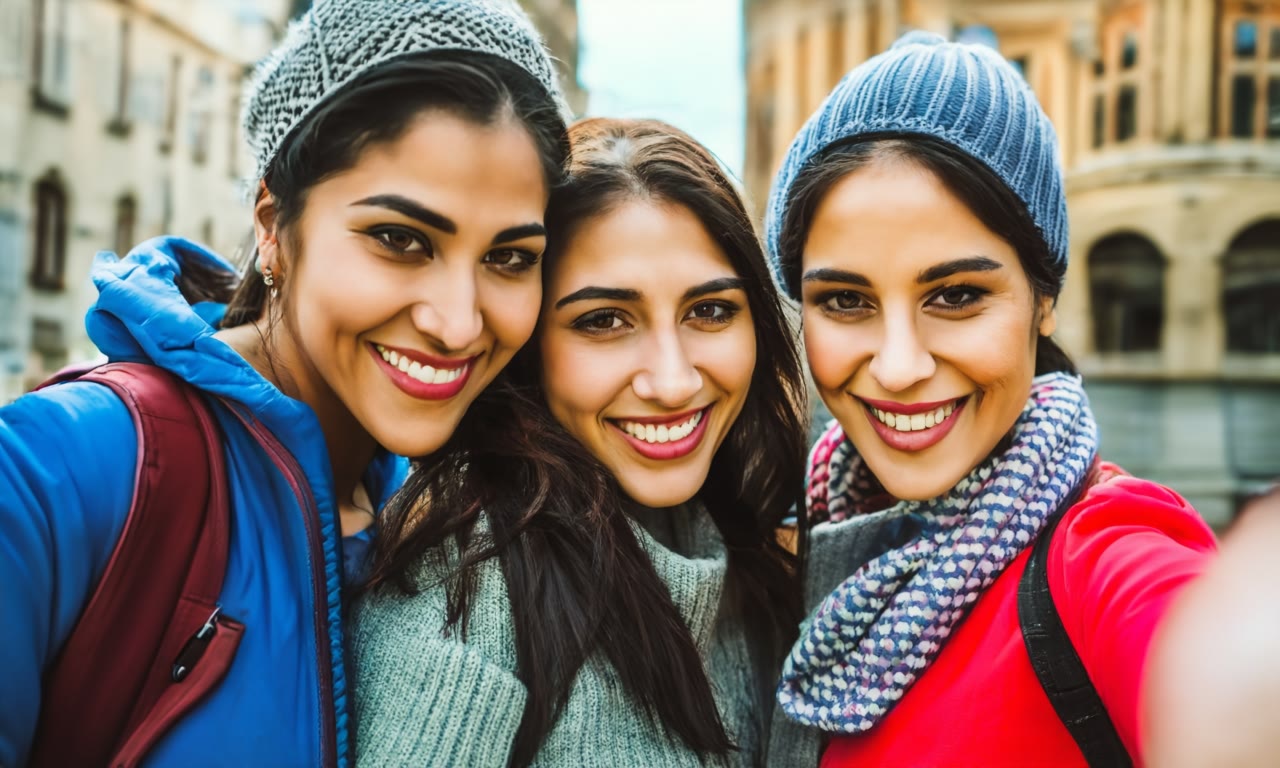} &
        \includegraphics[width=0.32\textwidth]{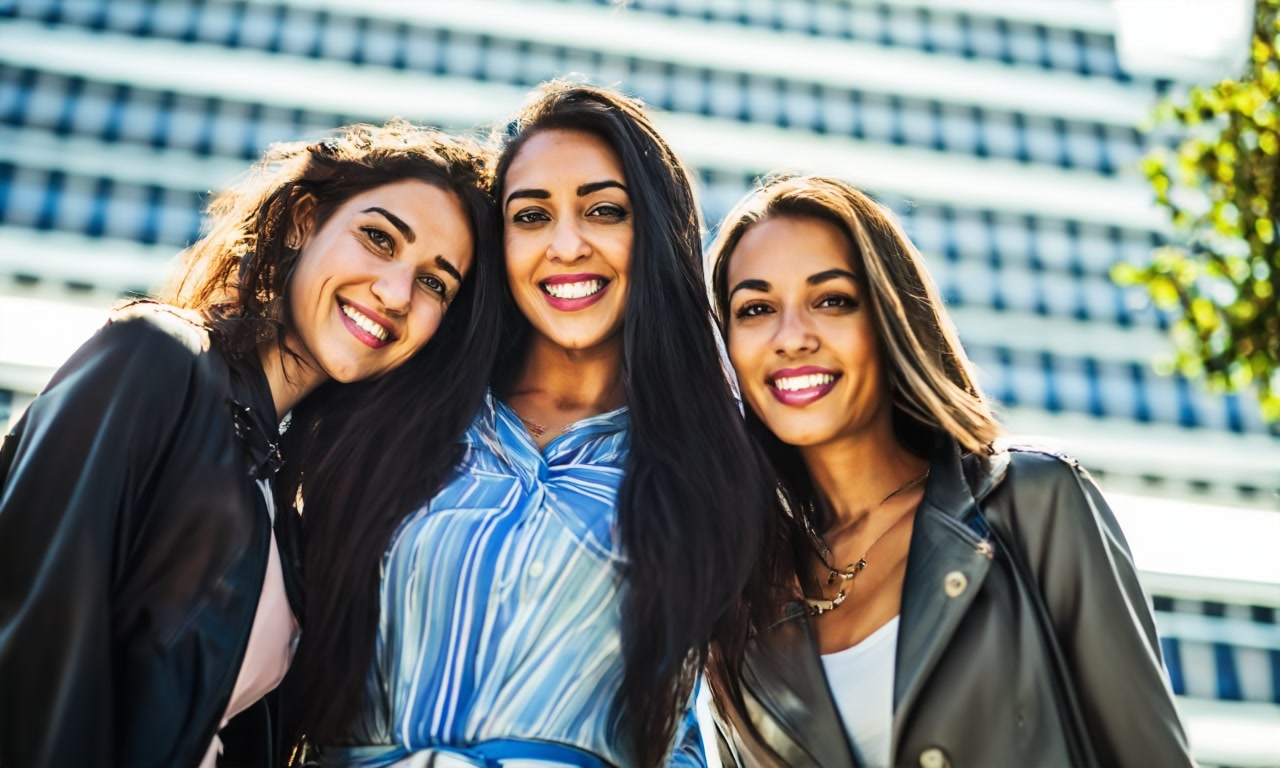} \\[2mm]
        \multicolumn{3}{c}{\emph{A photo of three women hanging out in a city}}
\end{tabular}
\caption{\textbf{Camera controls - Pitch.}}
\label{fig:camera_controls_pitch}
\end{figure*}


\begin{figure*}[tb!]
    \centering
    \begin{tabular}{cc}
        {\centering Shallow depth of field} & {\centering Deep depth of field}  \\[2mm] 
        \includegraphics[width=0.5\textwidth]{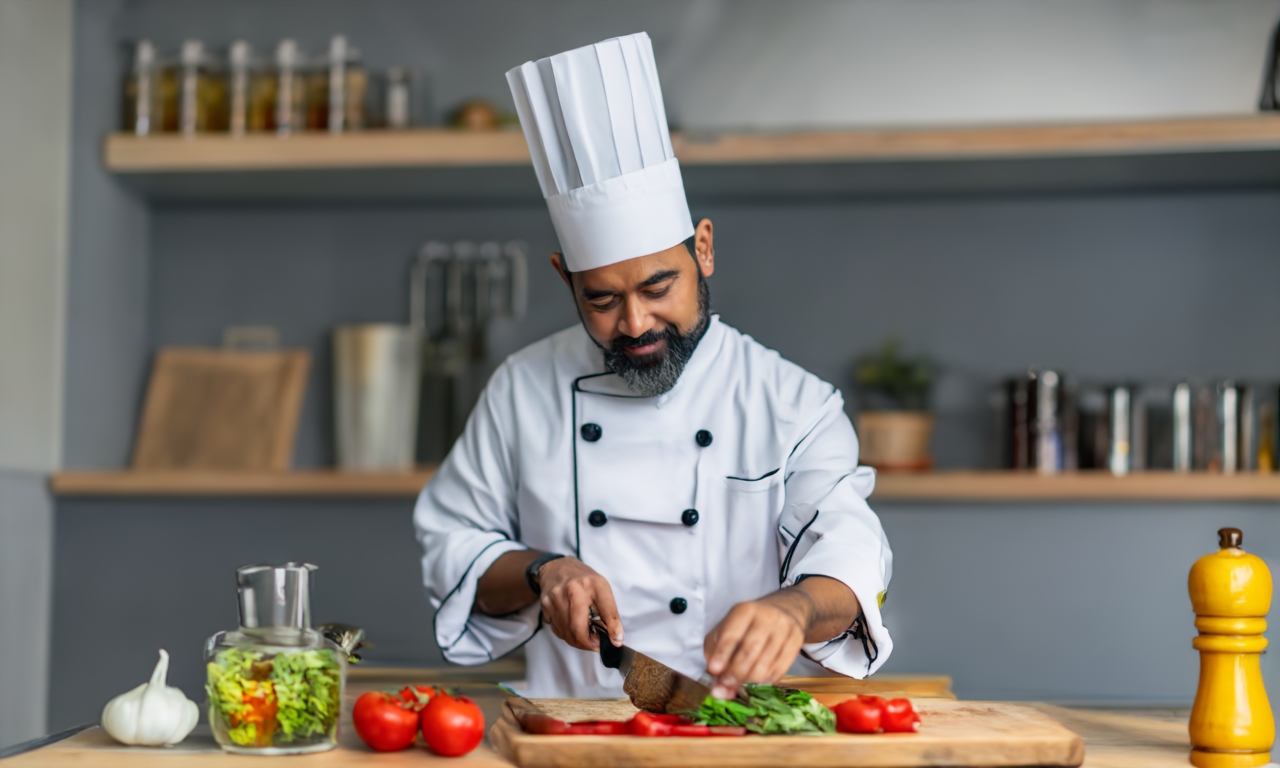} &
        \includegraphics[width=0.5\textwidth]{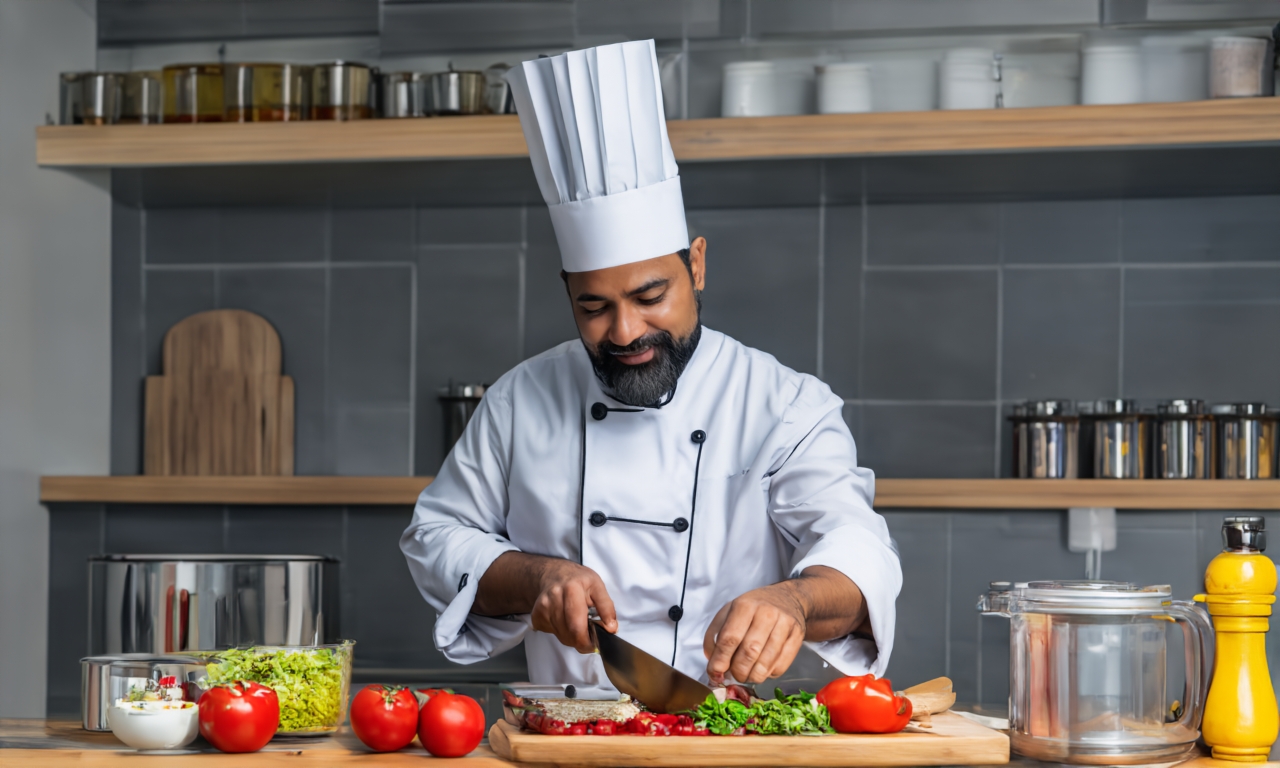} \\[2mm]
        \multicolumn{2}{c}{\emph{A photo of a chef wearing a chef hat cooking in a kitchen. The background has kitchen supplies}} \\[2mm]
        \includegraphics[width=0.5\textwidth]{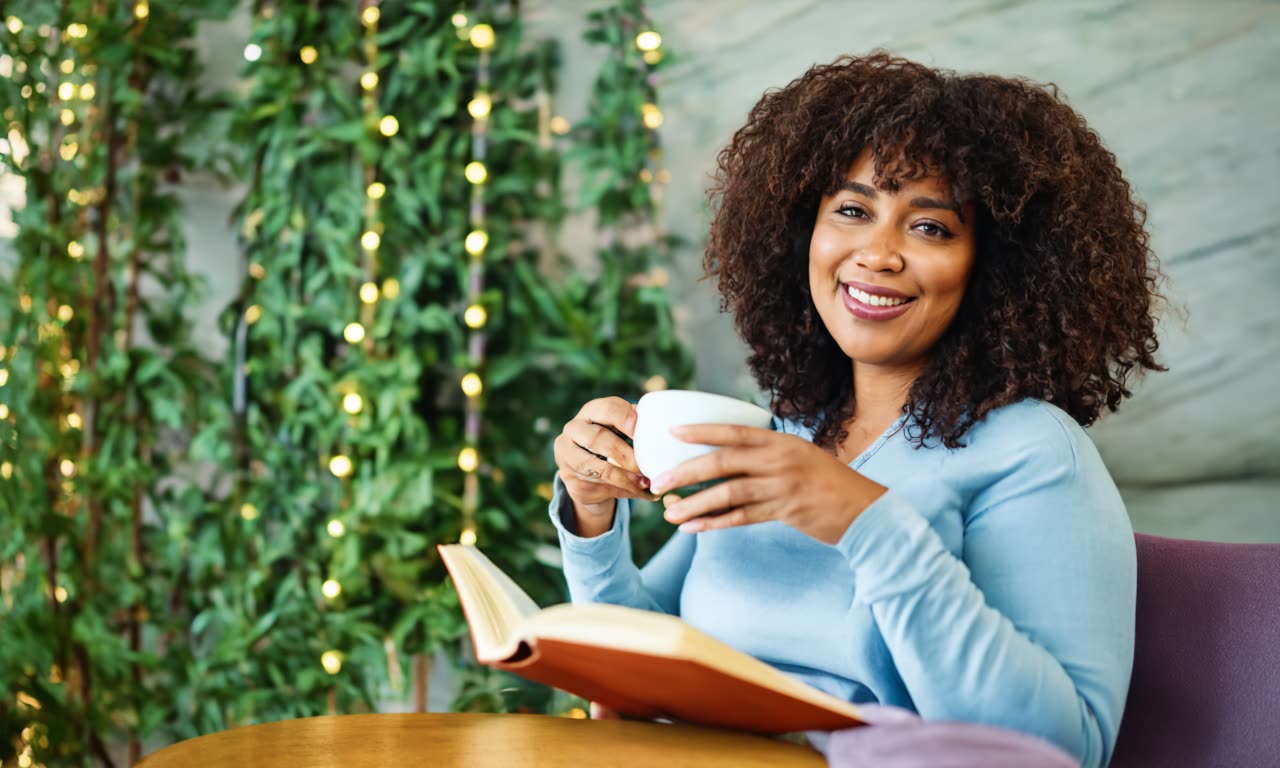} &
        \includegraphics[width=0.5\textwidth]{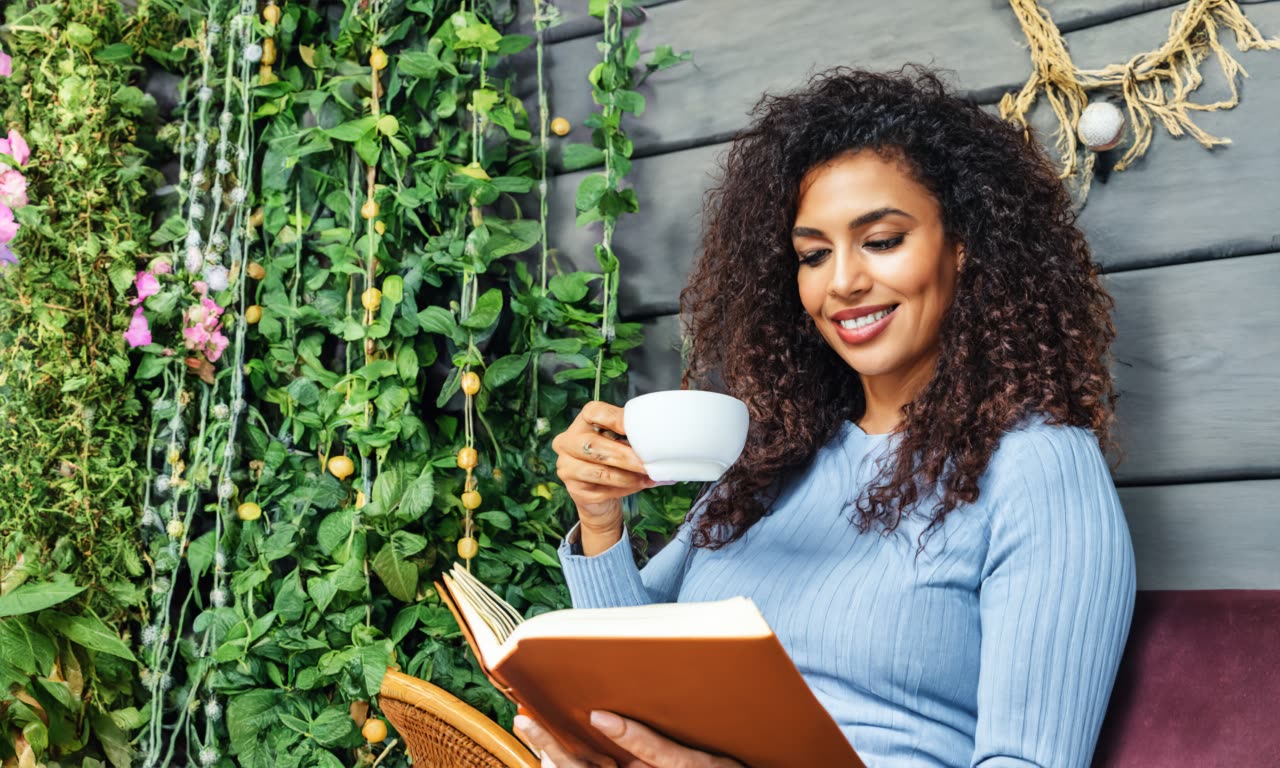} \\[2mm]
        \multicolumn{2}{c}{\emph{A photo of a woman drinking coffee in a coffee shop}}
\end{tabular}
\caption{\textbf{Camera controls - Depth of field.}}
\label{fig:camera_controls_depthoffield}
\end{figure*}

\subsection{Aspect Ratios}
We support five common aspect ratios in our image generator - $1:1$, $4:3$, $3:4$, $16:9$, and $9:16$. Samples in the dataset are first grouped into one of these five bins according to the closest aspect ratio. During each training iteration, we randomly sample a batch of examples from a bin and train the diffusion model. We provide the aspect ratio information to the model using learnable spatial positional encodings. The positional encoding parameters are defined for the base $1:1$ aspect ratio. For all other aspect ratios, we perform spatial interpolation to the required feature dimensions. We observed that the aspect ratios in our dataset had an imbalanced distribution. Despite this imbalance, the model was able to perform well across all aspect ratios.

\subsection{Training}
We train both base and upsampler models for $2.7M$ iterations. The base model was trained with a global batch size of $4096$, while the upsampler was trained with a batch size of $2048$. We use AdamW optimizer with a constant learning rate and a warmup following ~\cite{ediffI}. After 1.5M iterations, we use aesthetic weighted training, in which loss per sample is multiplied by a normalized aesthetic score computed using an aesthetic model. 

\subsection{Results}

Samples generated by our text-to-image model are shown in~\cref{fig:generations_2d_main}. Our model is able to generate highly detailed photorealistic images adhering to the input text prompt across a diverse set of categories - nature, humans, animals, food, etc. We also show results across three aspect ratios - $16:9$, $1:1$, and $9:16$. The model can generate high-quality images in both aspect ratios despite having very few $9:16$ images in the dataset. Our model can also generate images adhering to long and descriptive captions, as shown in~\cref{fig:generations_long_prompts}. 

\subsubsection{Fairness and Diversity}

In~\cref{fig:race_diversity}, we show the results of our text-to-image model on some human generation prompts. We observe that our model can generate images with sufficient race and gender diversity.

\subsubsection{Camera Control}
As discussed in Section~\ref{sec:conditioning_inputs},  we condition our diffusion models on pitch and depth of field attributes during training.~\cref{fig:camera_controls_pitch} shows the generations as we vary the pitch to \textit{ascending}, \textit{eye level}, and \textit{descending view} while using the same text prompt. We observe that the pitch of the resulting image changes as specified in the input. In~\cref{fig:camera_controls_depthoffield}, we vary the depth of field attribute to shallow and deep. The images with shallow depth of field have blurred backgrounds, while those with deep depth of field have all regions in focus.

\section{$4K$ Upsampling}\label{sec:4K_upsampling}

Our Edify suite of image generation models also includes $4K$ upsampling, which helps users generate highly detailed images. While the $1K$ generator generates high-quality images with strong adherence to the input text prompts, the $4K$ upsampler adds additional fine-grained details to the $1K$ resolution image and outputs $4K$ resolution images.

\begin{figure}[htbp]
    \centering
    \begin{subfigure}{\textwidth}
        \centering
        \includegraphics[trim={0 5cm 0 20cm},clip,width=\textwidth]{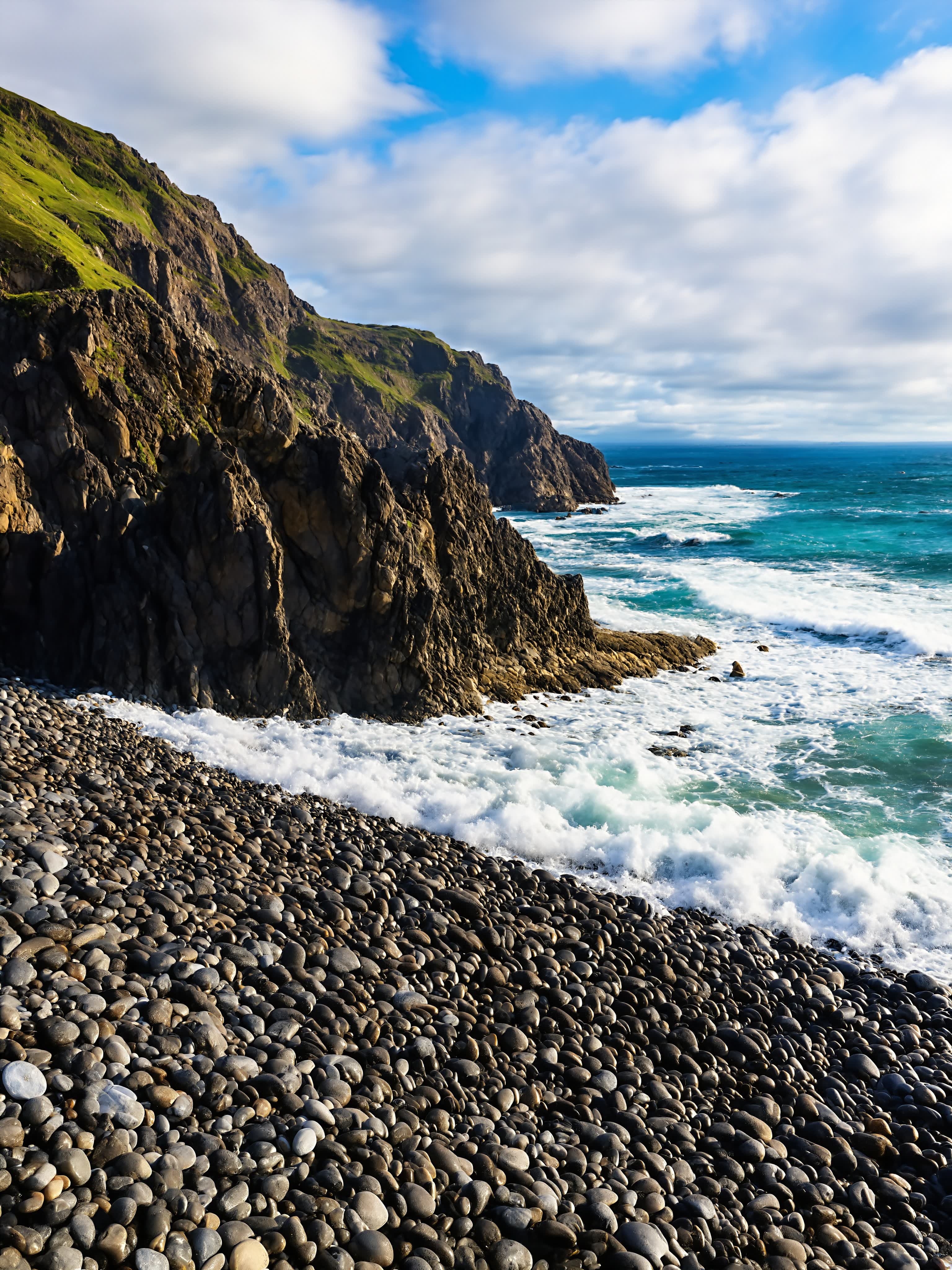}
    \end{subfigure}
    
    \vspace{0.25em}
    
    \begin{subfigure}{0.496\textwidth}
        \centering
        
        \begin{tikzpicture}
            \node[anchor=south west,inner sep=0] (image) at (0,0) {\includegraphics[width=\textwidth]{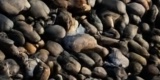}};
            \begin{scope}[x={(image.south east)},y={(image.north west)}]
                \node[text=white, font=\bfseries] at (0.5,0.9) {1K Resolution};
            \end{scope}
        \end{tikzpicture}
    \end{subfigure}
    \hfill
    \begin{subfigure}{0.496\textwidth}
        \centering
        
        \begin{tikzpicture}
            \node[anchor=south west,inner sep=0] (image) at (0,0) {\includegraphics[width=\textwidth]{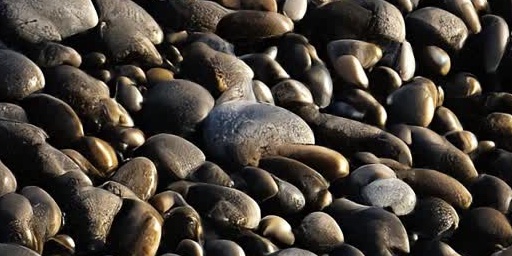}};
            \begin{scope}[x={(image.south east)},y={(image.north west)}]
                \node[text=white, font=\bfseries] at (0.5,0.9) {4K Resolution};
            \end{scope}
        \end{tikzpicture}
    \end{subfigure}
    
    \caption{\textbf{$4K$ Upsampling results.} Full (top) and zoomed-in images (bottom) show the additional details.}
    \label{fig:upsampling_1}
\end{figure}

\begin{figure}[htbp]
    \centering
    \begin{subfigure}{\textwidth}
        \centering
        \includegraphics[trim={0 5cm 0 20cm},clip,width=\textwidth]{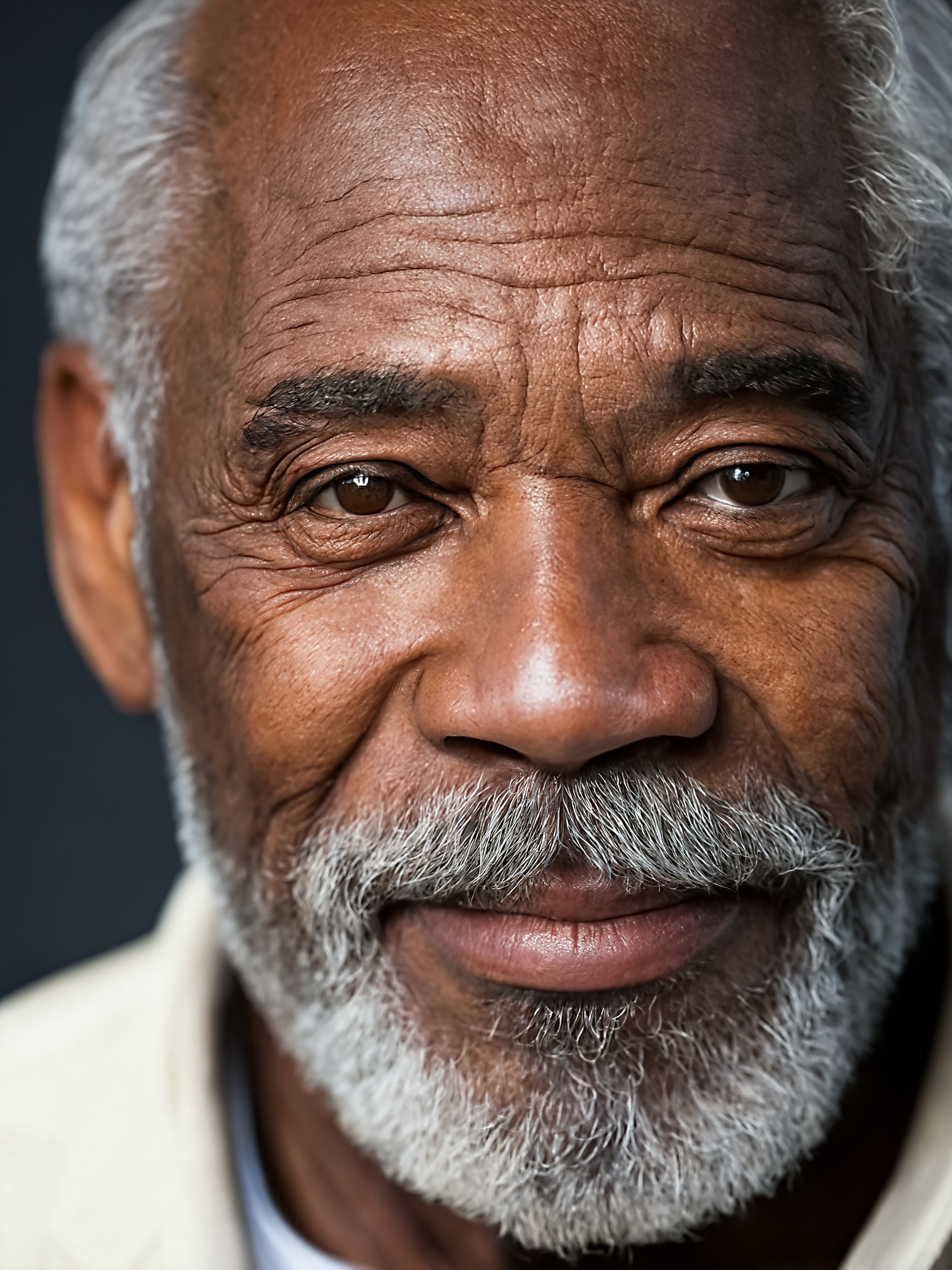}
    \end{subfigure}
    
    \vspace{0.5em}
    
    \begin{subfigure}{0.495\textwidth}
        \centering
        
        \begin{tikzpicture}
            \node[anchor=south west,inner sep=0] (image) at (0,0) {\includegraphics[width=\textwidth]{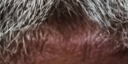}};
            \begin{scope}[x={(image.south east)},y={(image.north west)}]
                \node[text=white, font=\bfseries] at (0.5,0.9) {1K Resolution};
            \end{scope}
        \end{tikzpicture}
    \end{subfigure}
    \hfill
    \begin{subfigure}{0.495\textwidth}
        \centering
        
        \begin{tikzpicture}
            \node[anchor=south west,inner sep=0] (image) at (0,0) {\includegraphics[width=\textwidth]{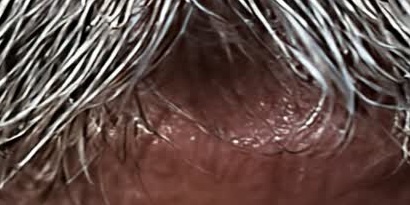}};
            \begin{scope}[x={(image.south east)},y={(image.north west)}]
                \node[text=white, font=\bfseries] at (0.5,0.9) {4K Resolution};
            \end{scope}
        \end{tikzpicture}
    \end{subfigure}
    
    \caption{\textbf{$4K$ Upsampling results.} Full (top) and zoomed-in images (bottom) show the additional details.}
    \label{fig:upsampling_2}
\end{figure}

\subsection{Approach}

In theory, given our model formulation, it is very easy to train a $4K$ resolution diffusion model by simply adding a new resolution level while training. However, there is typically a large gap in the amount of data available at higher resolutions compared to lower-resolution data. This is indeed the case for our dataset as well, wherein the number of good-quality images with $4K$ or higher resolution is less than 1\% of the data available to train the $4K$ model. Here, we refer to `good-quality' images as those images that pass our criteria for containing high-frequency content and are aesthetically pleasing, which is required to train a good-quality upsampling model. To address this, we opt to utilize the existing $1K$ generator as the base for the upsampling model.

By scaling the noise levels appropriately, we can generate a high-quality $4K$ image directly from a pre-trained $1K$ generator. In the case of upsampling, similar to SDEdit, we can start with a low-resolution image, resize it to the desired resolution, add noise to it based on the forward diffusion process discussed in~\cref{sec:dim_varying_edm}, and finally denoise it iteratively using our base $1K$ model to obtain the upsampled image. One issue with this approach, however, is that the model may change the content in the initial low-res image to a degree that may not be desirable to the user. To overcome this challenge, we design the upsampler as a ControlNet which conditions the base model on the clean low-resolution input image. Finally, we fine-tune the base model with the low-resolution ControlNet on a smaller number of $4K$ images available to us. This helps the model in two ways:
\begin{itemize}
    \item The pre-trained base model has not seen any high-frequency content which is crucial for generating $4K$ images. Fine-tuning on the $4K$ data enables the model to generate such details.
    \item The clean low-resolution image conditioning allows the model to access the original content of the noisy input image and prevents it from deviating too much from the original.
\end{itemize}

Additionally, we utilize reconstruction guidance~\cite{ho2022video} during sampling to further control the degree of change to the original low-resolution image.

\subsection{Results}

\Cref{fig:upsampling_1} and \cref{fig:upsampling_2} show how the upsampling model is able to add more details to the $1K$ resolution output from the base model. The difference is clearly observed by zooming into a region in the image showing the high-frequency content. Note that images in~\Cref{fig:upsampling_1} and \cref{fig:upsampling_2} are all upsampled to $4K$ using our upsampler.

\section{Generation with Additional Control}
\begin{figure*}[t]
    \centering
    \includegraphics[width=\textwidth]{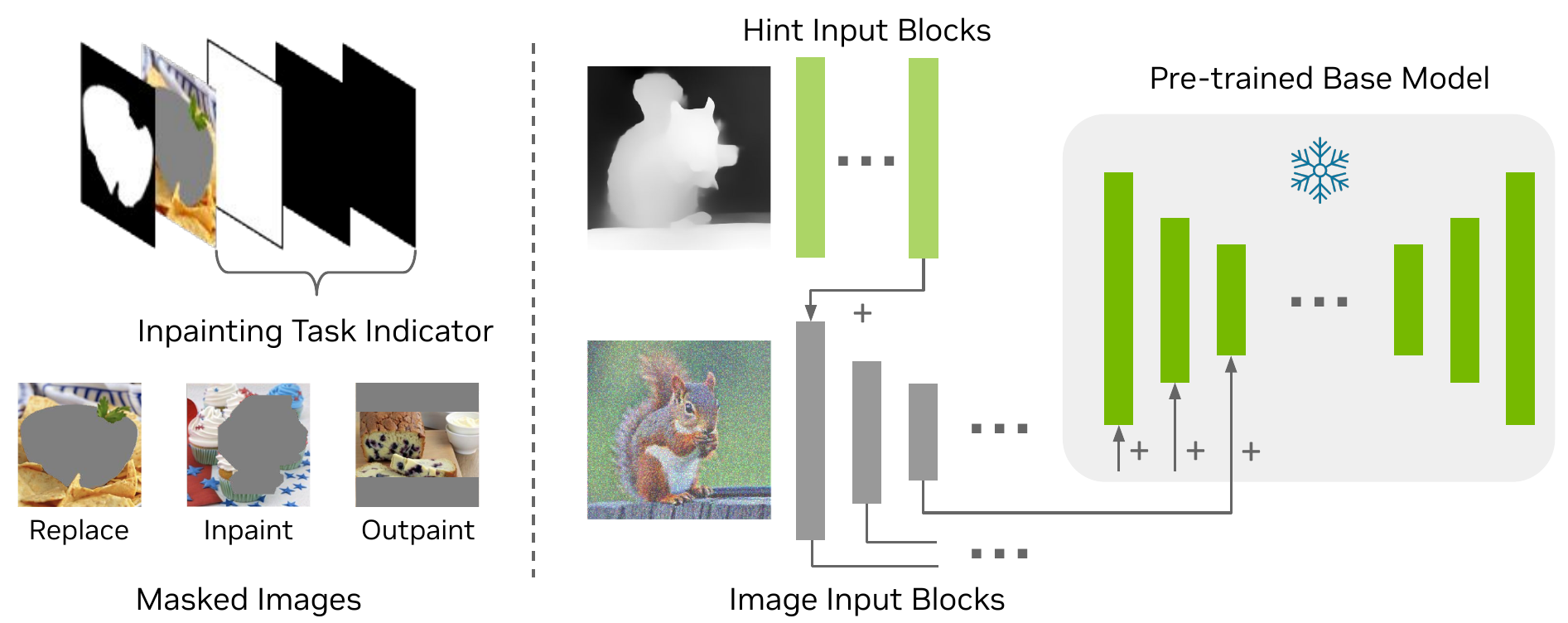}
    \caption{\textbf{Model architecture with additional control inputs.} The base model is frozen when training the ControlNet encoders. The Image Input Blocks are initialized from the base model U-Net. The Hint Input Blocks are randomly initialized. }
    \label{figcontrolnet}
\end{figure*}

We add additional control to the Edify Image model by training ControlNet encoders following \cite{zhang2023adding}. 
\cref{figcontrolnet} illustrates the architecture of the Edify Image model with ControlNet, which will be detailed in the following sections. 

\setlength{\tabcolsep}{1pt}
\renewcommand{\arraystretch}{0.5}
\begin{figure*}[tb!]
    \centering
\includegraphics[width=0.95\textwidth]{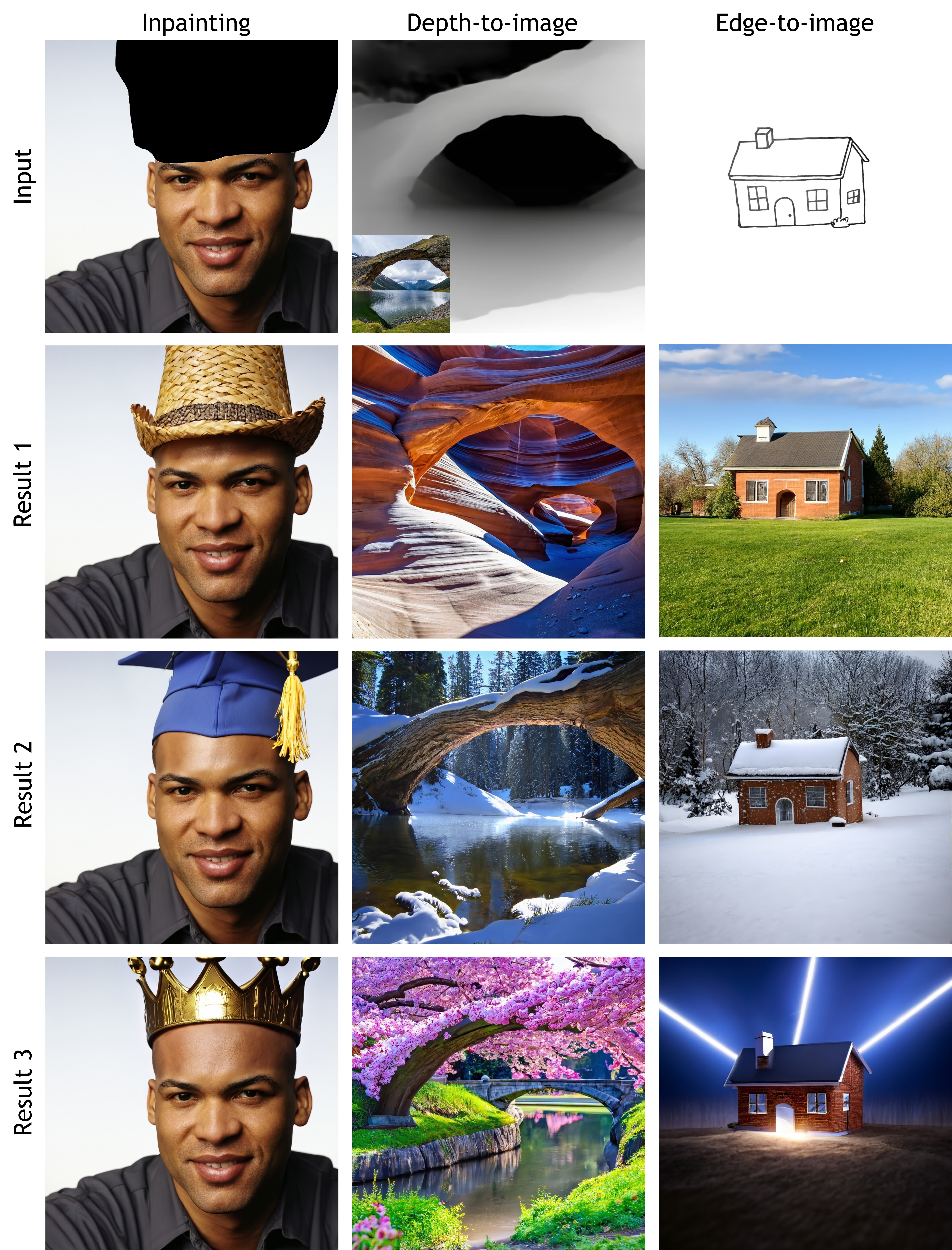}
\caption{\textbf{Results with additional control inputs for inpainting, depth, and edge.} For each input condition, we generate 3 variants using different text prompts.}
\label{fig:controlnetresult}
\end{figure*}
\subsection{Approach}

After the base model is pre-trained, we freeze the model parameters and introduce an additional encoder whose parameters are partially initialized from the first half of the base model's U-Net. As the control input, such as depth and sketch maps, may have different dimensions from images, we add several extra blocks, called Hint Input Blocks, to transform the control input into feature maps that will be added to the features from the noisy image input. Additionally, by scaling the control input feature maps (\ie, control weight), we can achieve the controllability of different strengths. 

We view inpainting as another controlled image generation problem, similar to sketch and depth-controlled generation, with the partial image and inpainting mask as the control input. We consider three sub-tasks for inpainting: 
\begin{itemize}
\item Replace: the unknown area in the image is an entire semantic area, which means the mask shape strictly follows the object shape. This is useful for replacing objects or backgrounds while we do not want to change the object shape. 
\item Inpaint: the unknown area is not a semantic area and could partially cover both the background and foreground. 
\item Outpaint: the unknown area is at the image boundary. This is usually called outpainting in the literature but can be viewed as a special case of inpainting. 
\end{itemize}
The left side of \cref{figcontrolnet} showcases the examples of the three sub-tasks. Only one shared inpainting model is trained for all sub-tasks. We use a one-hot vector to indicate different tasks, which is expanded to the image size and concatenated with the masked image and inpainting mask to serve as the control input. 

\setlength{\tabcolsep}{1pt}
\renewcommand{\arraystretch}{0.5}
\begin{figure*}[tb!]
    \centering
    \includegraphics[width=\textwidth]{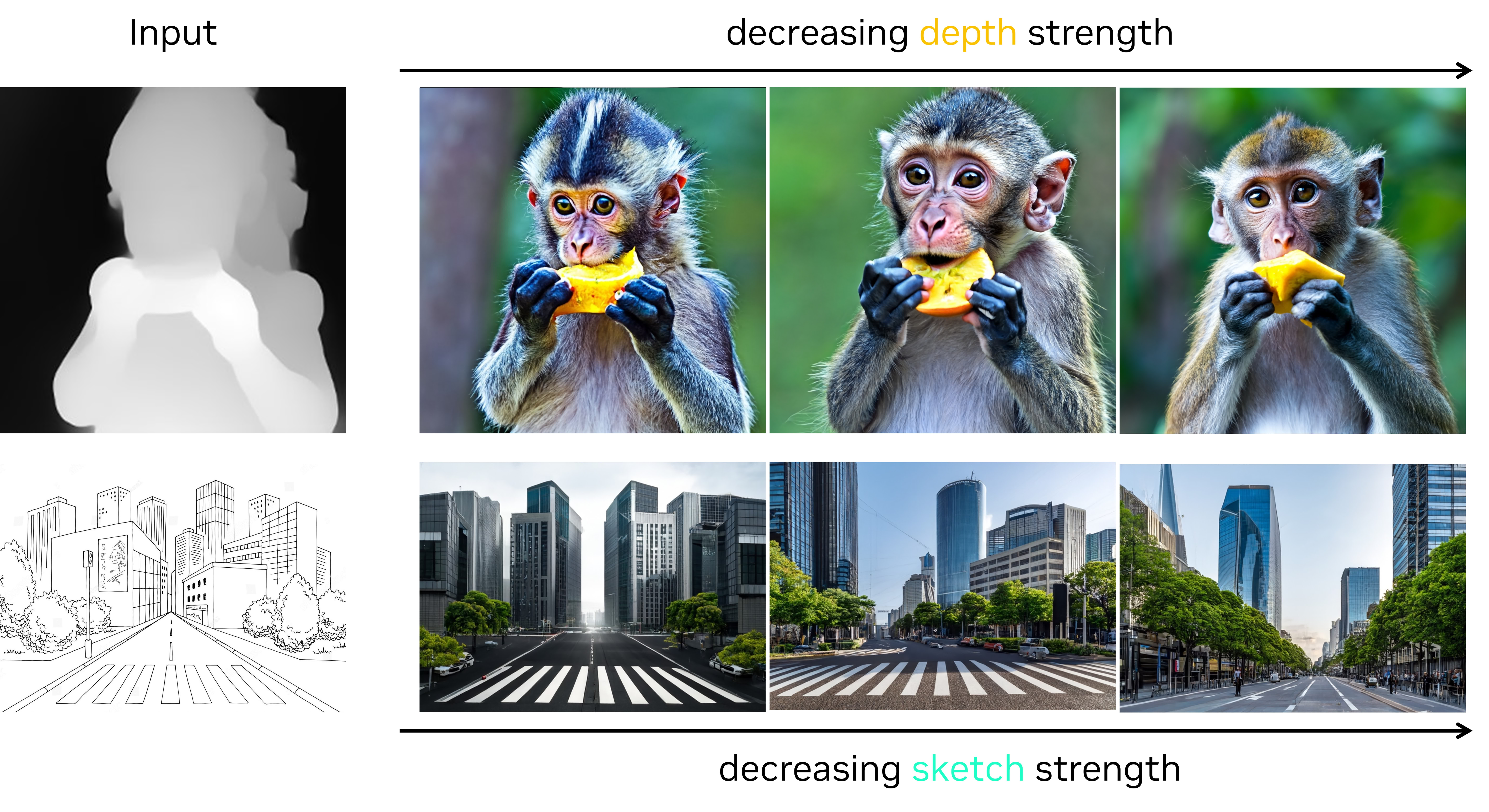} 
\caption{\textbf{Results with different control weight values for depth-to-image and edge-to-image.} }
\label{fig:controlnetweight}
\end{figure*}

\subsection{Training}
We compute Canny edges, HED edges, and depth maps from input RGB images and use them to train the edge/depth2image models. For inpainting, we generate random masks or use object masks to train the inpainting model. Following~\cite{zhang2023adding}, we only train the additional encoder and keep the base model frozen during training.

\subsection{Results}
\cref{fig:controlnetresult} shows example results with various control inputs. The model can generate high-quality images while following the image structure indicated by the control input. 
Furthermore, we demonstrate the effect of different control weights in \cref{fig:controlnetweight}, using edge and depth inputs as an example. The generated image can be aligned to the input more strictly with a higher control strength.

\section{360$^{\circ}$ HDR Panorama Generation}\label{sec:360_generation}



\begin{figure}[htbp]
\centering
\begin{subfigure}{\linewidth}
\includegraphics[width=\textwidth]{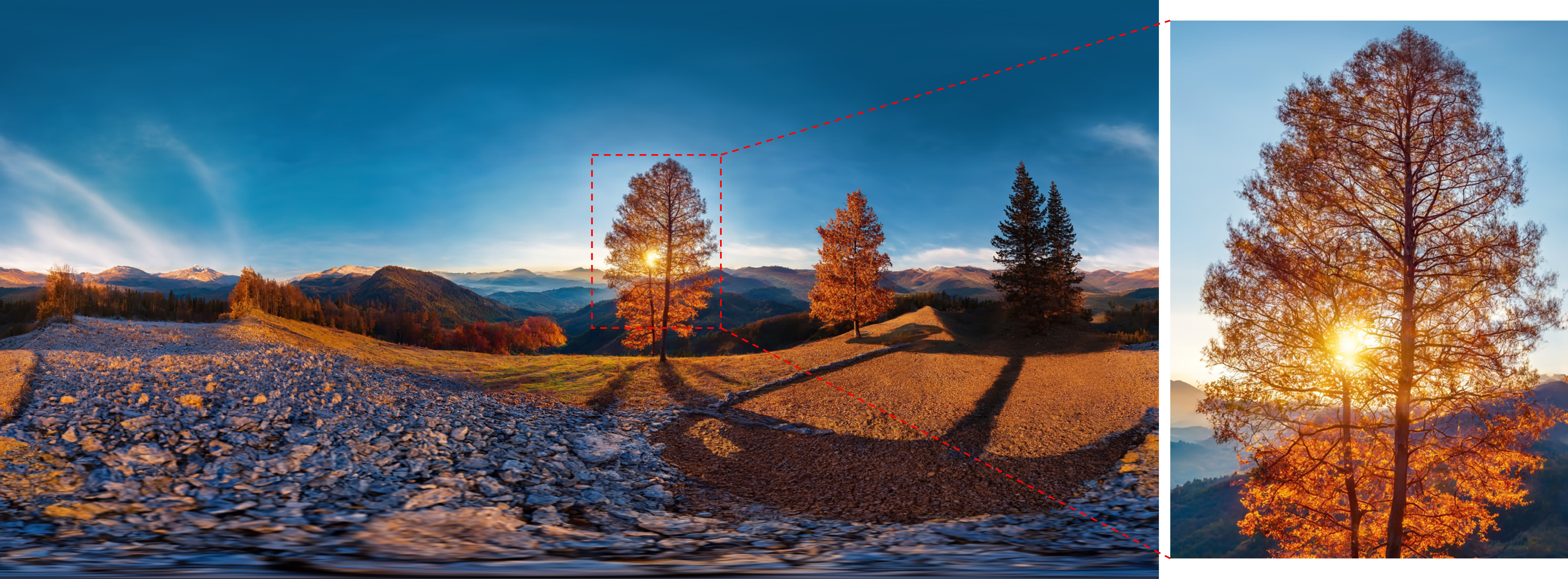}
\caption{\emph{sunset at a lookout point in a gravel parking lot with blue sky and a few autumn maple trees and beautiful smokey mountains in the background, scenic nature, inspiring, landscape panoramic, mountains.}}
\end{subfigure}
\\
\begin{subfigure}{\linewidth}
\includegraphics[width=\textwidth]{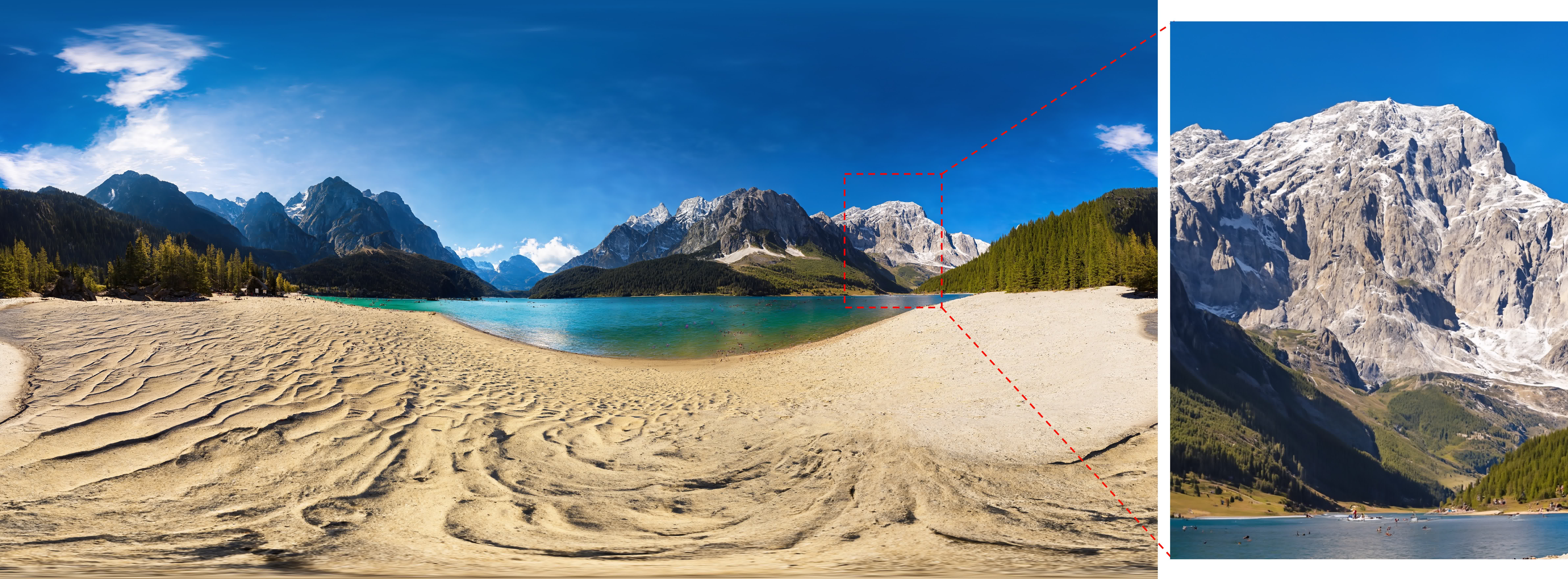}
\caption{\emph{flat sand beach by a lake in the swiss alps mountains at noon with beautiful swiss alps mountains in the background, god rays, scenic nature, inspiring, landscape panoramic.}}
\end{subfigure}
\\
\begin{subfigure}{\linewidth}
\includegraphics[width=\textwidth]{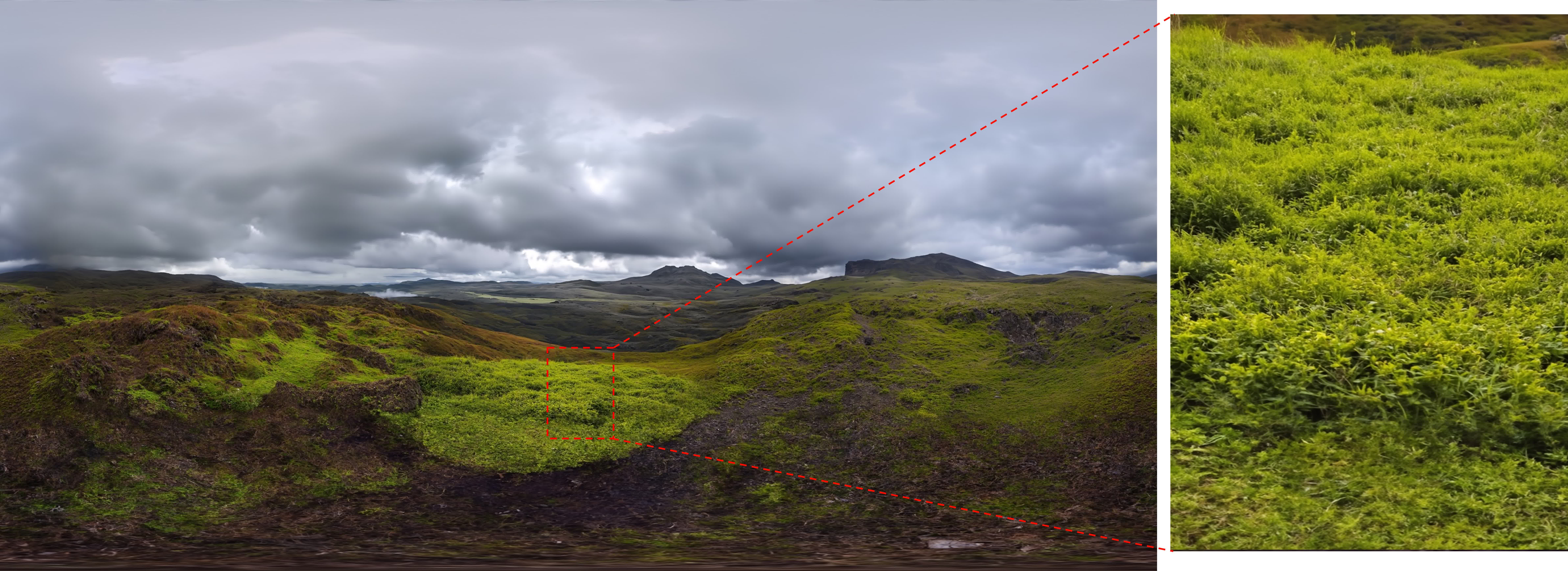}
\caption{\emph{moss and grass plains in scottish highlands, scotland, remote, photography, wilderness, moody cloudy sky, rain, bluffs in background.}}
\end{subfigure}
\caption{\textbf{Example 360 panorama generation results.} The input prompts are described under each image. We also show zoomed-in crops at the right to better show the details in the images.}
\label{fig:pano360}
\end{figure}


\begin{figure}[htb]
\centering
\begin{subfigure}{0.25\linewidth}
    \includegraphics[width=.99\textwidth]{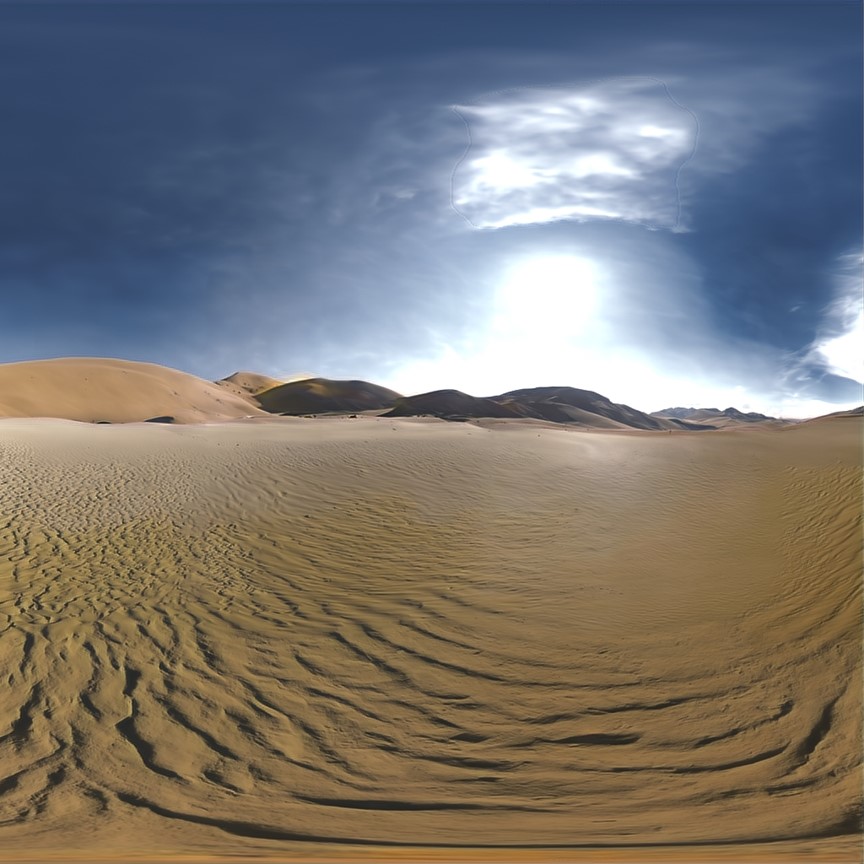}
    \subcaption{\centering ev+0}
\end{subfigure}%
\begin{subfigure}{0.25\linewidth}
    \includegraphics[width=.99\textwidth]{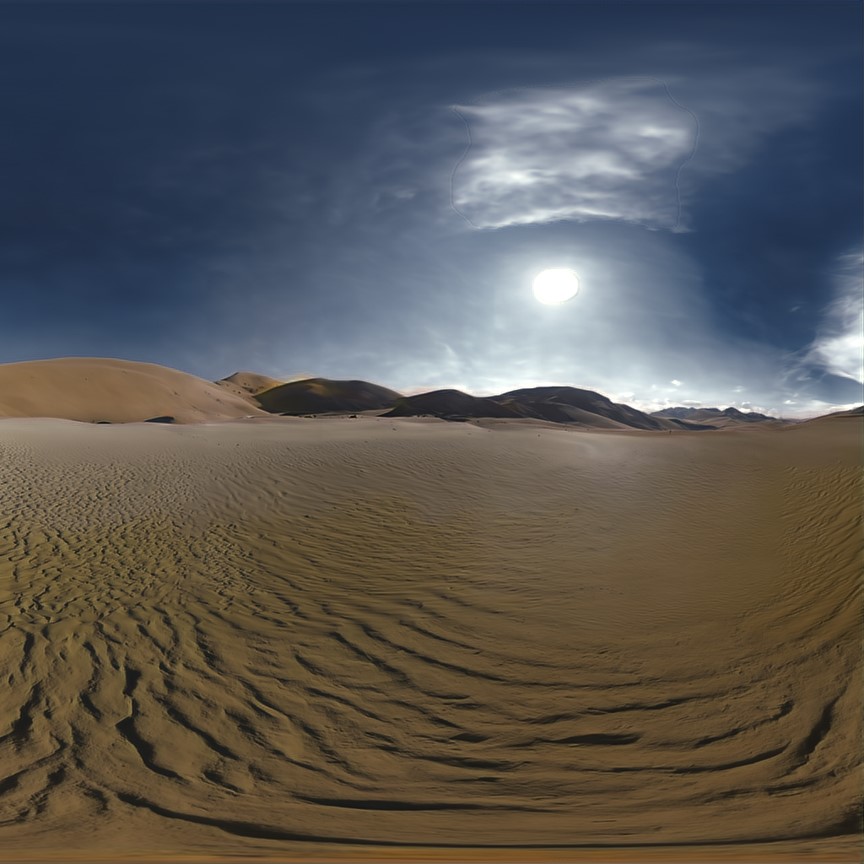}
    \subcaption{\centering ev-2}
\end{subfigure}%
\begin{subfigure}{0.25\linewidth}
    \includegraphics[width=.99\textwidth]{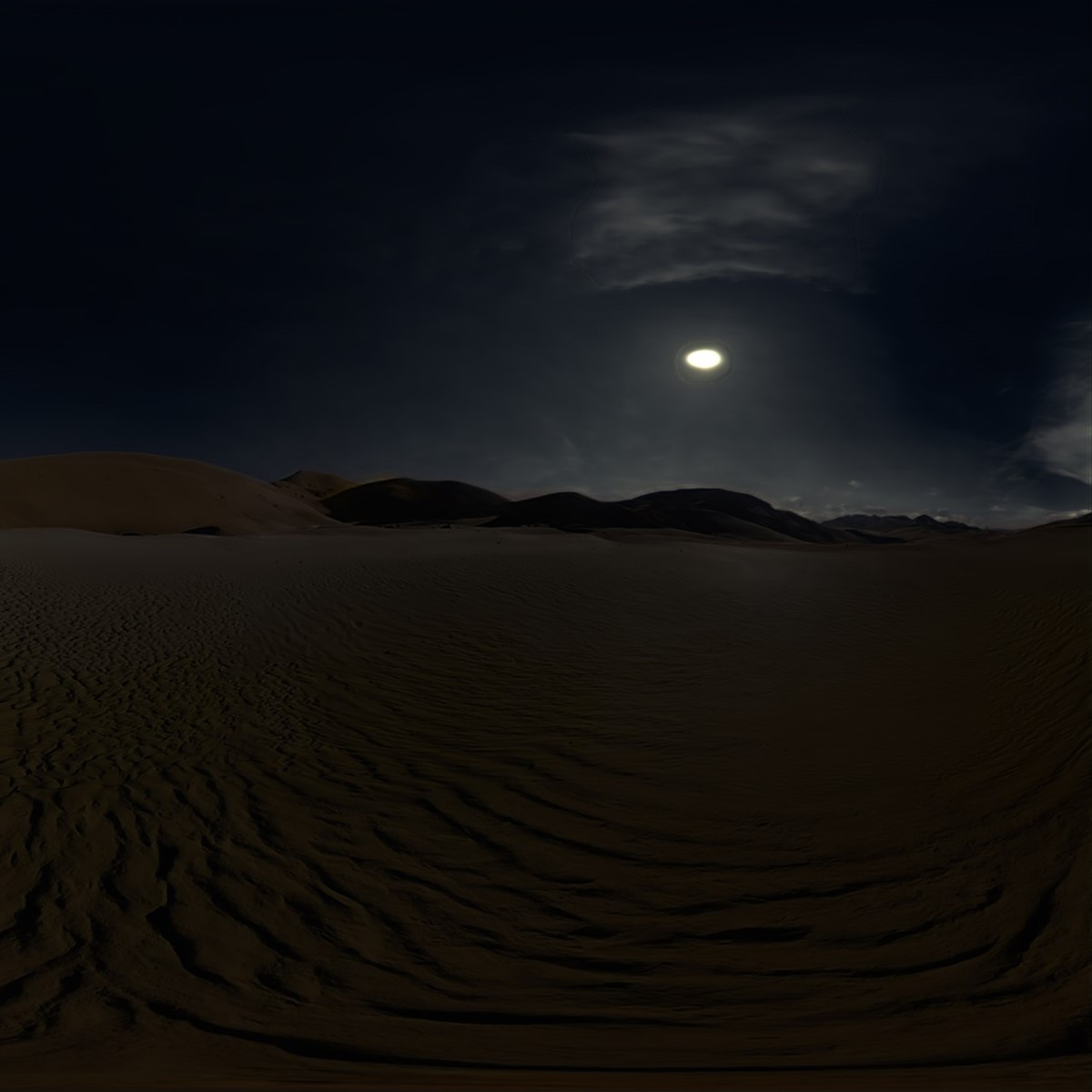}
    \subcaption{\centering ev-5}
\end{subfigure}%
\begin{subfigure}{0.25\linewidth}
    \includegraphics[width=.99\textwidth]{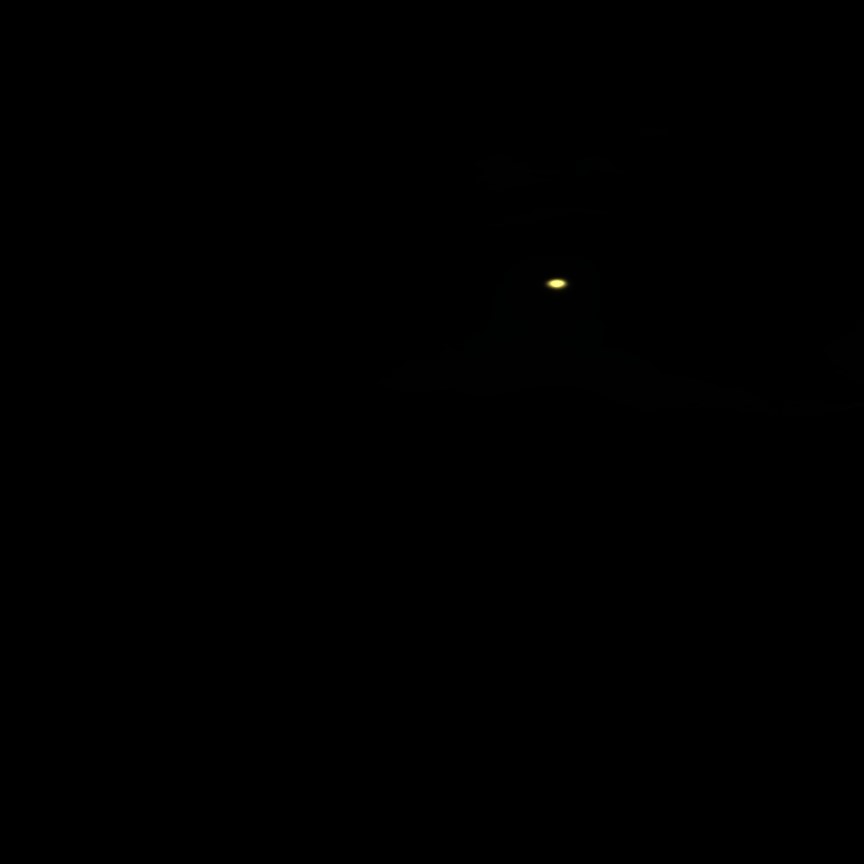}
    \subcaption{\centering ev-9}
\end{subfigure}
\caption{\textbf{Crops of an HDR panorama at different exposure stops.}  This panorama image has 19 stops of dynamic range, showing the sun and bright clouds with values ``above white''.}
\label{fig:360hdr}
\end{figure}

Building on the foundation described in previous sections, we developed a high-dynamic range (HDR) 360-degree panorama generator. Given a text prompt and (optionally) a corresponding example image from a single viewpoint, the system generates omnidirectional equirectangular projection panoramas at $4K$, $8K$, or $16K$ resolution (\cref{fig:pano360}).
The generated outputs can provide content for 3D virtual reality headsets and backdrops for movies and games. Thanks to the high-dynamic range output, it can be used as image-based lighting (IBL).

\subsection{Approach}

Unlike the case of images, which are cheap to obtain and available at scale on the Internet, gathering HDR panoramas is time-consuming. A single panorama requires capturing and combining multiple images across different directions and exposure levels. The amount of available HDR panorama data is orders of magnitude less than that used to train successful foundation image models. To address this data limitation, our algorithm relies on the base Edify Image model to provide a general text-to-image capability and assemble multiple generated images into the desired panorama. The limited panorama data are used only to fine-tune this process and for HDR estimation.

The algorithm adopts a sequential inpainting approach in which a number of conventional perspective images are synthesized with the foundation model and stitched together, with overlap from preceding images, to ensure continuity. 
During synthesis, each image is warped into equirectangular coordinates and projected into the coordinates of the neighboring image to provide the overlap region. 
The zenith (sky) and nadir (ground) images are inpainted with overlaps from all longitudinal images.  
The inpainting is trained as a controlnet, with an image containing the overlap area providing the control signal.

After we generate the panorama, we feed it to an LDR2HDR network to convert the low dynamic range (LDR) image to a high dynamic range (HDR) image. The LDR2HDR network is a multi-scale U-Net where we first generate a low-resolution HDR image and then concatenate it with the high-resolution LDR input to generate the high-resolution HDR output. To train the network, we convert the ground truth HDR dataset into LDR images and ask the network to reconstruct the original HDR input. For better training stability, we train the network to predict intensity values in logarithmic space.

\subsection{Results}
\cref{fig:pano360} shows the results for our panorama synthesis at 16k resolution. Note that we are able to generate consistent panoramic scenes that properly follow the input prompt. We are also able to synthesize fine details for the trees, grass, etc, which are essential to make the results look realistic.

On the other hand, \cref{fig:360hdr} demonstrates the result of our HDR generation from LDR input, which correctly assigns high-intensity values to bright objects like the sun and clouds. It also predicts a wide dynamic range (\eg,\ 19 stops) of intensities (crucial for IBL applications).

\subsection{Limitations}
The panorama generator application shares assumptions and limitations of 360$^{\circ}$ panoramas in general; specifically, the panorama shows views in any direction from a single location. 
As a consequence, parallax is not possible, and translating the viewpoint results in visible distortions unless the translation is very small.

Another limitation is that while the generated panels are pairwise consistent, there is not necessarily any global consistency to the lighting. We hope to address this issue in future work. Note that this is not a crippling limitation in either backdrop or IBL applications: In the case where the panorama is visible as a backdrop, every individual view is plausible, and viewers may not notice the issue without additional study. In image-based lighting applications, when the sun is visible, it alone is usually responsible for most of the lighting effect due to its vastly greater intensity relative to reflected light.

\section{Finetuning for Customization}\label{sec:finetuning}

We explore the Edify Image model's capability to adapt to new personalization and stylization tasks. First, we describe our approach and then showcase several use cases, including single and multi-subject personalization, as well as single and multi-subject stylization. Finally, we demonstrate how the finetuned model can seamlessly integrate with various pre-trained frozen Edify ControlNets.

\subsection{Approach}

Our finetuning approach does not modify the architecture of the Edify Image models, and we keep the text encoders frozen. We finetune only a subset of parameters in the cross-attention layers of the U-Nets, which accounts for just 3\% of the total U-Net parameters. We finetune both the $256$ and $1024$-resolution U-Nets for 1500 steps.

\subsection{Results}

We finetuned our models on four different datasets, each demonstrating the model's ability to handle various customization tasks: single-subject personalization, multi-subject personalization, single-subject stylization, and multi-subject stylization. All the images in this section are upsampled by our $4K$ model in \cref{sec:4K_upsampling}.

\begin{figure}[htbp]
    \centering
    \begin{tabular}{cccc}
    \multicolumn{2}{c}{Age Variation} & 
    \multicolumn{2}{c}{Scenario Variation} \\ [2mm]
    \includegraphics[width=0.25\textwidth]{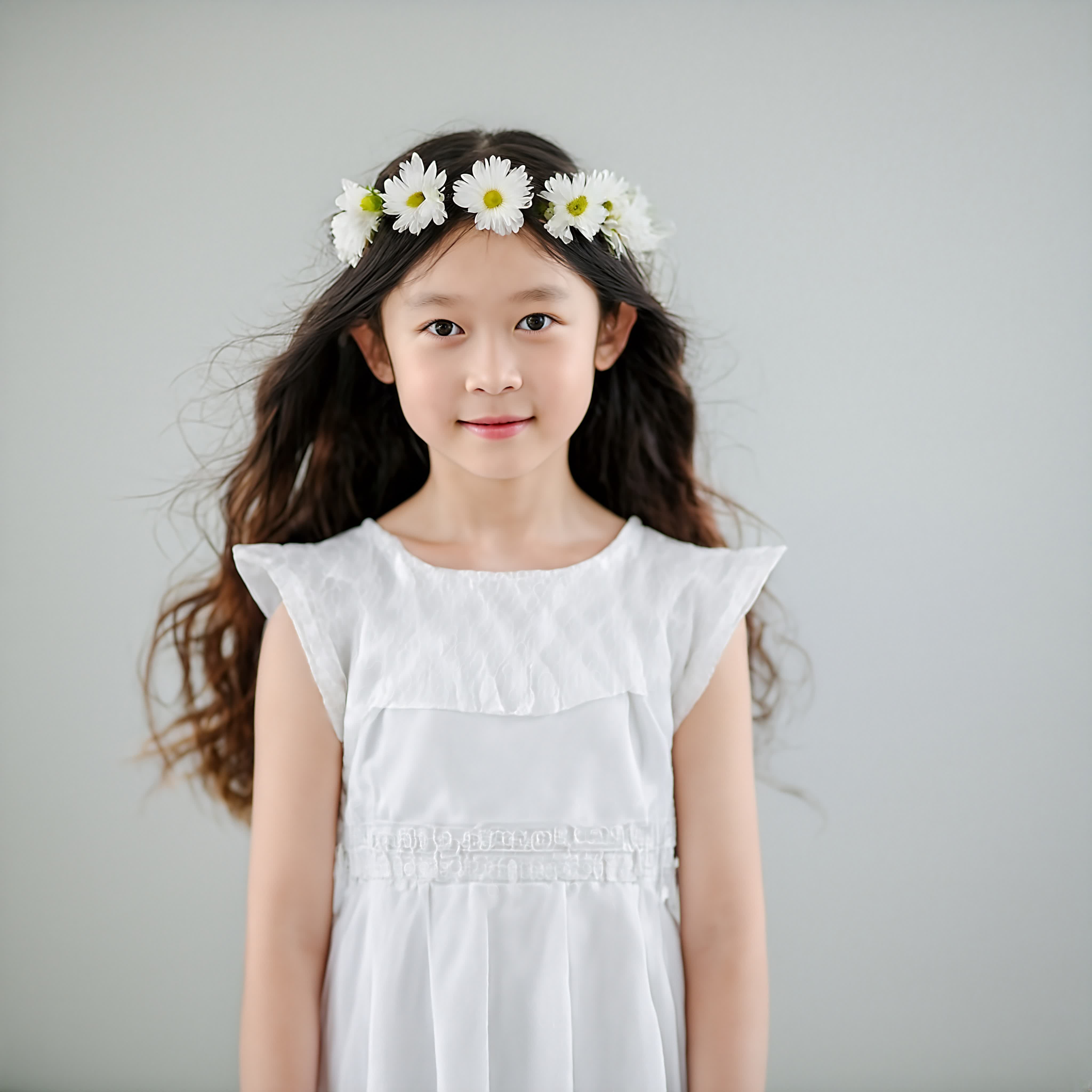} &  
    \includegraphics[width=0.25\textwidth]{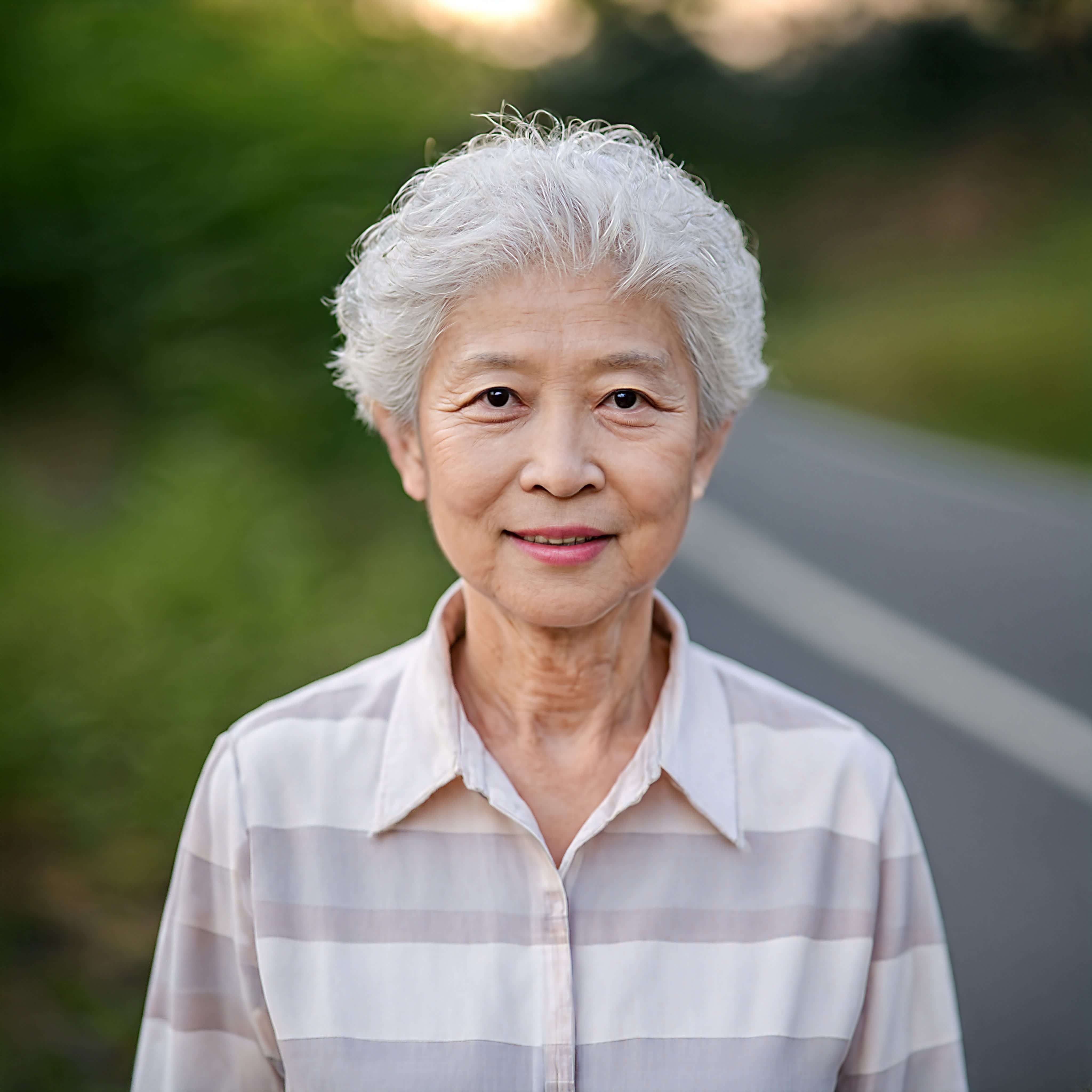} &
    \includegraphics[width=0.25\textwidth]{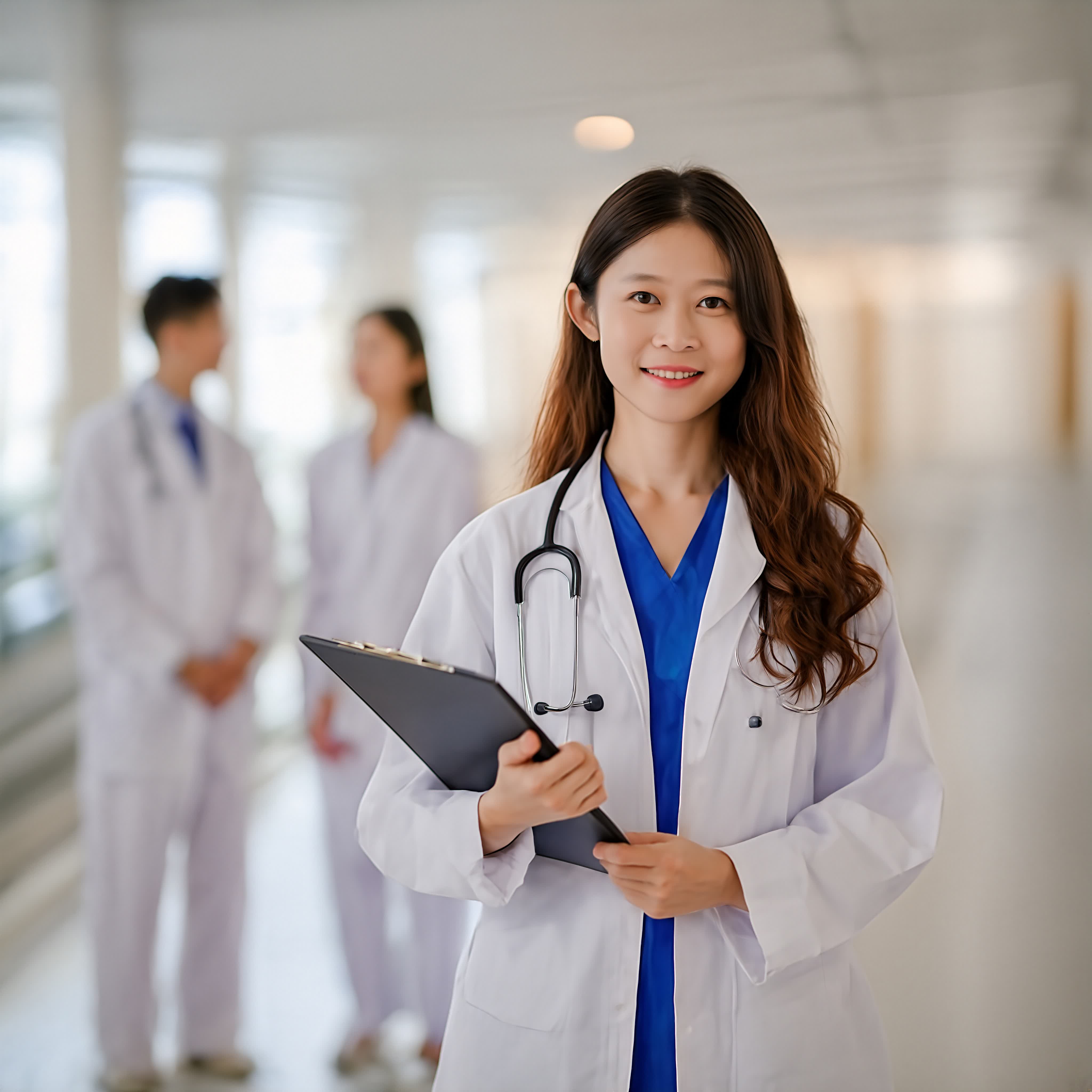} &
        \includegraphics[width=0.25\textwidth]{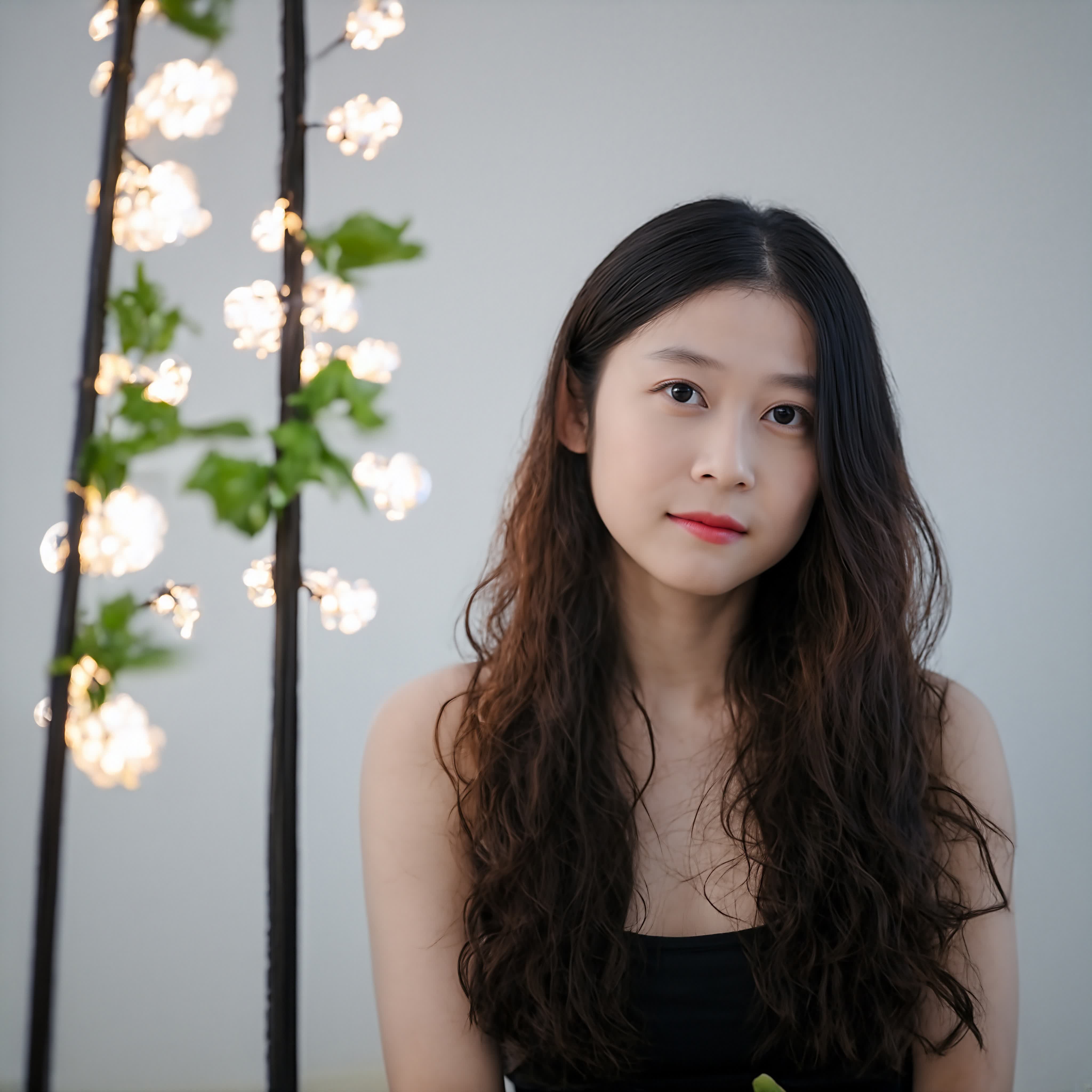}\\ 
    \end{tabular}
    \caption{
The finetuned model is capable of generating realistic images of her at different ages and in a variety of scenarios that were not included in the finetuning dataset 
    (\cref{fig:fangyin_train}).}
    \label{fig:fangyin}
\end{figure}

\paragraph{Learning a single human} We finetuned the Edify Image model using the finetuning data shown in \cref{fig:fangyin_train}. 
\Cref{fig:fangyin} demonstrates the model's capability to generate images of the person at different ages and in various outfits, none of which were included in the training data.

\paragraph{Learning multiple humans} We also finetuned the model on a dataset (\cref{fig:ss_train}) that included multiple subjects. Interestingly, this dataset contains only three images with two individuals together, while the remaining 96\% of the training images feature a single person. Despite the limited instances of multiple subjects, the finetuned model accurately generates images depicting both individuals engaging in various activities, as shown in \cref{fig:sid_sj}. To distinguish between multiple subjects, distinct names were used for each individual in the training prompts. \\

\begin{tabularx}{\textwidth}{M{0.33\textwidth} M{0.33\textwidth} M{0.33\textwidth}}
    \includegraphics[width=0.33\textwidth]{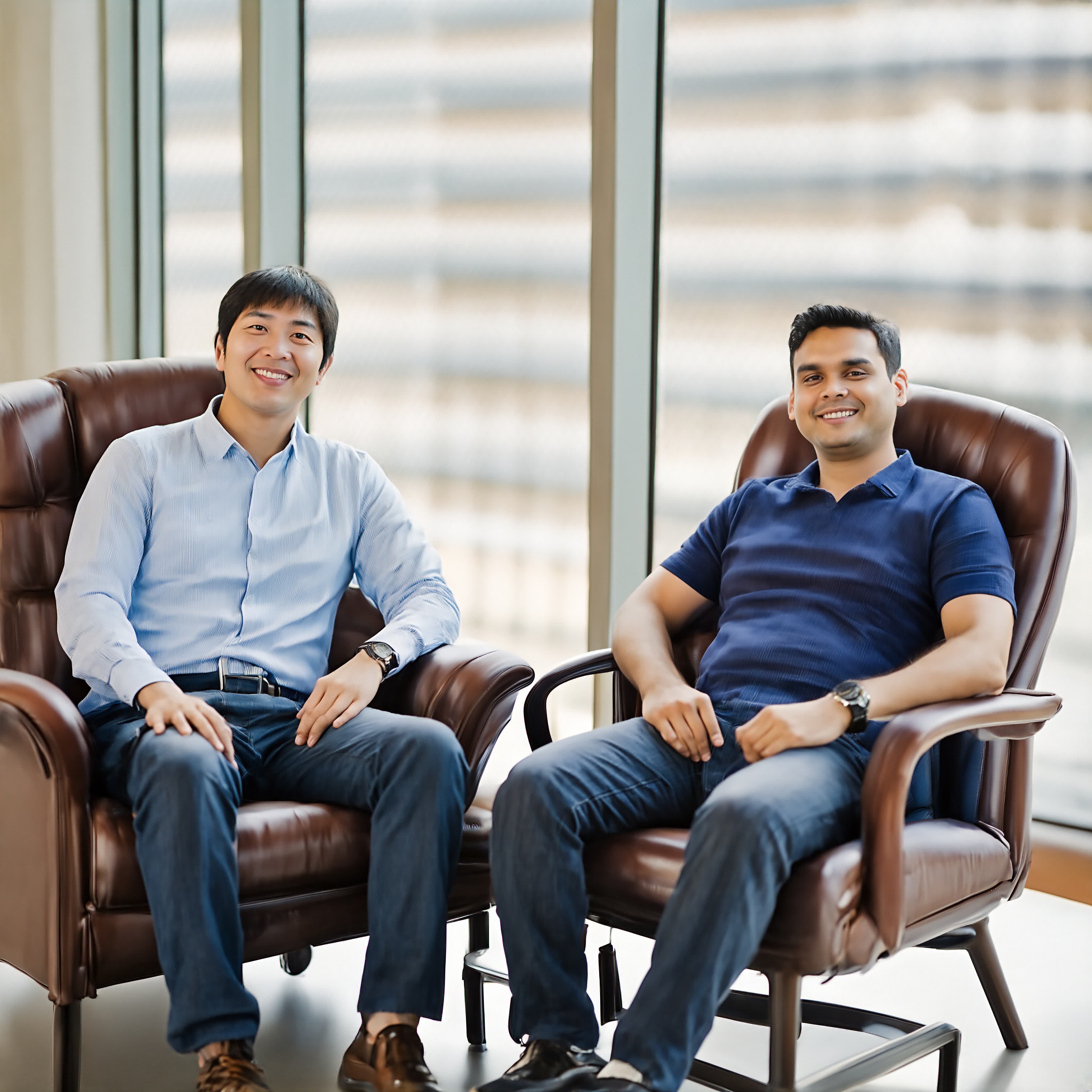} & 
    \includegraphics[width=0.33\textwidth]{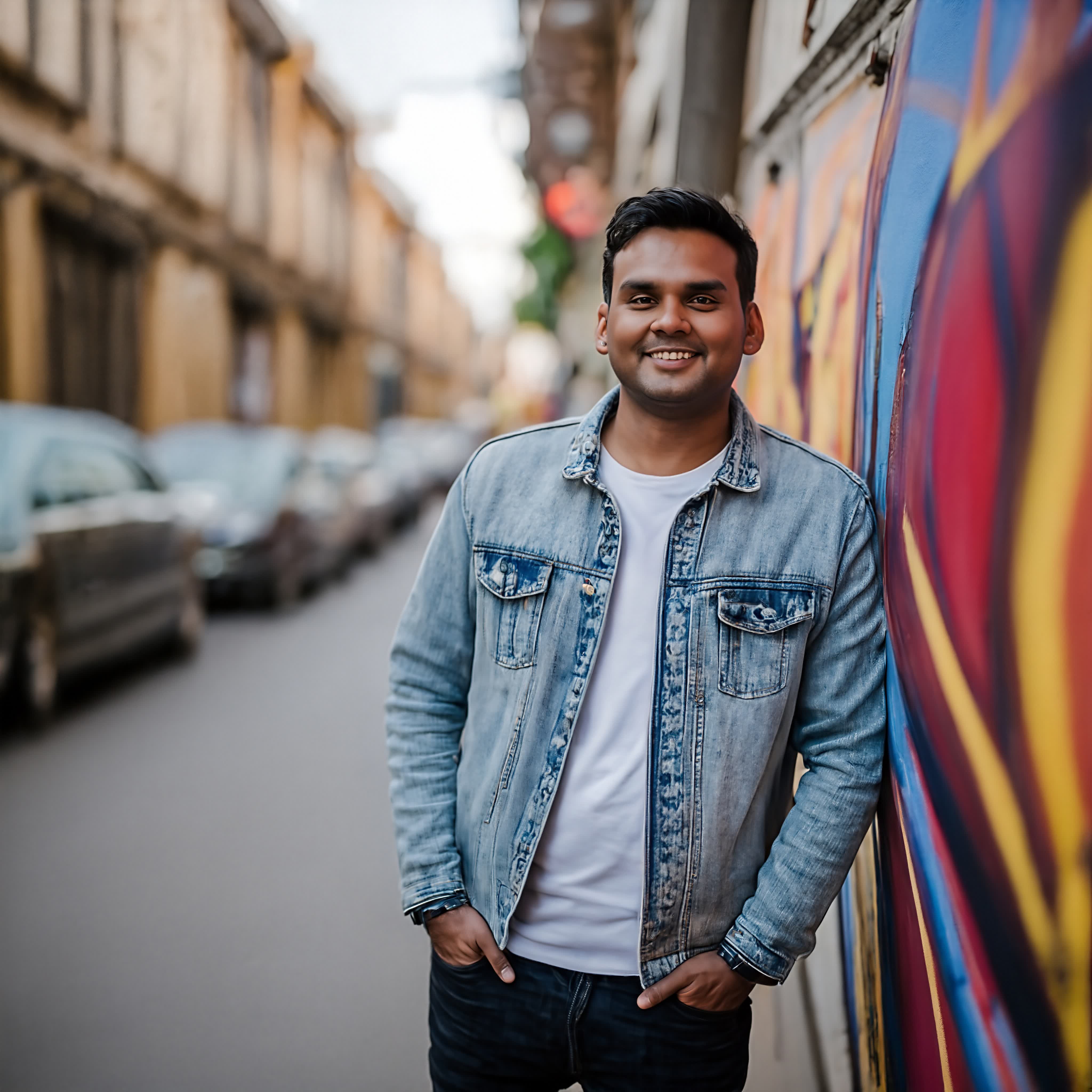} & 
    \includegraphics[width=0.33\textwidth]{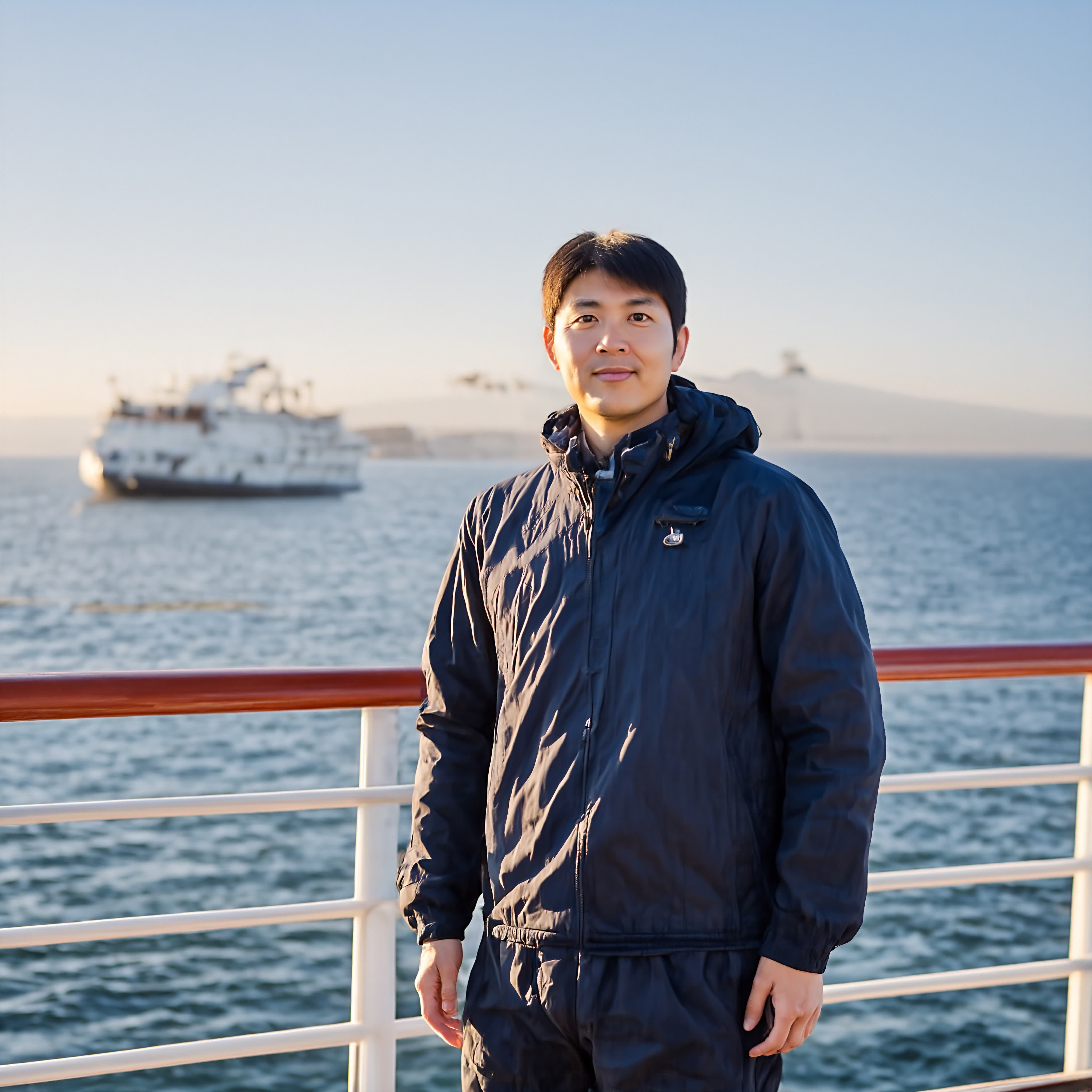}  \\
 \end{tabularx}
\captionof{figure}{
The finetuned model can generate images featuring two individuals in various settings and correctly distinguish between them based on their names in the text prompts.}
\label{fig:sid_sj}

\paragraph{Learning a single style} We finetuned the model on an icon dataset shown in \cref{fig:icon_train}. The results are presented in \cref{fig:icon}, demonstrating the model's ability to produce clear and sharp line drawings. \\

\begin{tabularx}{\textwidth}{M{0.25\textwidth} M{0.25\textwidth} M{0.25\textwidth} M{0.25\textwidth}}
    \includegraphics[width=0.25\textwidth]{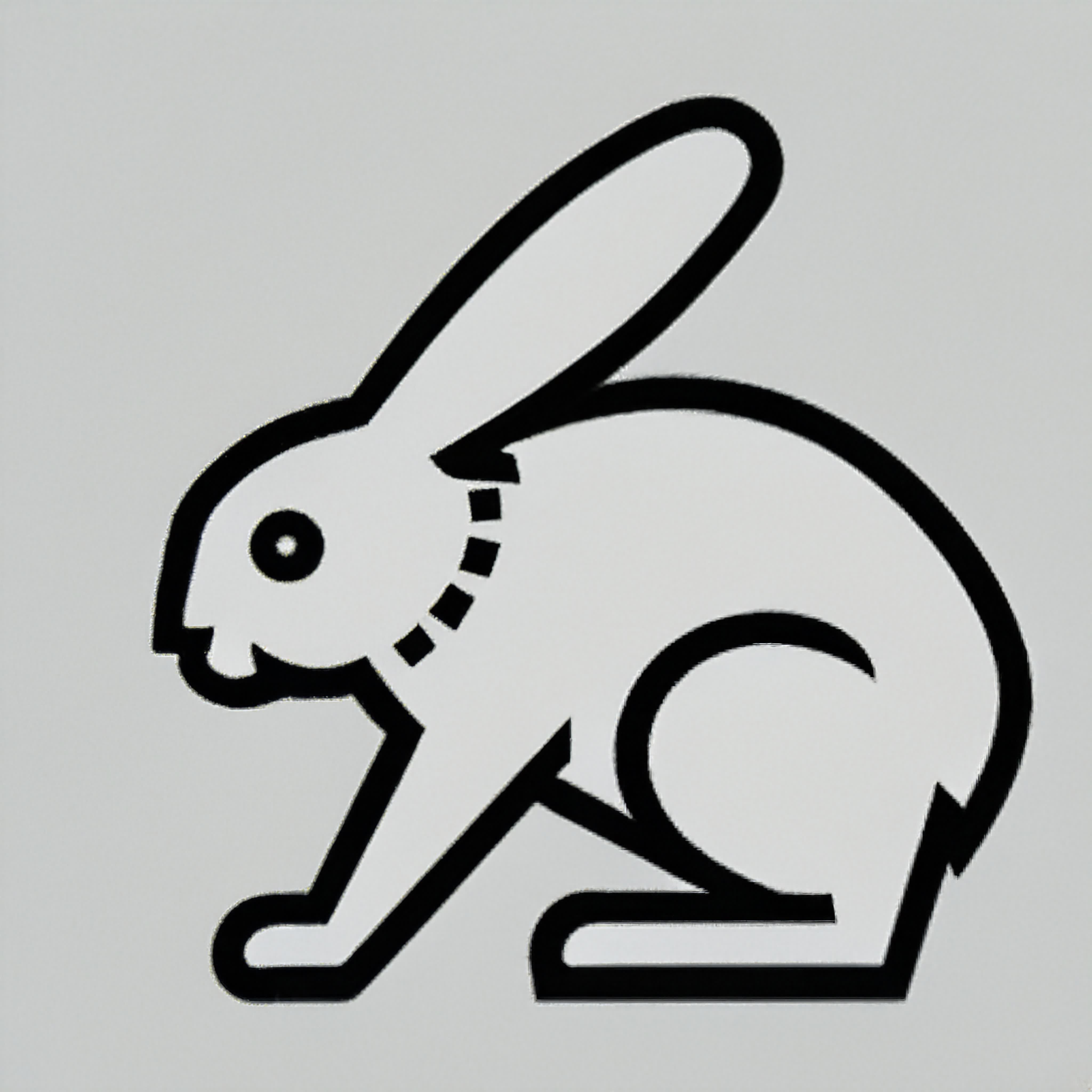} & \includegraphics[width=0.25\textwidth]{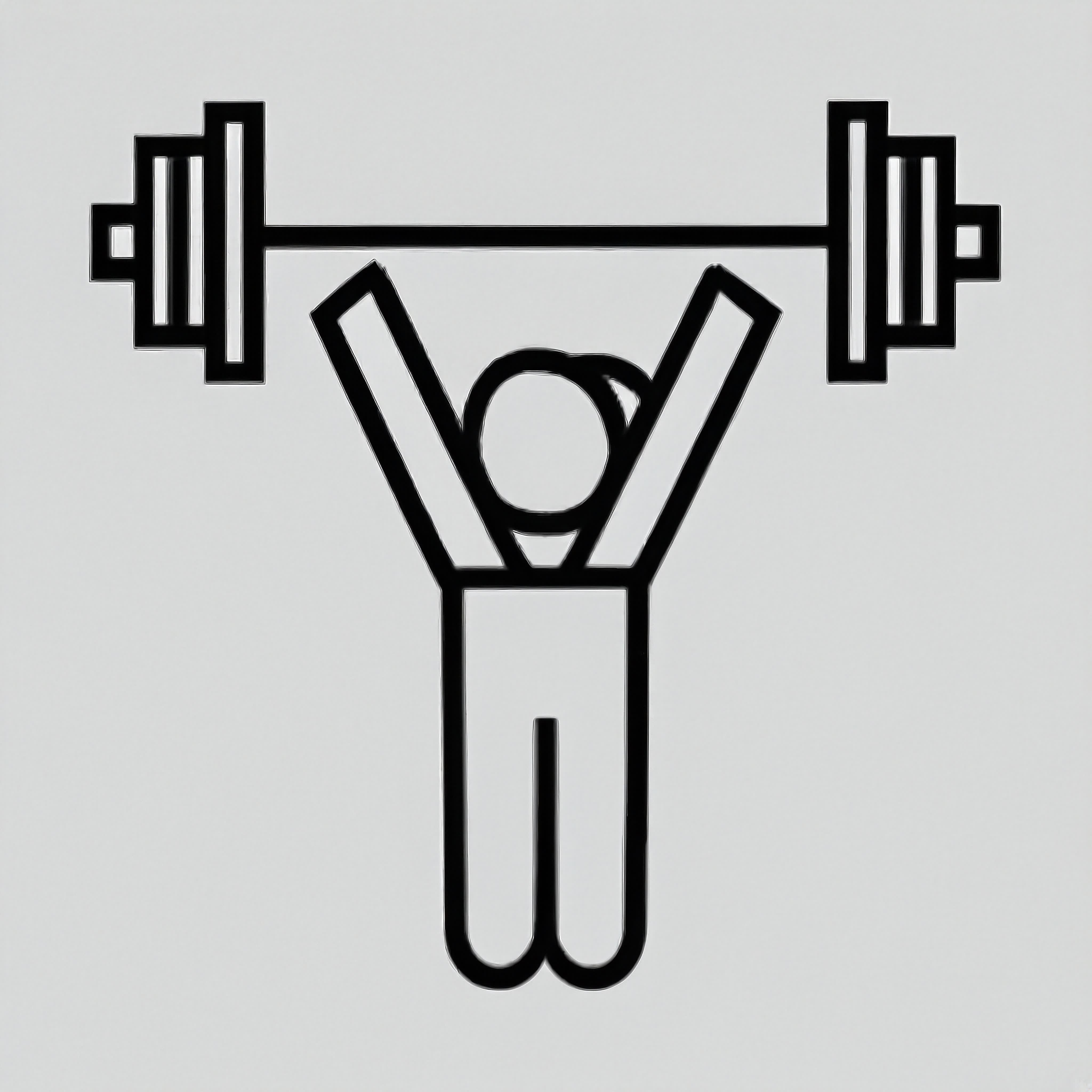}  &
    \includegraphics[width=0.25\textwidth]{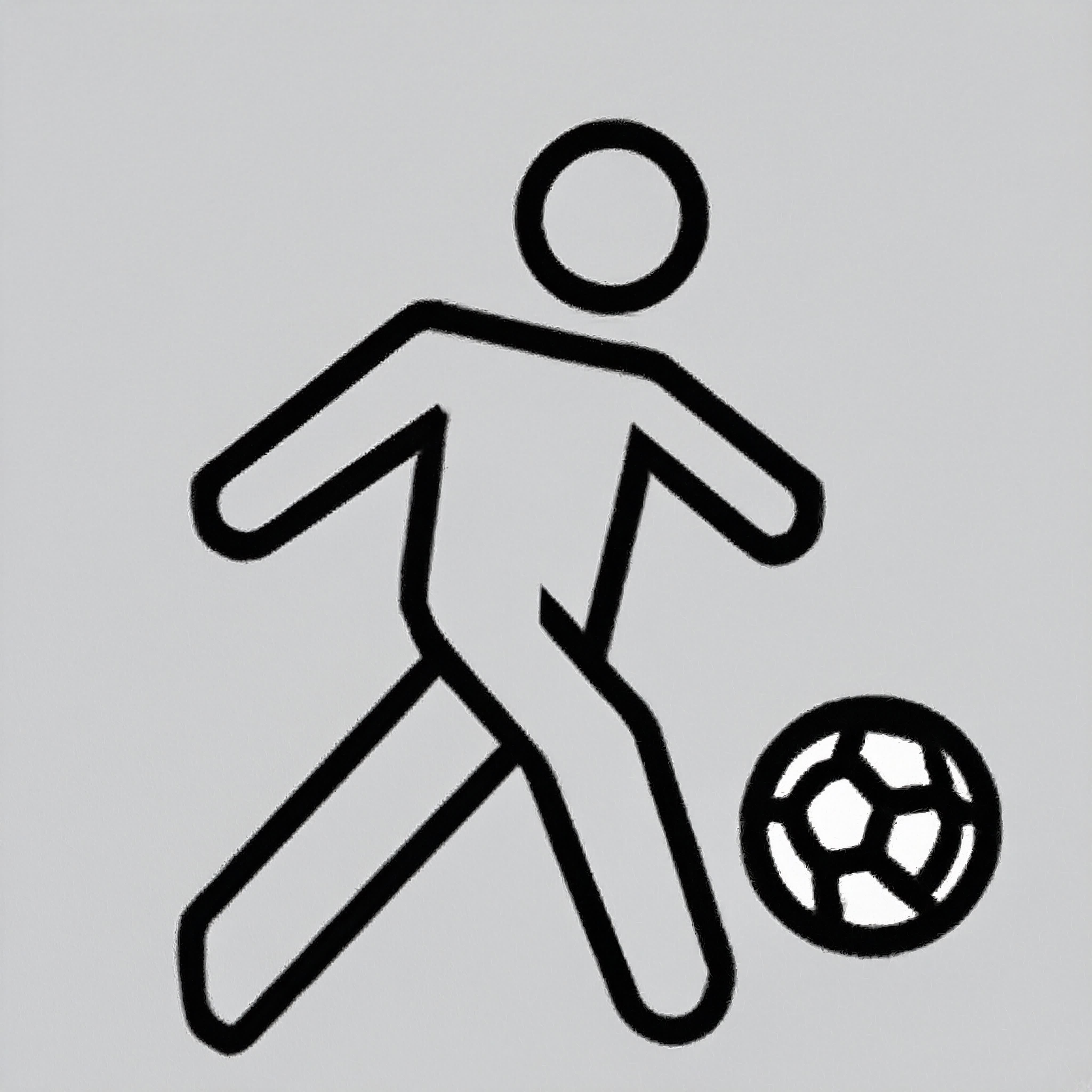} &
    \includegraphics[width=0.25\textwidth]{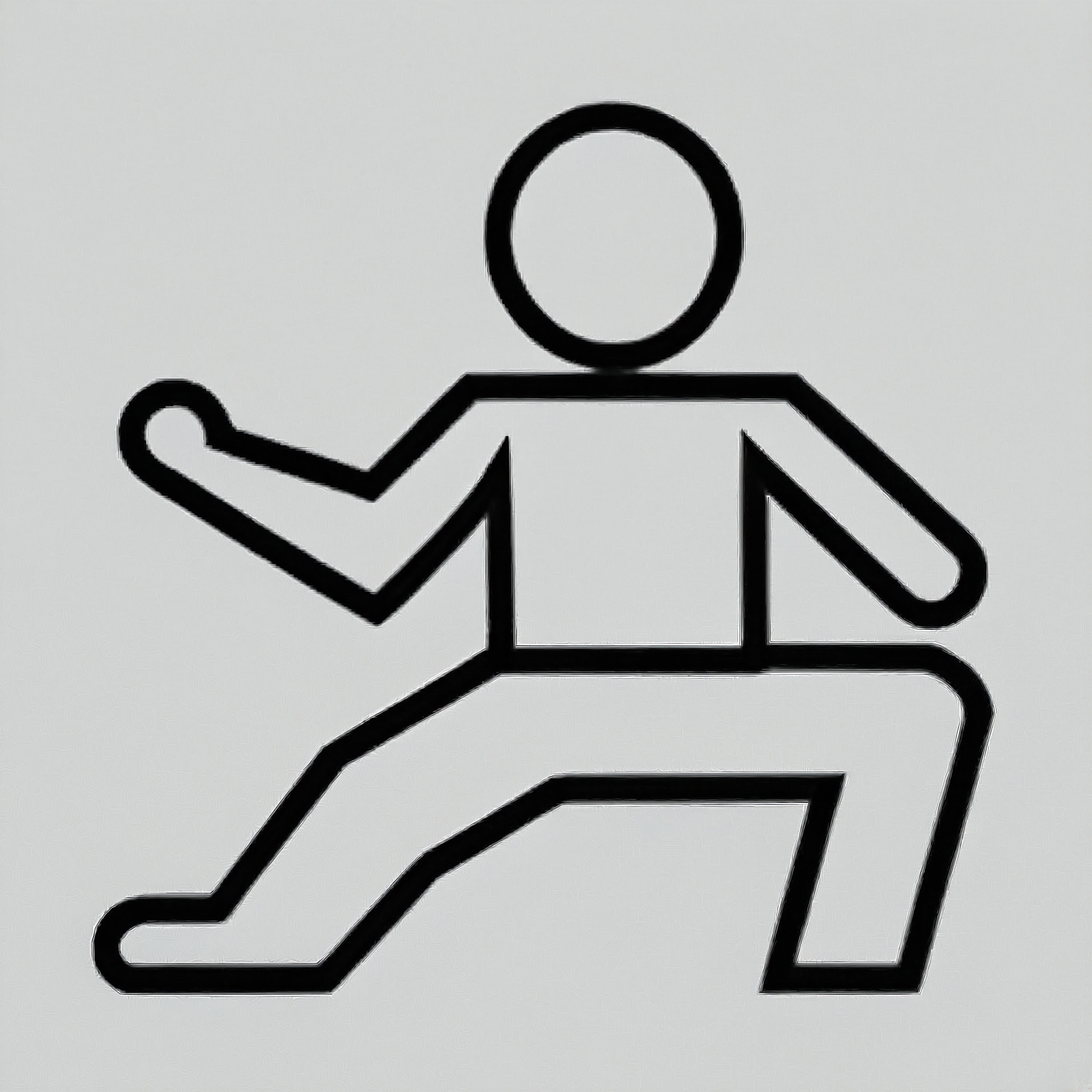} 
    \\    
    \makecell[{{p{0.24\textwidth}}}]{
            \emph{\small{A rabbit.}}
    } & 
    \makecell[{{p{0.24\textwidth}}}]{
            \emph{\small{A woman is weightlifting.}}
    } & 
    \makecell[{{p{0.24\textwidth}}}]{
            \emph{\small{A man is playing soccer.}}
    } & 
    \makecell[{{p{0.24\textwidth}}}]{
            \emph{\small{A man is practicing Tai Chi.}}
    } \\    
\end{tabularx}
\captionof{figure}{The finetuned model successfully replicates the iconographic style with high fidelity.}
\label{fig:icon}

\begin{figure}[htbp]
    \centering
    \begin{tabular}{cccc}
    \multicolumn{2}{c}{{Learned styles from finetuning data}} & 
    \multicolumn{2}{c}{{Known styles in the base model}} \\ [2mm]
    \includegraphics[width=0.25\textwidth]{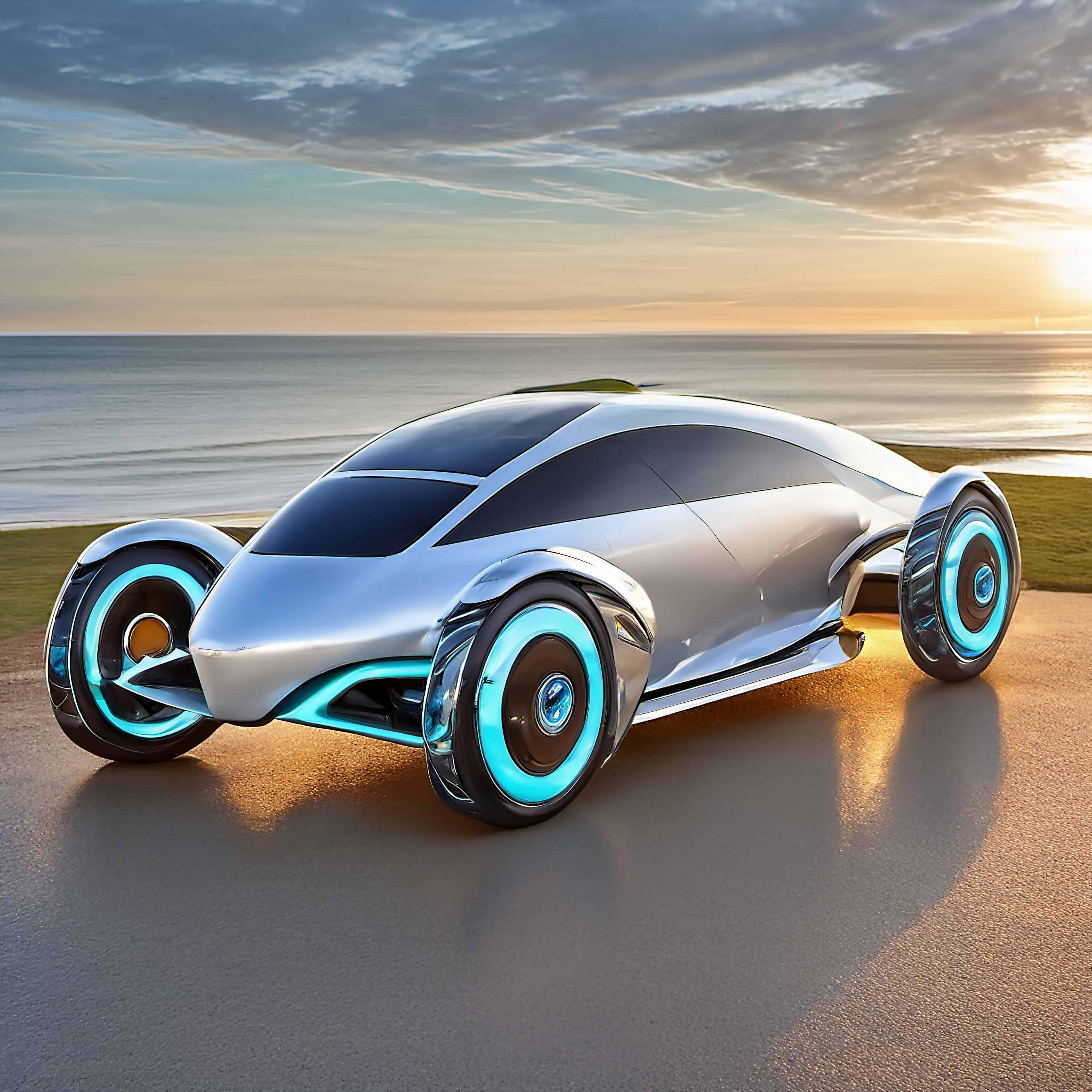} &  
    \includegraphics[width=0.25\textwidth]{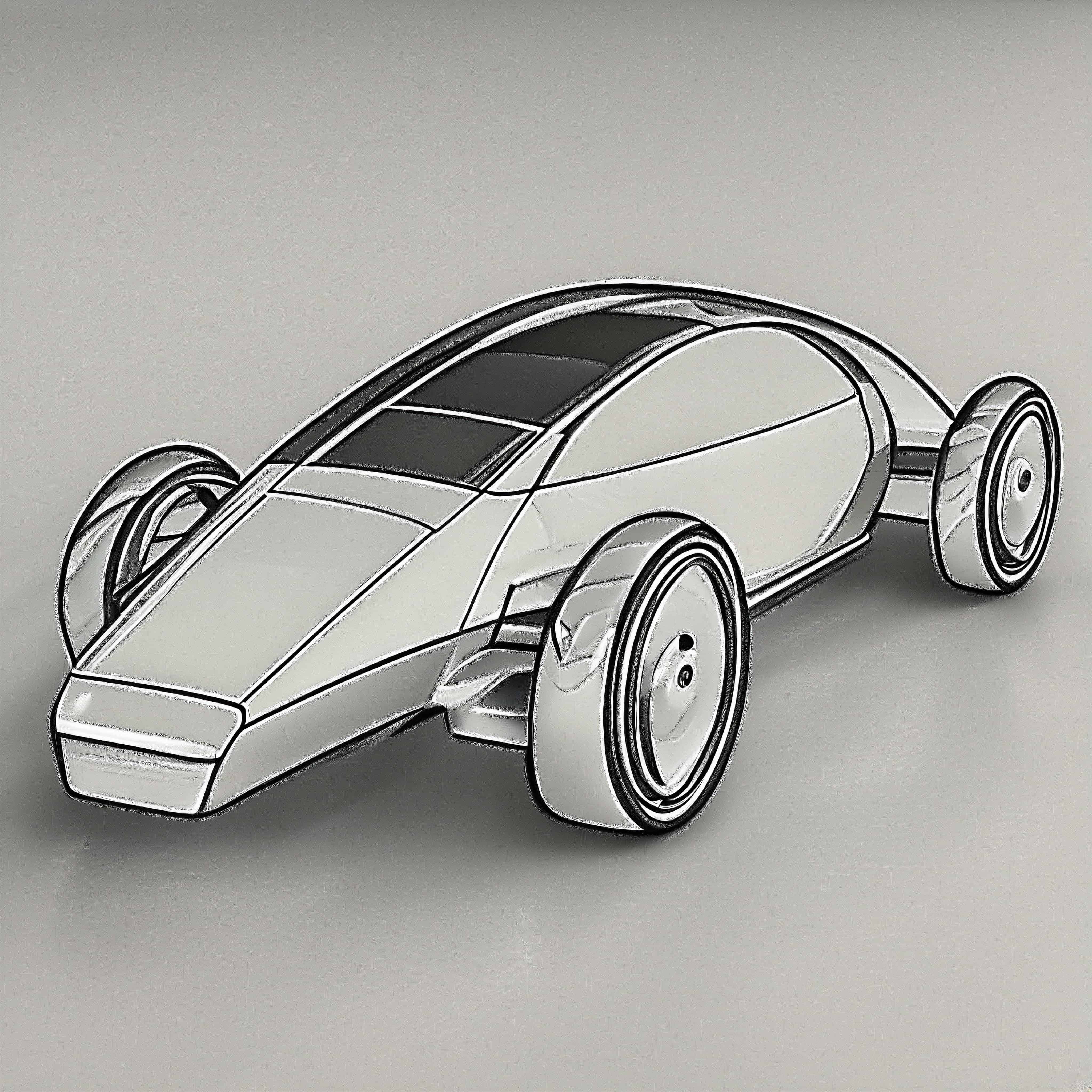} &
    \includegraphics[width=0.25\textwidth]{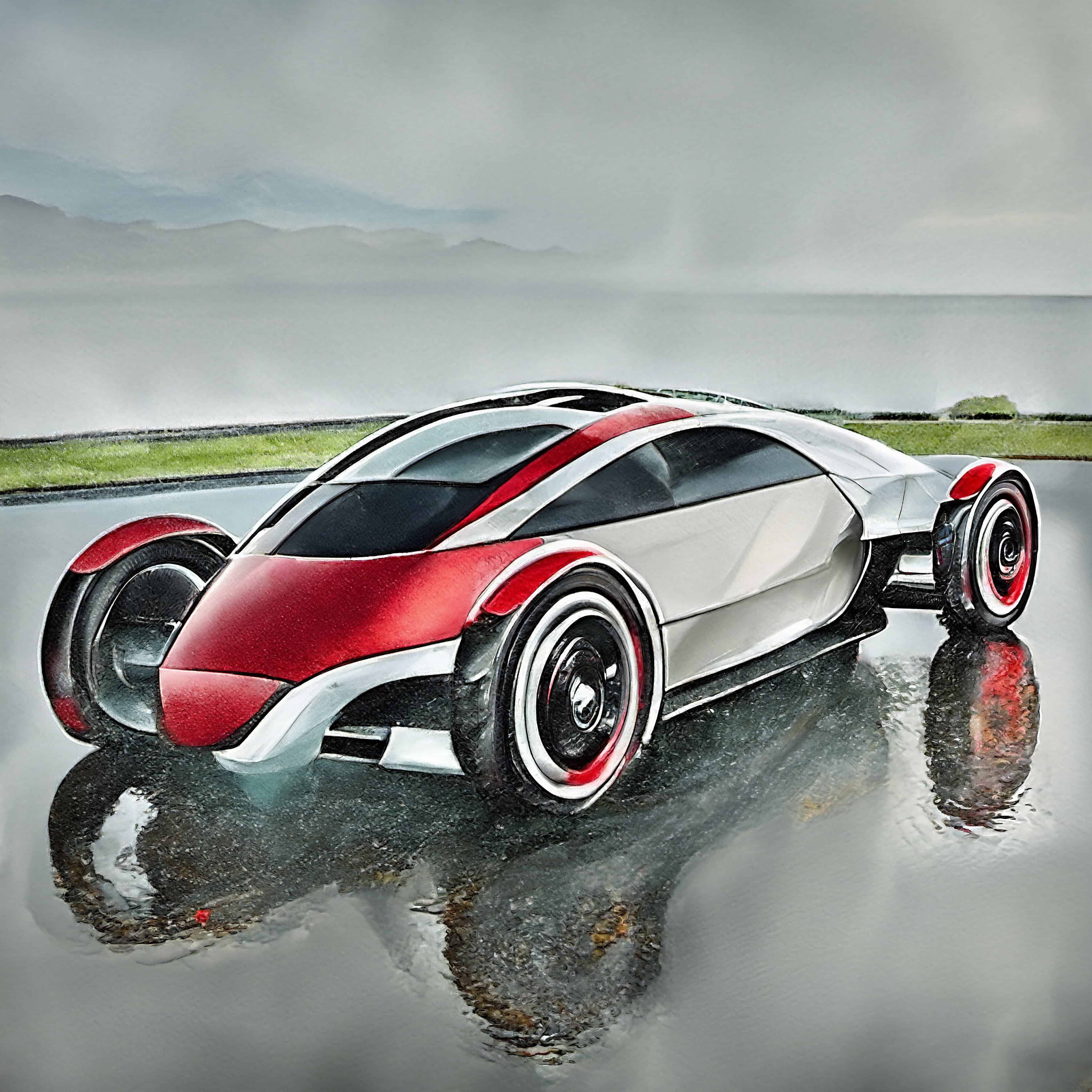}&
    \includegraphics[width=0.25\textwidth]{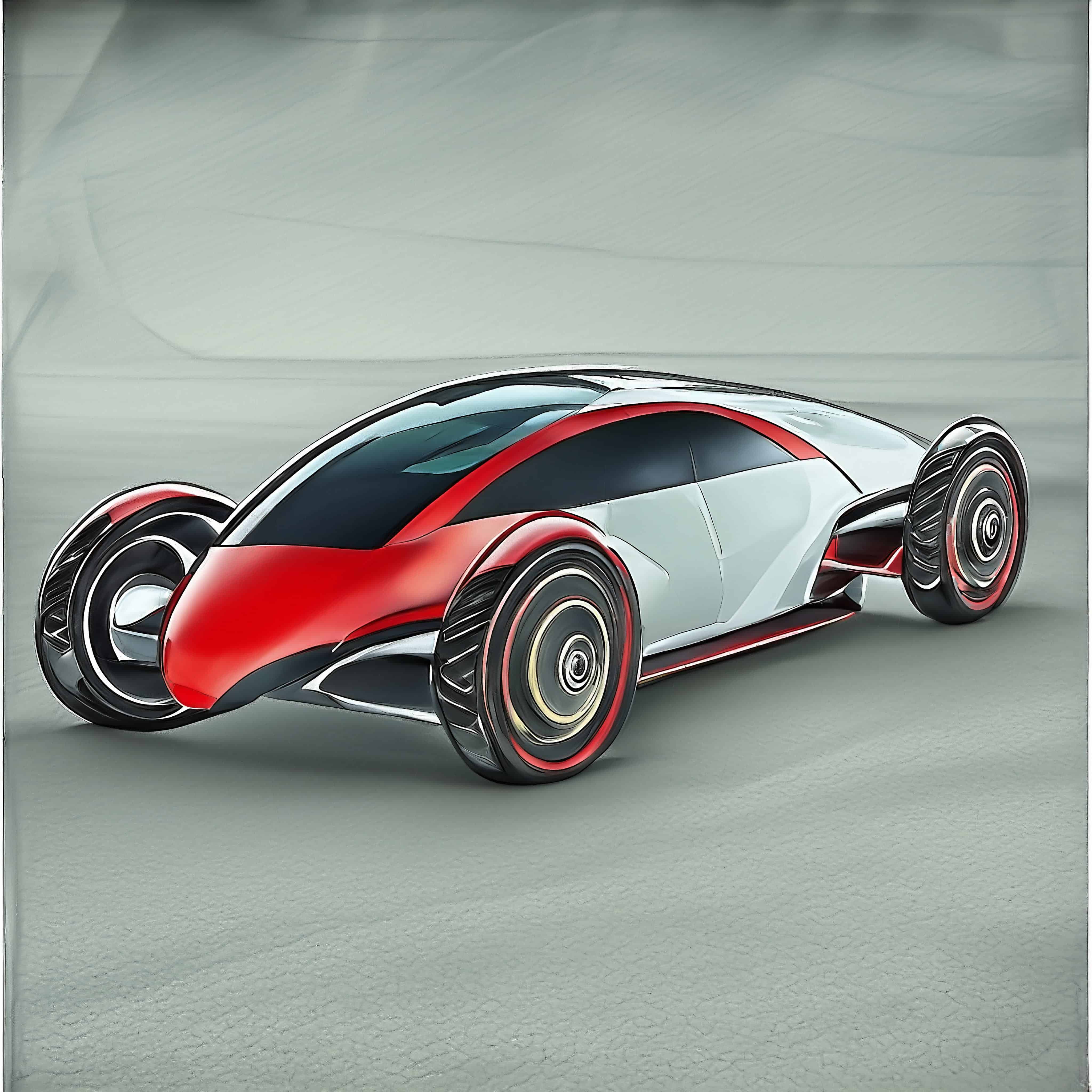} \\ [1mm]
    \emph{\small{Epic}} & 
    \emph{\small{Line art}} & 
    \emph{\small{Watercolor}} & 
    \emph{\small{Comic sketch}} \\
    \end{tabular}
    \caption{The finetuned model is capable of learning multiple new styles while retaining knowledge of existing ones.}
    \label{fig:car}
\end{figure}

\paragraph{Learning multiple styles} We also finetuned the model on the dataset in \cref{fig:car_train} to enable it to learn multiple styles. Different style names, such as "Epic" and "Line Art" were used in the training prompts to help the model distinguish among various styles. The results are shown in \cref{fig:car}. 

\subsection{Compatibility with Edify ControlNet}

Our finetuning approach maintains the model architecture, allowing the finetuned model to be easily integrated with pre-trained, frozen ControlNet modules. In \cref{fig:ft_ctrl}, we demonstrate that the finetuned U-Net can still effectively work with inpainting, sketch, and depth ControlNets, while faithfully preserving the learned subject in the controlled generation. \\

\begin{tabularx}{\textwidth}{M{0.33\textwidth} M{0.33\textwidth} M{0.33\textwidth}}
    \includegraphics[width=0.33\textwidth]{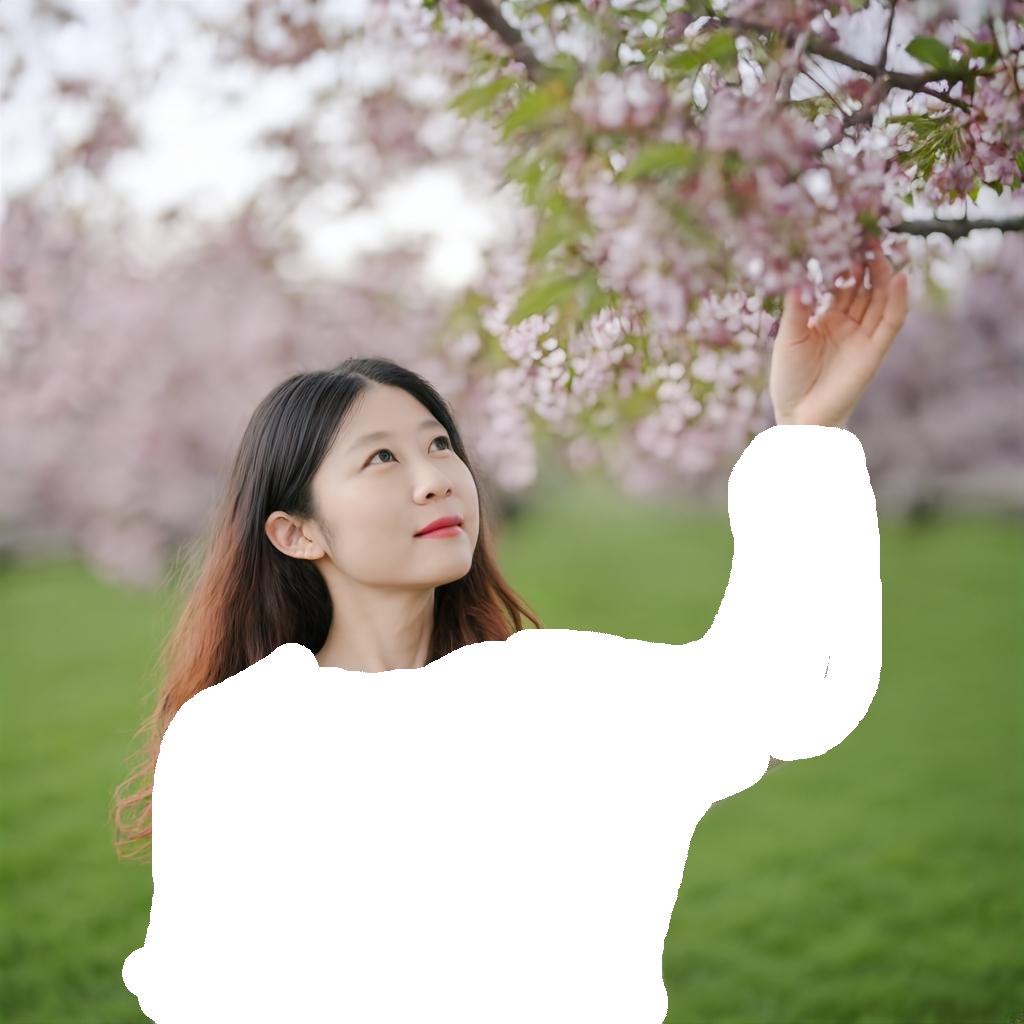} & \includegraphics[width=0.33\textwidth]{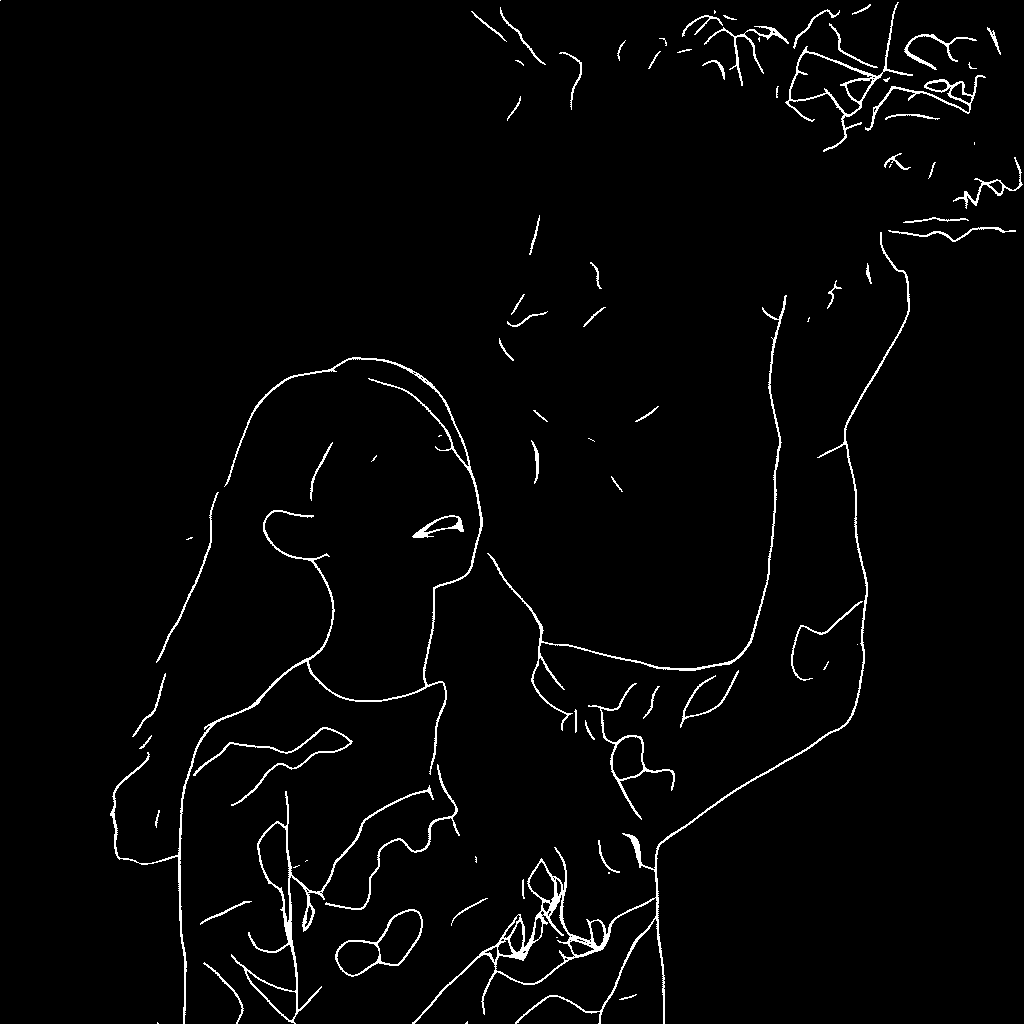} & 
\includegraphics[width=0.33\textwidth]{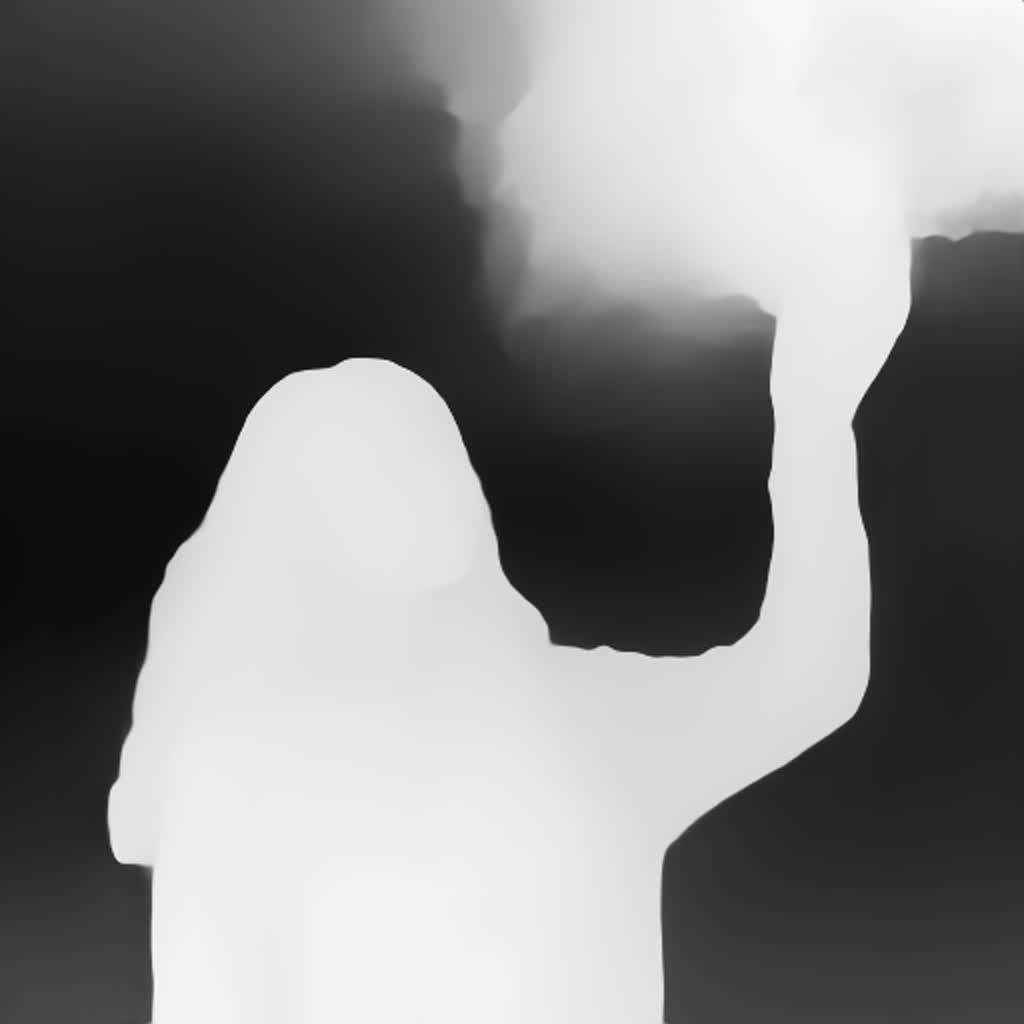}    
    \\
    \includegraphics[width=0.33\textwidth]{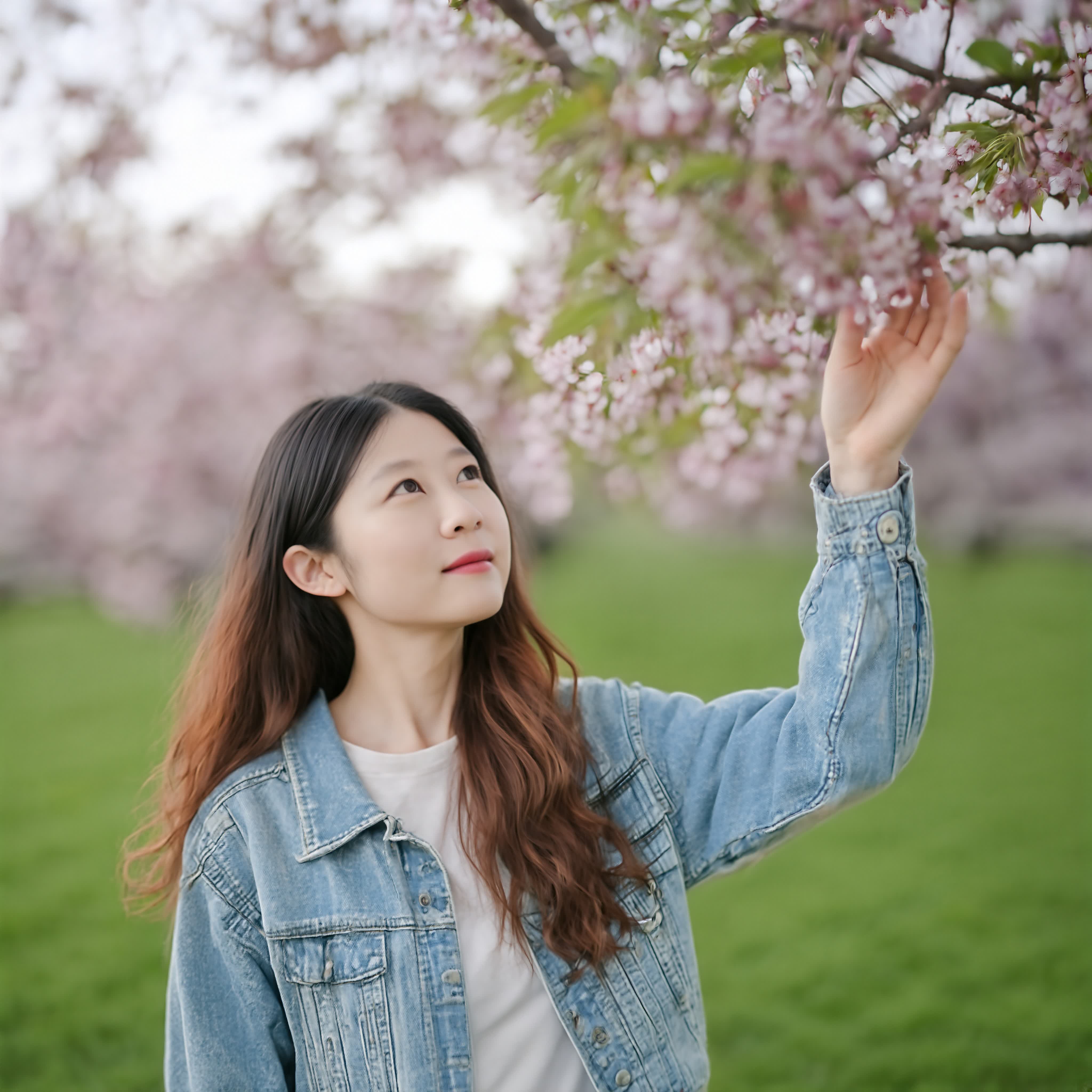} & 
\includegraphics[width=0.33\textwidth]{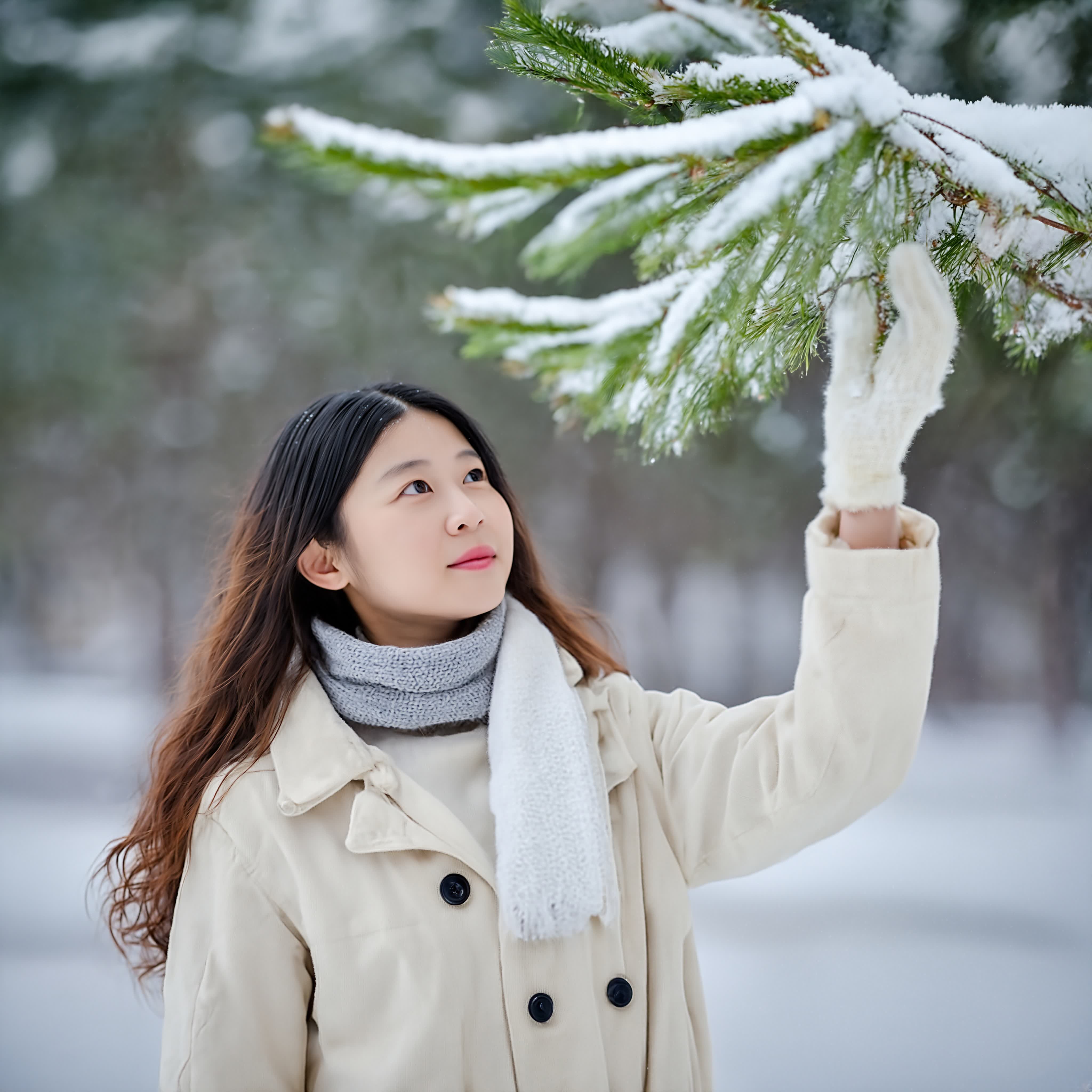}  &   
\includegraphics[width=0.33\textwidth]{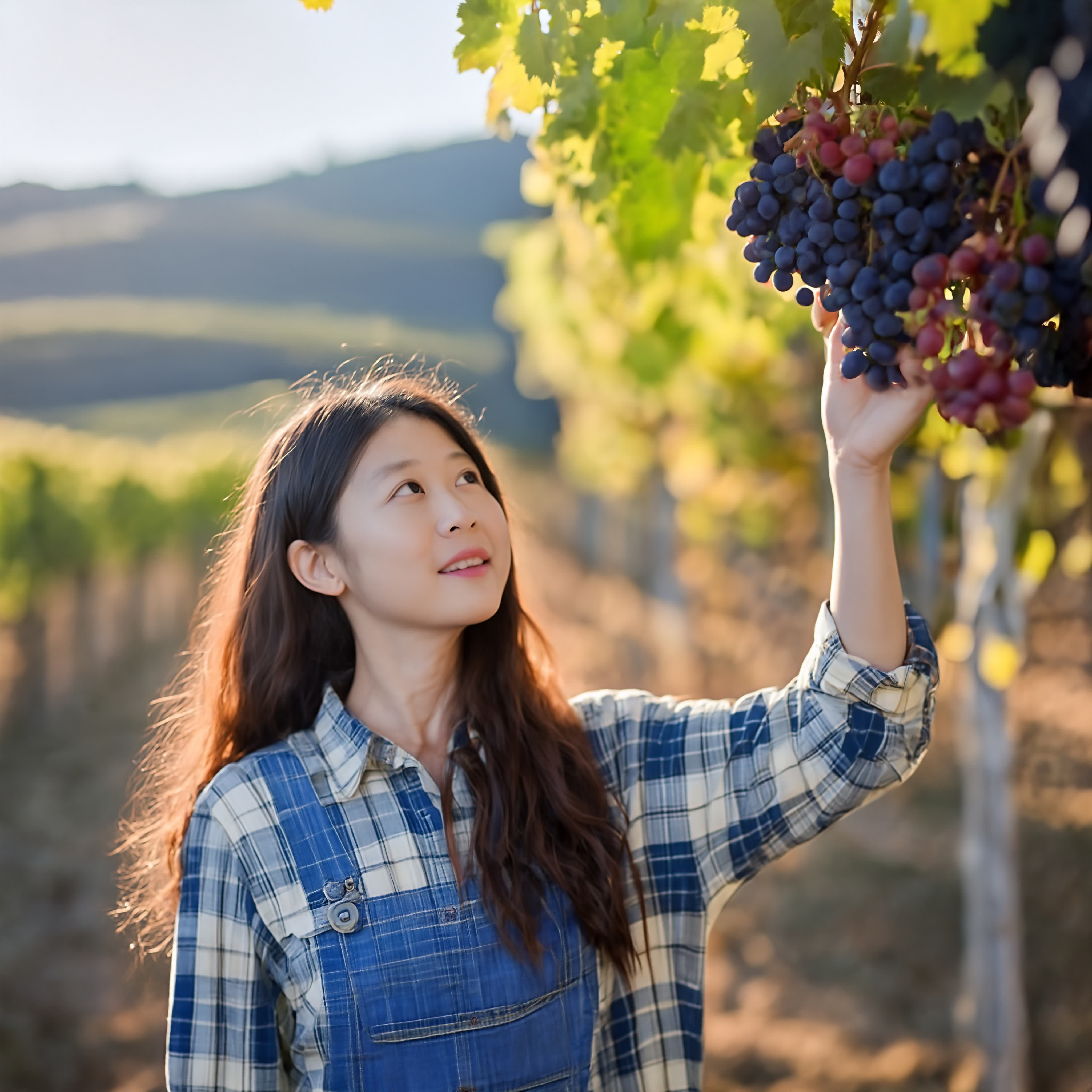} 
    \\   
\end{tabularx}
\captionof{figure}{\textbf{ControlNet compatibility.} The finetuned U-Nets remain compatible with our inpainting, sketch, and depth ControlNets, which are not finetuned on the personalization datasets. The top row shows the control inputs, while the bottom row shows the generations.}
\label{fig:ft_ctrl}

\subsection{Ablation Study}

\begin{figure}[htbp]
    \centering
    \begin{tabular}{m{0.12\textwidth}cccc}    
    \makecell[{{p{0.115\textwidth}}}]{\raggedright \vspace{-2.5cm} \small{Training data includes photos taken over 6 years}} &    
    \includegraphics[width=0.2\textwidth]{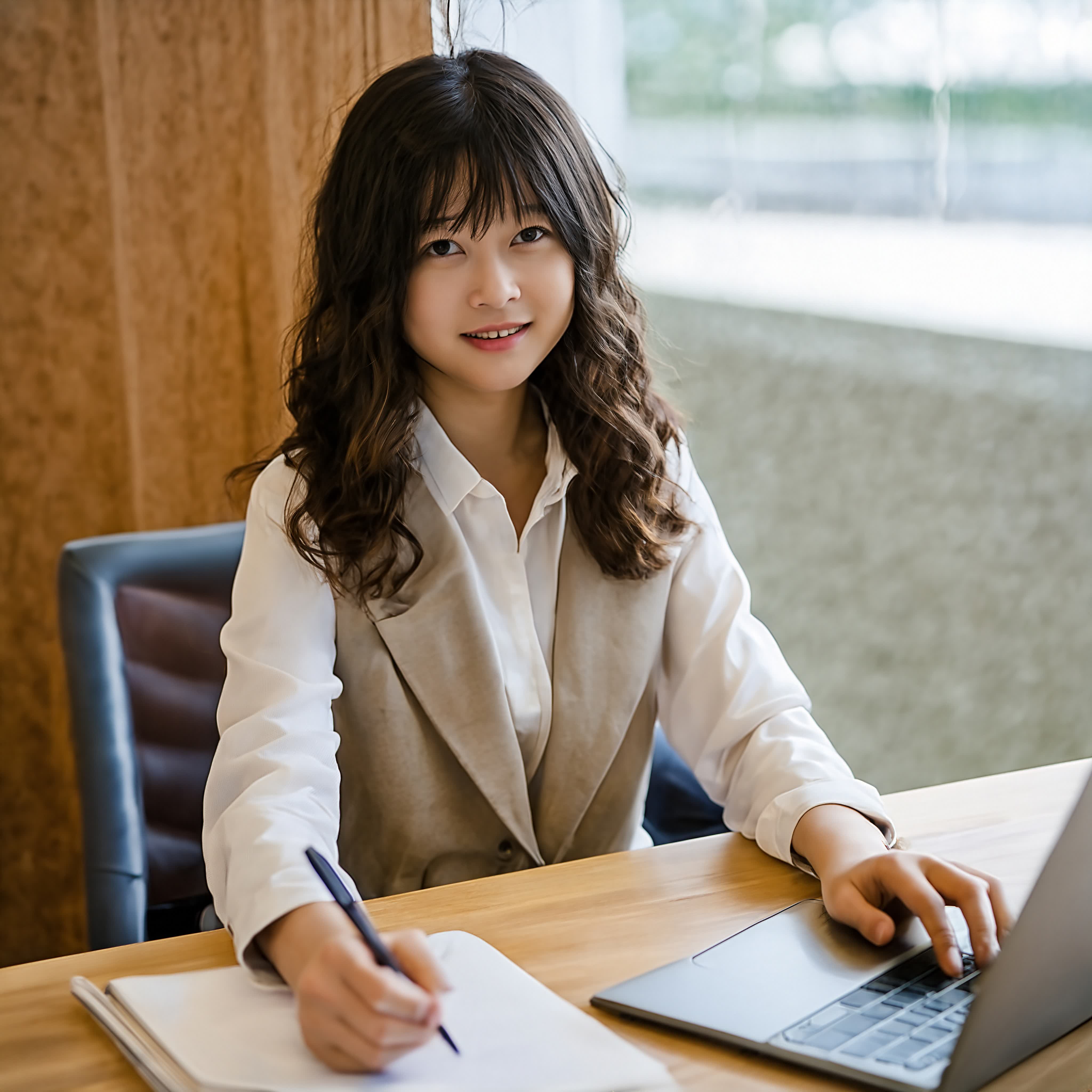} &
    \includegraphics[width=0.2\textwidth]{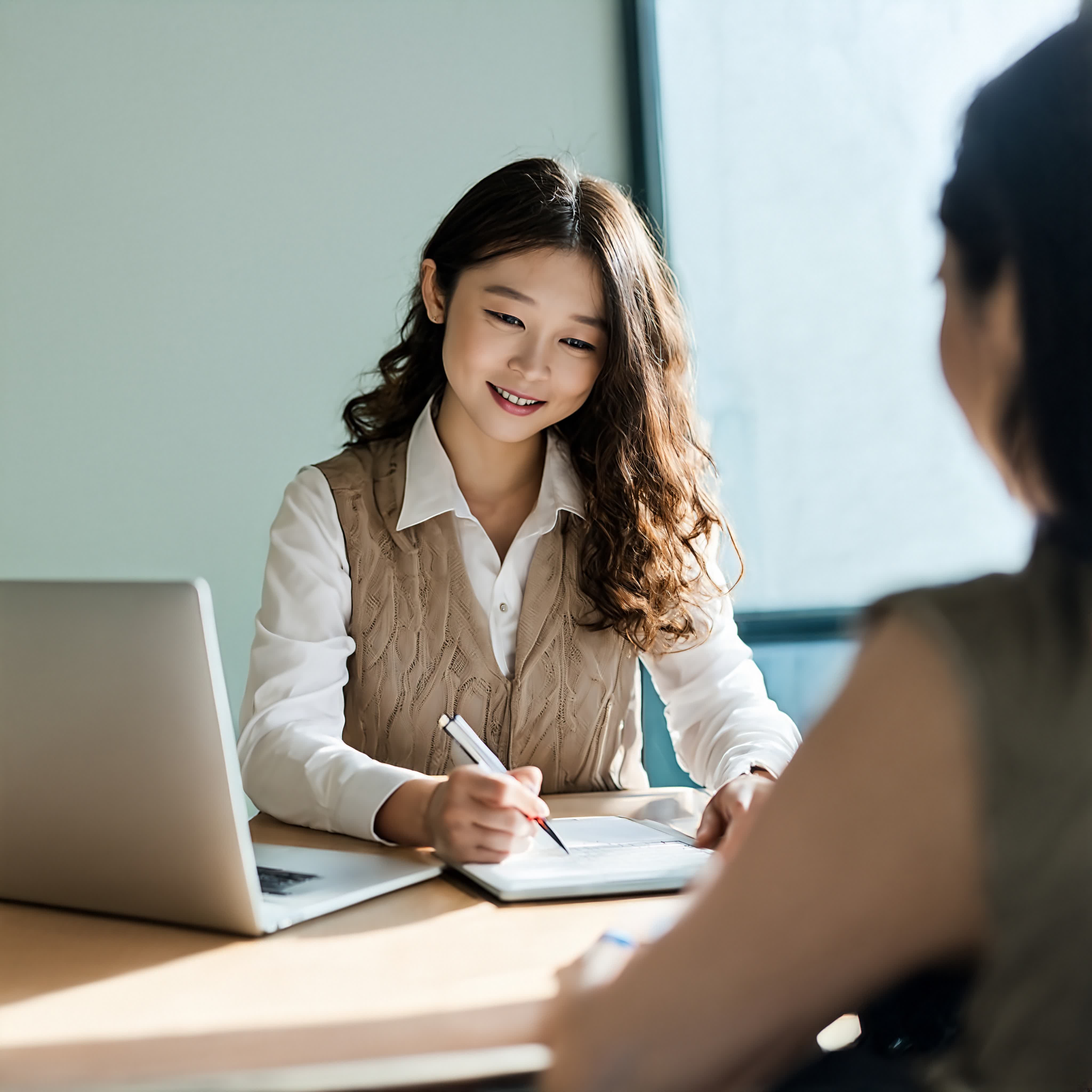} &
    \includegraphics[width=0.2\textwidth]{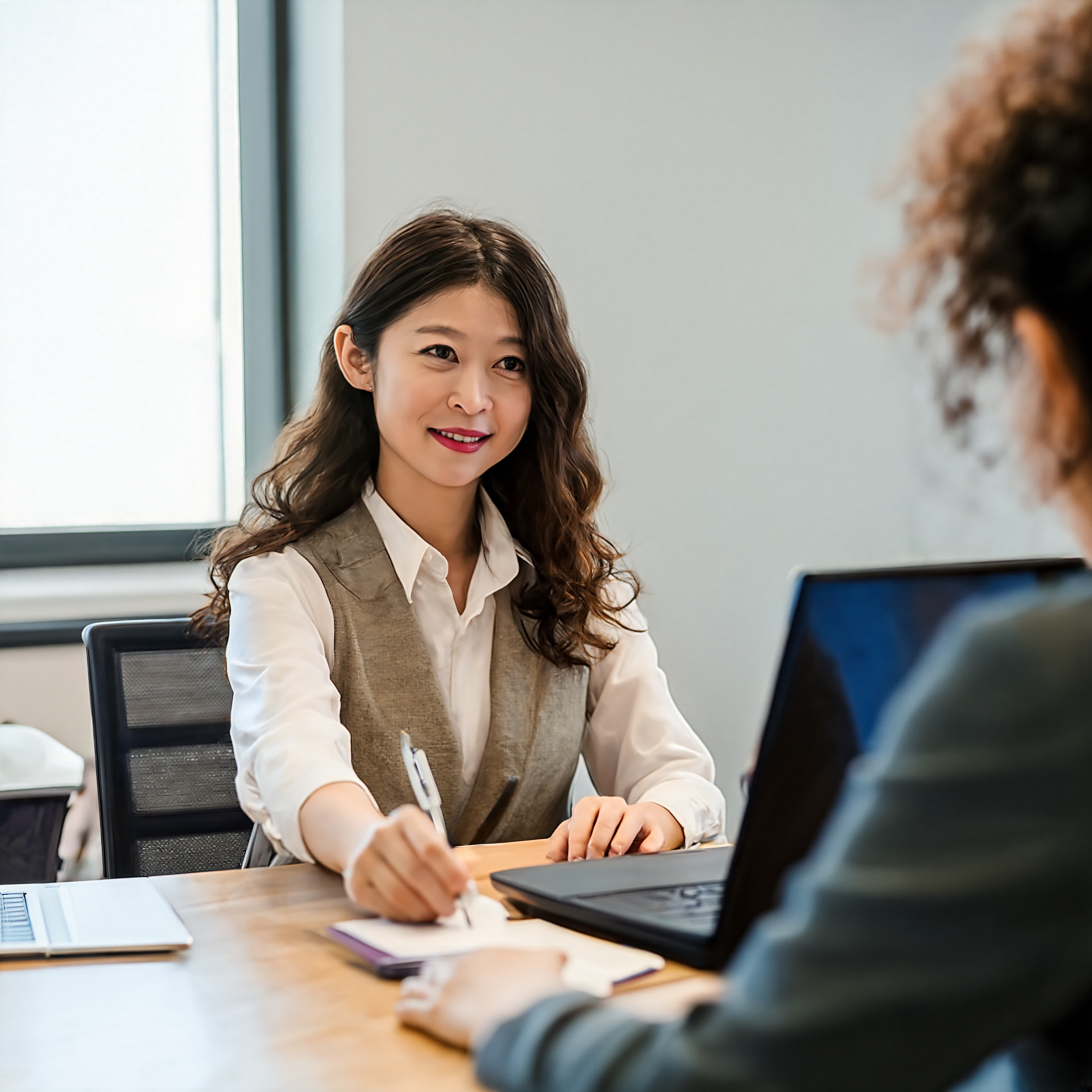} &
    \includegraphics[width=0.2\textwidth]{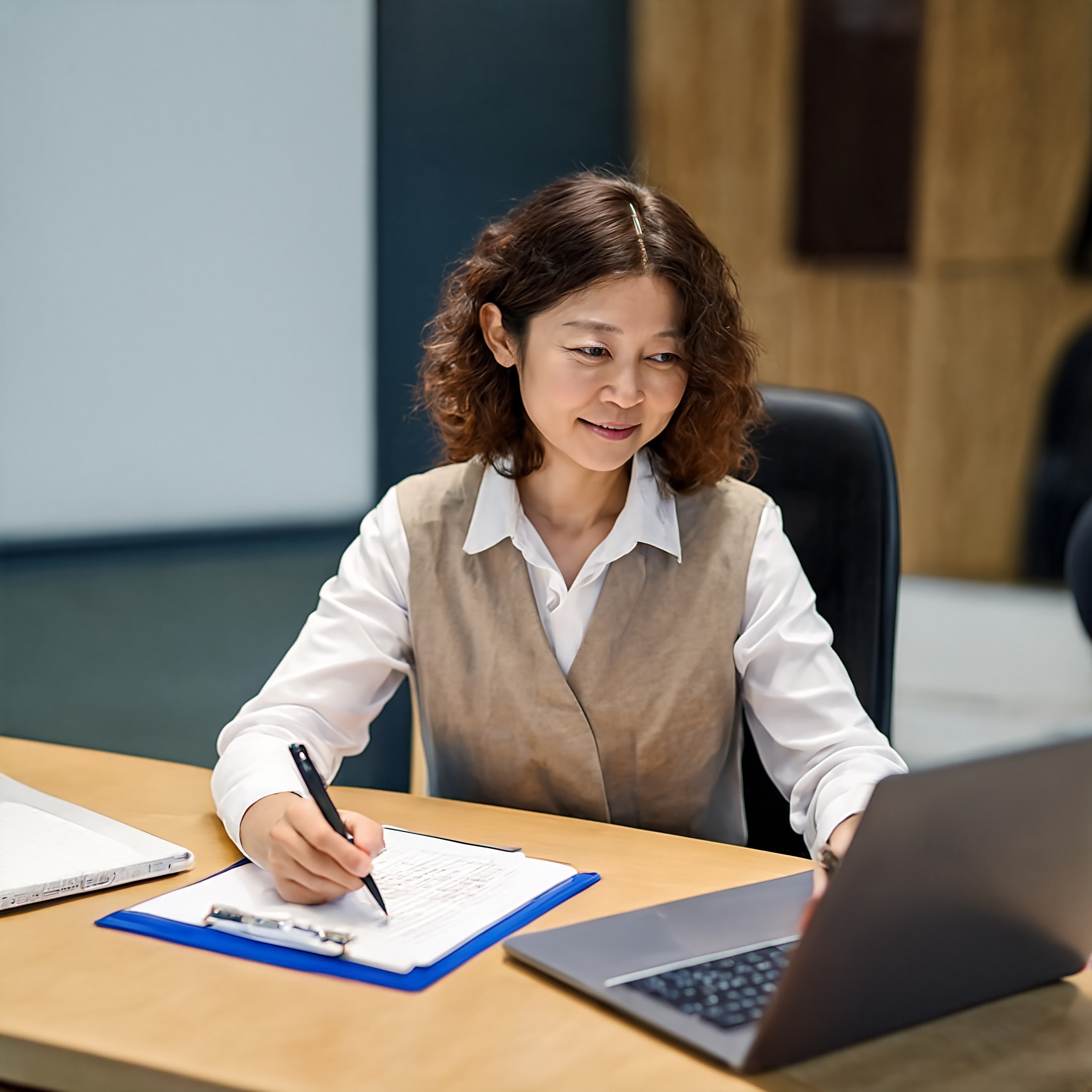} \\ [1mm]

    \makecell[{{p{0.115\textwidth}}}]{\raggedright \vspace{-2.2cm} \small{Training data includes only recent photos}} &    
    \includegraphics[width=0.2\textwidth]{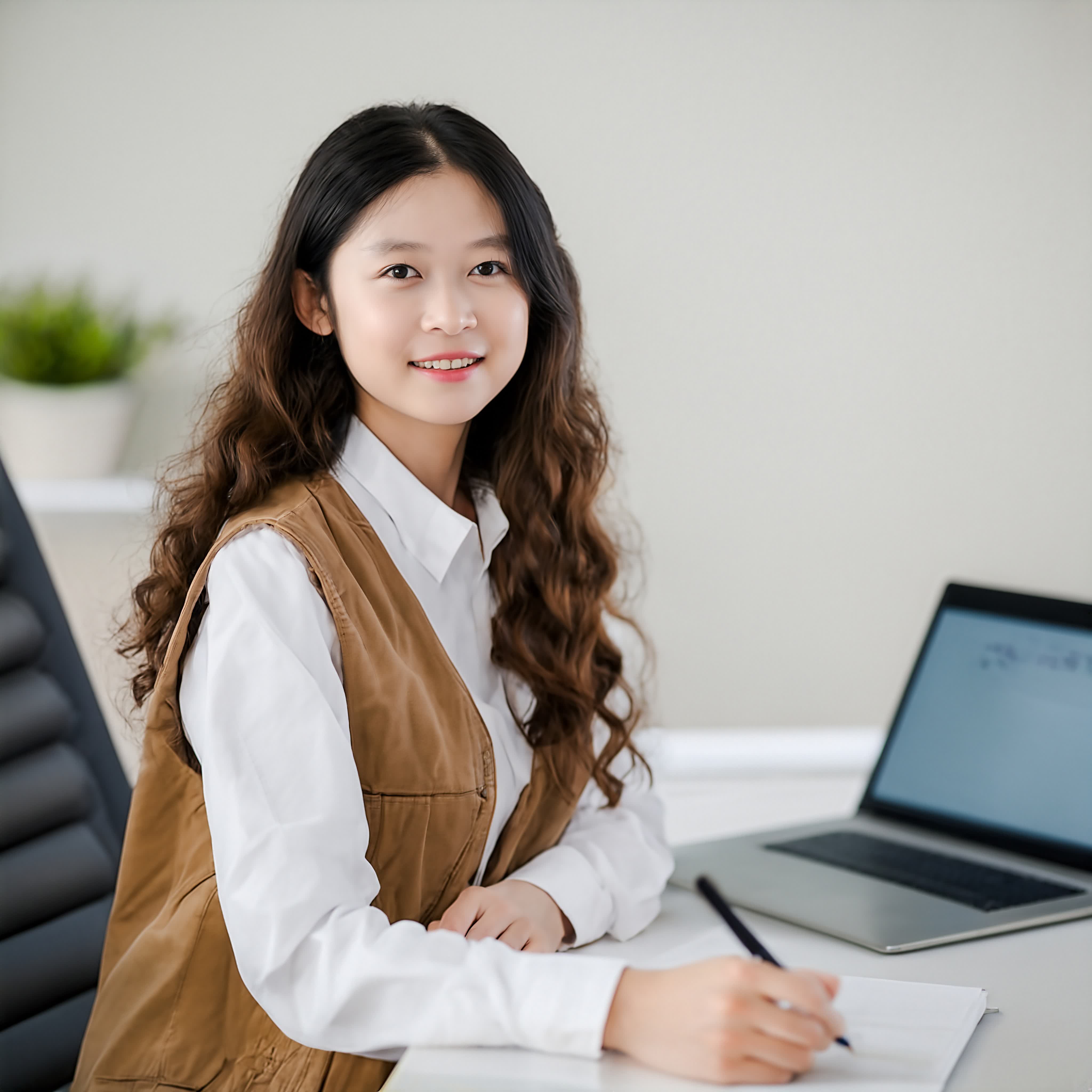} &
    \includegraphics[width=0.2\textwidth]{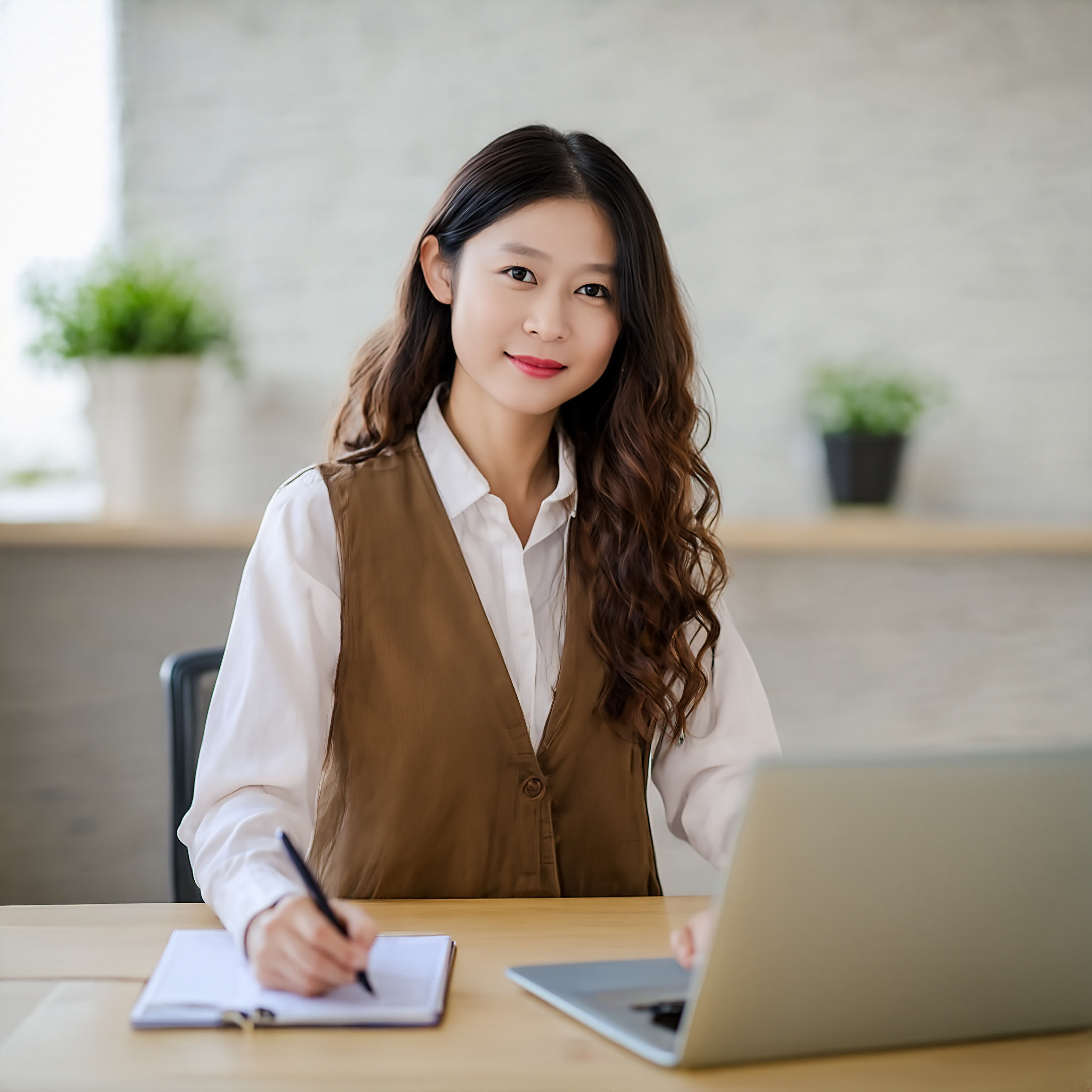} &
    \includegraphics[width=0.2\textwidth]{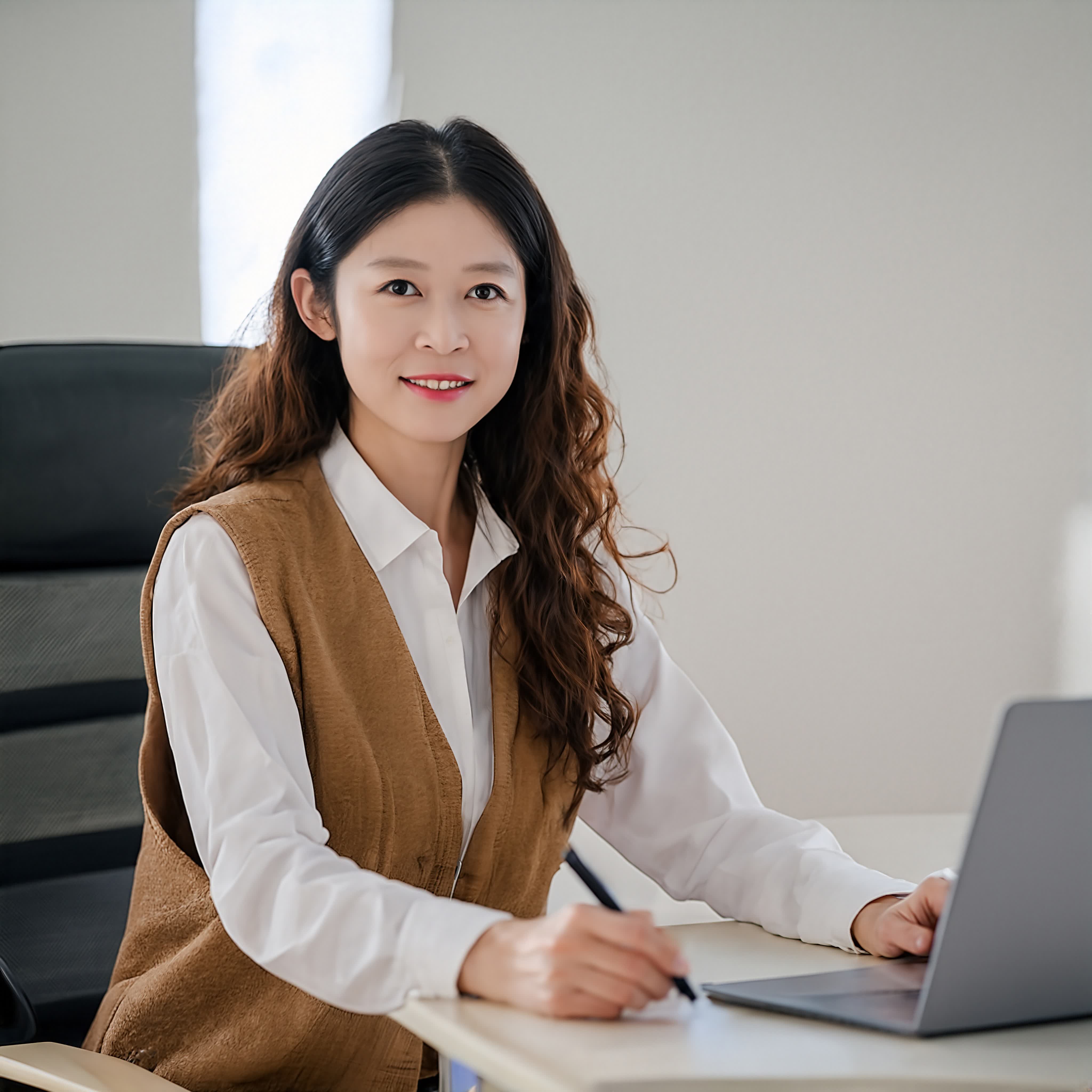} &
    \includegraphics[width=0.2\textwidth]{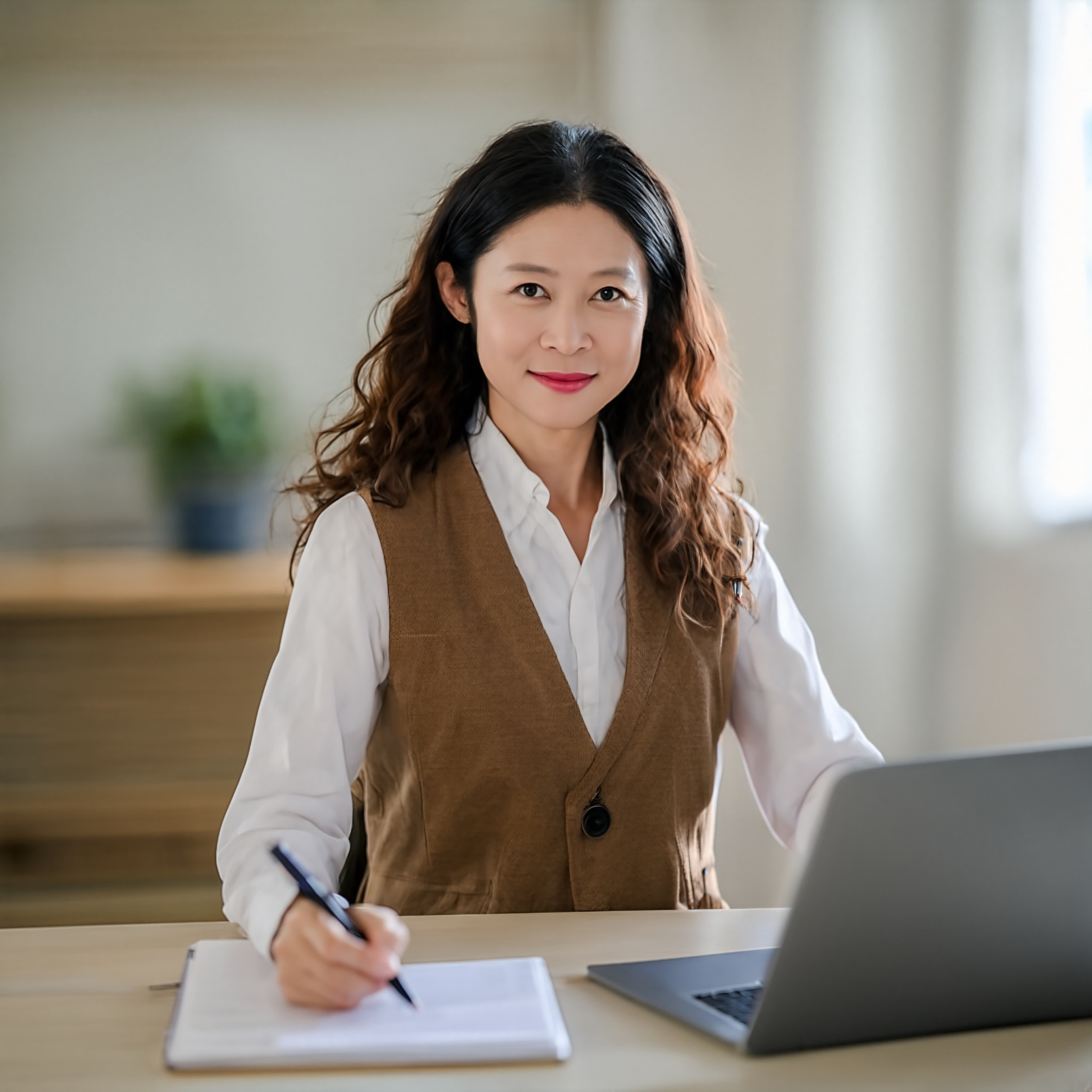} \\ [1mm]

    \multicolumn{1}{c}{} &
    \texttt{\small{age = 16}} & 
    \texttt{\small{age = 26}} & 
    \texttt{\small{age = 36}} & 
    \texttt{\small{age = 46}} \\ [2mm]

    \multicolumn{1}{c}{} &
    \multicolumn{4}{p{0.8\textwidth}}{\small{\emph{FangyinWei, a professional} \{\texttt{age}\}\emph{-year old woman sitting upright at a desk with one hand holding a pen in an office, smiling and engaged with her laptop, short, curly hair that frames her face, dressed in a white shirt and light brown vest.}}} \\
    \end{tabular}
    \caption{\textbf{Effect of data diversity.} Both training datasets lack explicit age labels in the captions and contain only images of the subject in her 20s. However, a more diverse training dataset facilitates generating the character across a broader range of ages.}
    \label{fig:data_diversity}
\end{figure}

We conducted an ablation study to investigate the impact of training data diversity on the generalization ability of the finetuned model. As shown in \cref{fig:data_diversity}, we finetuned the model using two datasets: one containing the subject's photos taken over six years, and another containing only recent photos. Although both datasets include images of the subject in her 20s, the first dataset makes it easier to generate the subject at younger or older ages (\eg, in her 30s or 40s) and yields more diverse outputs.

\section{Related Work}\label{sec:relatedwork}

\paragraph{Text-to-Image Generation} Diffusion models have emerged as the dominant approach for high-resolution image generation since the seminal work of \cite{ho2020denoisingdiffusionprobabilisticmodels}. For text-to-image synthesis, two main paradigms have gained popularity: pixel-space~\citep{saharia2022photorealistic, DALLE2} and latent-space diffusion~\citep{rombach2022high, podell2023sdxl, baldridge2024imagen, betker2023improving}. Pixel-space models typically employ a cascaded architecture, where a base model generates low-resolution images, and subsequent models progressively upscale the generated images to higher resolutions. DALLE2~\citep{DALLE2} conditions the pixel-space diffusion on CLIP~\citep{radford2021learning} text embeddings, while Imagen~\citep{saharia2022photorealistic} uses T5~\citep{raffel2020exploring} embeddings. eDiff-I~\citep{ediffI} trains an ensemble of expert models, each specializing in a specific noise range, to enhance the generation quality.
Latent-space models, on the other hand, employ an autoencoder to compress images into a low-dimensional latent representation, upon which a diffusion model is trained. Stable diffusion and Stable Diffusion XL~\citep{podell2023sdxl} utilize U-Net based architectures for both autoencoders and diffusion models, with additional CLIP text conditioning. DALLE3~\citep{betker2023improving} trains the diffusion model using upsampled prompts from LLMs for generating images with long descriptive prompts. Stable Diffusion 3~\citep{esser2024scaling} adopts a Diffusion Transformer (DiT)~\citet{peebles2023scalable} based architecture to scale the latent space models up to $8B$ parameters.

\paragraph{Generation with Additional Control} In addition to purely text-based image generation, many other methods aim at adding more control signals. They can be easily classified as training-free methods \citep{meng2022sdedit,xue2023freestylenet,chen2023trainingfree,bansal2023universal}, or methods that require further training to existing text-to-image models \citep{huang2023composer,mou2024t2i,zhang2023adding,zhao2024uni,qin2023unicontrol,li2023gligen,ju2023humansd}. In general, training-based methods perform better than training-free methods. The most well-known is ControlNet \citep{zhang2023adding}, which adds an identical encoder branch and exclusively trains that branch.

\paragraph{Panorama and HDR Synthesis}
Panorama generation conditioned on text input has seen its emergence following the popularity of text-to-image synthesis using diffusion models. Most existing methods adopt a divide-and-conquer approach by first generating smaller image patches and then stitching them together \citep{chen2022text2light,zhang2023diffcollage,tang2023mvdiffusion,li2023panogen,wang2024customizing,zhang2024taming}. For HDR synthesis from LDR images, recent trends have been using deep neural networks with the aid of either a perceptual loss \citep{liu2020single,marcel2020single} or a GAN loss \citep{wang2022stylelight,wang2023glowgan}.

\paragraph{Finetuning} A substantial amount of work has been devoted to personalization and stylization for text-to-image generative models. Training-based personalization approaches include \citet{kumari2023multi,tewel2023key,yeh2023navigating,ruiz2022dreambooth,gal2022image,hu2021lora,xie2023difffit,han2023svdiff,voynov2023p+,arar2024palp,chen2023disenbooth,zhao2023dreamdistribution,marjit2024diffusekrona,shah2023ziplora}. And training-based stylization methods are explored in \citet{sohn2023styledrop,dong2022dreamartist,sinha2023text,zhao2023dreamdistribution,shah2023ziplora}. Our finetuning approach is more closely related to these training-based methods. Another branch of research focuses on fast finetuning or tuning-free techniques, which rely on pretrained image encoders or hypernetworks for personalization or stylization. The works in this area include, but are not limited to, \citet{ruiz2024hyperdreambooth,rout2024rb,arar2023domain,gal2023encoder,wei2023elite,zeng2024jedi,xiao2023fastcomposer,ma2023unified,shi2024instantbooth,jia2023taming,chen2024subject,li2024blip,valevski2023face0,yuan2023inserting,wang2024instantstyle,wang2024instantid}.

\section{Conclusion}

In this work, we presented Edify Image, a suite of image generation models producing high-fidelity images trained on a large dataset of image-text pairs. We proposed a novel multi-scale diffusion model, called \emph{Laplacian Diffusion Model}, in which different image frequency bands are decayed at different rates in the diffusion process. Additionally, we explored several intriguing capabilities that can be adapted from our base model, including ControlNets, $4K$ upsampling, finetuning, and $360^{\circ}$ panorama generation. 

\clearpage
\appendix
\section{Contributors and Acknowledgements}
\label{sec:contributors}

\subsection{Core contributors}

\textbf{Text-to-image model:} Yogesh Balaji, Qinsheng Zhang, Jiaming Song, Ming-Yu Liu

\textbf{Super-resolution:} Ting-Chun Wang, Siddharth Gururani, Seungjun Nah, Ming-Yu Liu

\textbf{ControlNets:} Ting-Chun Wang, Yu Zeng, Grace Lam, Ming-Yu Liu

\textbf{$360^{\circ}$ Panaroma Generation:} Ting-Chun Wang, J. P. Lewis, Seungjun Nah, Ming-Yu Liu

\textbf{Finetuning:} Jiaojiao Fan, Xiaohui Zeng, Yin Cui, Ming-Yu Liu

\textbf{Data Processing:} Jacob Huffman, Yunhao Ge, Siddharth Gururani, Fitsum Reda, Seungjun Nah, Yin Cui, Arun Mallya, Ming-Yu Liu

\subsection{Contributors} Yuval Atzmon, Maciej Bala, Tiffany Cai, Ronald Isaac, Pooya Jannaty, Tero Karras, Aaron Licata, Yen-Chen Lin, Qianli Ma, Ashlee Martino-Tarr, Doug Mendez, Chris Pruett, Fangyin Wei

\subsection{Acknowledgements}
We thank Timo Aila, Samuli Laine, Gal Chechik, Tsung-Yi Lin, Chen-Hsuan Lin and Zekun Hao for useful research discussions. We are grateful to Alessandro La Tona, Amol Fasale, Arslan Ali, Aryaman Gupta, Brett Hamilton, Devika Ghaisas, Gerardo Delgado Cabrera, Joel Pennington, Jason Paul, Jashojit Mukherjee, Jibin Varghese, Lyne Tchapmi, Mitesh Patel, Mohammad Harrim, Nathan Hayes-Roth, Raju Wagwani, Sydney Altobell, Thomas Volk and Vaibhav Ranglani for engineering and testing support.

Special thanks to Andrea Gagliano, Bill Bon, Si Moran and Grant Farhall of Getty Images for providing useful feedback on our image generators, and to Dade Orgeron, Steve Chappell, Lucas Brown and Alex Ambroziak of Shutterstock for providing feedback on our panaroma generations. We also thank the NVIDIA Creative team and Peter Pang for providing stylization finetuning data. Finally, we would like to thank Amanda Moran, Sivakumar Arayandi Thottakara, John Dickinson, Herb Woodruff, Dane Aconfora, Yazdan Aghaghiri, Yugi Guvvala, David Page and Andrew Morse for the computing infrastructure support.

\section{More Discussions on Laplacian Diffusion Models}
\label{app:formulation}
\subsection{Diffusing Signals at Different Resolution}

\begin{figure}[htbp]
    \centering
    \includegraphics[width=\textwidth]{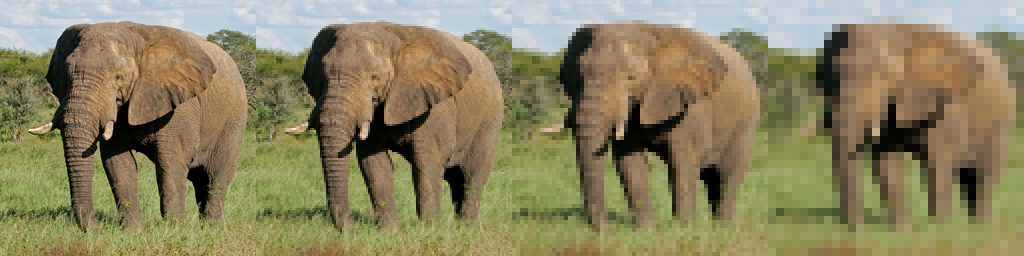}
    \includegraphics[width=\textwidth]{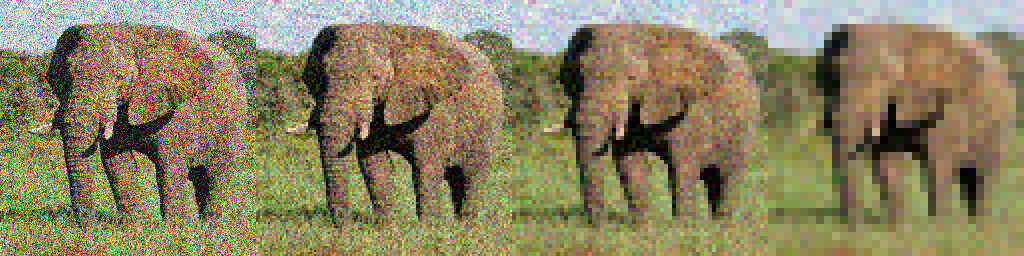}
    \caption{(Top) Noise-free images $\rvx$ at different resolutions, $256, 128, 64, 32$. (Bottom) Noisy images $\rvx + 0.02 \epsilon$ at different resolutions, $\epsilon$ has zero mean and identity matrix as covariance. From left to right in the bottom row, the images look less and less noisy, because the signal-to-noise ratio increases when we downsample the noisy images.}
    \label{fig:noisy-image-res}
\end{figure}

In addition to our work, other concurrent works~\citep{chen2023importance} also explore the diffusion effects at various resolutions and propose adjusting noise level sampling during training for different resolutions.
We reiterate this from the perspective of signal-to-noise ratio and derive noise scaling factors in the diffusion process.
First, let us consider the the average pooling (down) and nearest neighbor upsampling (up) operations. If we average pool the Gaussian noises, then each value in the downsampled tensor would have a lower variance. This is because in the case of $2\times 2$ average pooling,  assuming independent Gaussian random variables with $\gN(0, 1)$, we get
\begin{align}
    \frac{1}{4}(\gN(0, 1) + \gN(0, 1) + \gN(0, 1) + \gN(0, 1)) \sim \frac{1}{4} \gN(0, 4) \sim \frac{1}{2} \gN(0, 1).
\end{align}
The reduced variance comes from summing independent Gaussian variables.
From the definition of signal-to-noise ratio, we can see that signal-to-noise ratio will double when we downsample the noisy image. 
We illustrate the point in~\cref{fig:noisy-image-res}. 

\subsection{Derivations}

\paragraph{Forward Diffusion Process}

For simplicity, we consider splitting the current sample $\rvx(0)$ into two subspaces, corresponding to the effective resolution of $r$ and $R$. As the signal to noise ratio is different for different resolutions, we use the variable $t$ to describe the ``time'' of the process. We use the notations $L$ and $H$ to represent low-frequency and high-frequency subspaces, respectively. 
\begin{align}
    \rvx^{(L)} &= \text{up}(\text{down}(\rvx(0), R/r)) \\
    \rvx^{(H)} &= \rvx - \rvx^{(L)}
\end{align}
The forward process also treat the two differently:
\begin{itemize}
    \item $\rvx^{(L)}(t) = \rvx^{(L)}(0) + \sigma(t) \epsilon^{(L)}$, where $\epsilon \sim \gN(0, I)$.
    \item $\rvx^{(R)}(t) = \alpha(t) \rvx^{(r)}(0) + \sigma(t) \epsilon^{(H)}$.
\end{itemize}
Specifically, $\alpha(t)$ is a function that vanishes after a certain time (denoted as $t^{(L)}$). We use EDM formulation to derive the drift and diffusion coefficients. 
\begin{align}
    \diff \rvx = f(t) \rvx + g(t) \diff \omega
\end{align}

leads to $\rvx(t) = s(t) \rvx(0) + s(t)^2 u(t)^2 \epsilon$, where:
\begin{align}
    s(t) = \exp\left(\int_0^t f(\xi) d(\xi) \right) \quad \text{and} \quad u(t) = \sqrt{\int_0^t \frac{g(\xi)^2}{s(\xi)^2} \diff \xi}.
\end{align}

For the $L$ subspace:
\begin{align*}
    s(t) & = 1, \\
    f(t) & = 0, \\
    u(t) & = \sigma(t), \\
    g(t) & = \sqrt{\frac{\diff \sigma(t)^2}{\diff t}},
\end{align*}
and for the $H$ subspace:
\begin{align*}
    s(t) & = \alpha(t), \\
    f(t) & = \frac{\diff \log \alpha(t)}{\diff t}, \\
    u(t) & = \frac{\sigma(t)}{\alpha(t)}, \\
    g(t) & = \sqrt{\frac{\diff (\sigma(t)^2 / \alpha(t)^2)}{\diff t}} \alpha(t)
\end{align*}

\paragraph{Backward Diffusion Process}
The backward diffusion process can be derived as:
\begin{align}
    \diff \rvx = \left[ f(t) \rvx - \frac{1}{2} g(t)^2 \nabla_{\rvx} \log p_t(\rvx) \right] \diff t,
\end{align}
or equivalently as:
\begin{align}
    \diff \rvx = \left[\dot{s}(t) \rvx / s(t) - s(t)^2 \dot{u}(t) u(t) \nabla_{\rvx} \log p_t(\rvx) \right] \diff t,
\end{align}
so we can derive the ODE as follows. For the $L$ subspace:
\begin{align}
    \diff \rvx = \left[ - \frac{1}{2} \frac{\diff \sigma(t)^2}{\diff t} \nabla_{\rvx} \log p_t(\rvx) \right] \diff t,
\end{align}
and for the $H$ subspace:
\begin{align}
    \diff \rvx = \left[\frac{\diff \log \alpha(t)}{\diff t} \rvx - \frac{\sigma(t) (\dot{\sigma}(t) \alpha(t) - \dot{\alpha}(t) \sigma(t))}{\alpha(t)} \nabla_{\rvx} \log p_t(\rvx) \right] \diff t.
\end{align}
\section{Finetuning Training Data}

The training images used for finetuning experiments in \cref{sec:finetuning} are provided below.





\begin{figure}[ht]
    \centering
    \includegraphics[width=\textwidth]{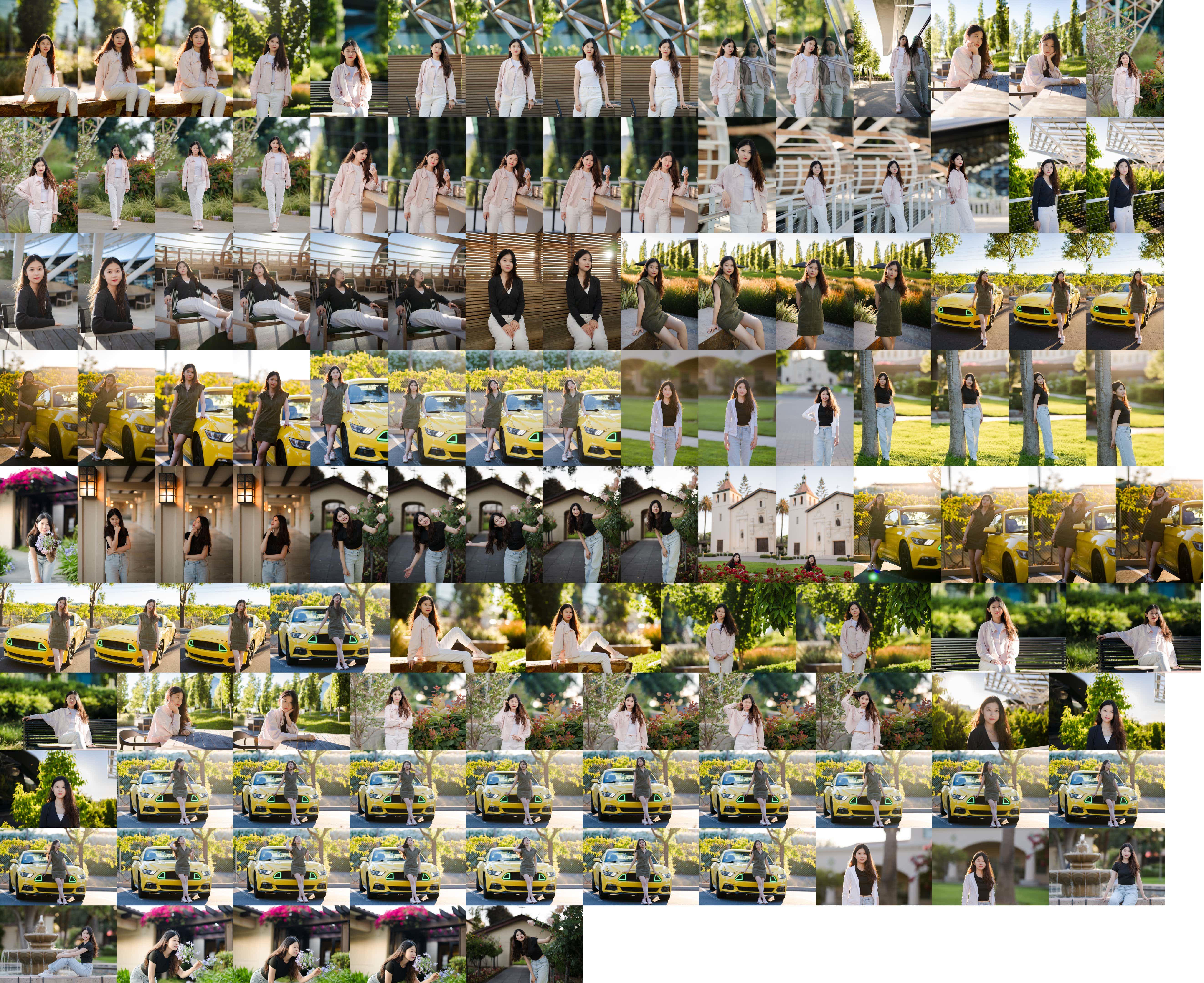}
    \caption{Training images used for single-subject personalization.}
    \label{fig:fangyin_train}
\end{figure}

\begin{figure}[ht]
    \centering
    \includegraphics[width=\textwidth]{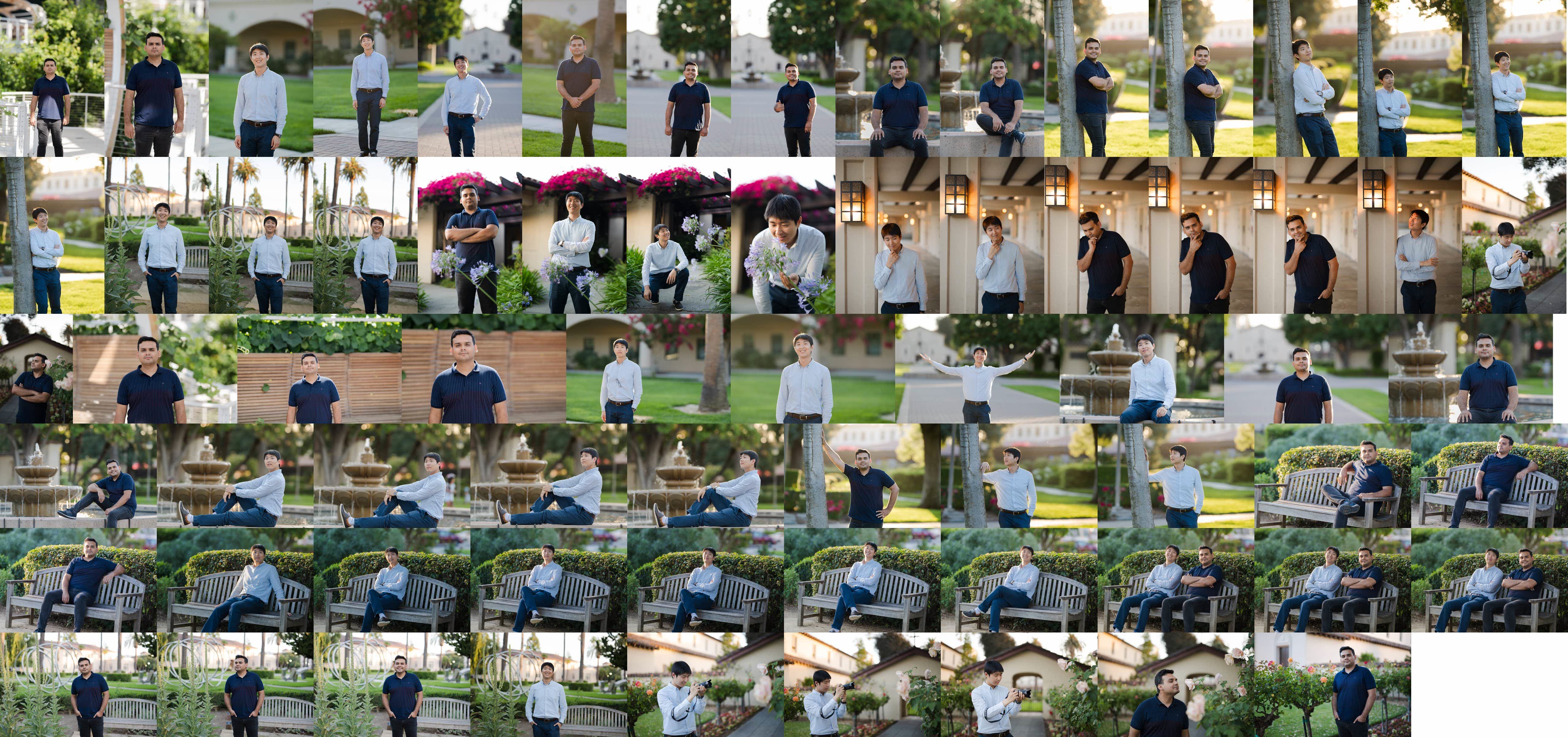}
    \caption{Training images used for multi-subject personalization.}
    \label{fig:ss_train}
\end{figure}

\begin{figure}[ht]
    \centering
    \includegraphics[width=\textwidth]{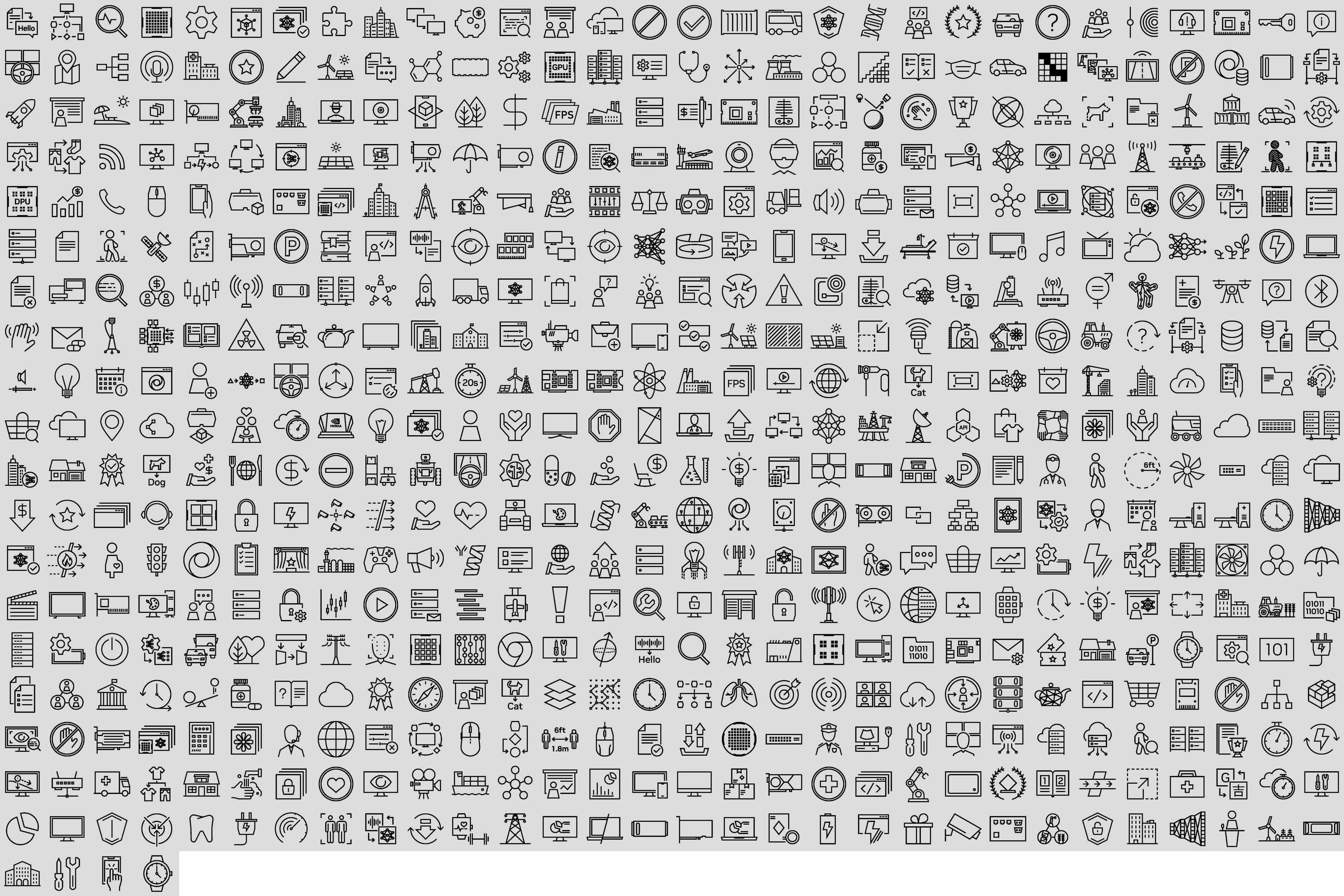}
    \caption{Training images used for single-subject stylization.}
    \label{fig:icon_train}
\end{figure}

\begin{figure}[ht]
    \centering
    \includegraphics[width=\textwidth]{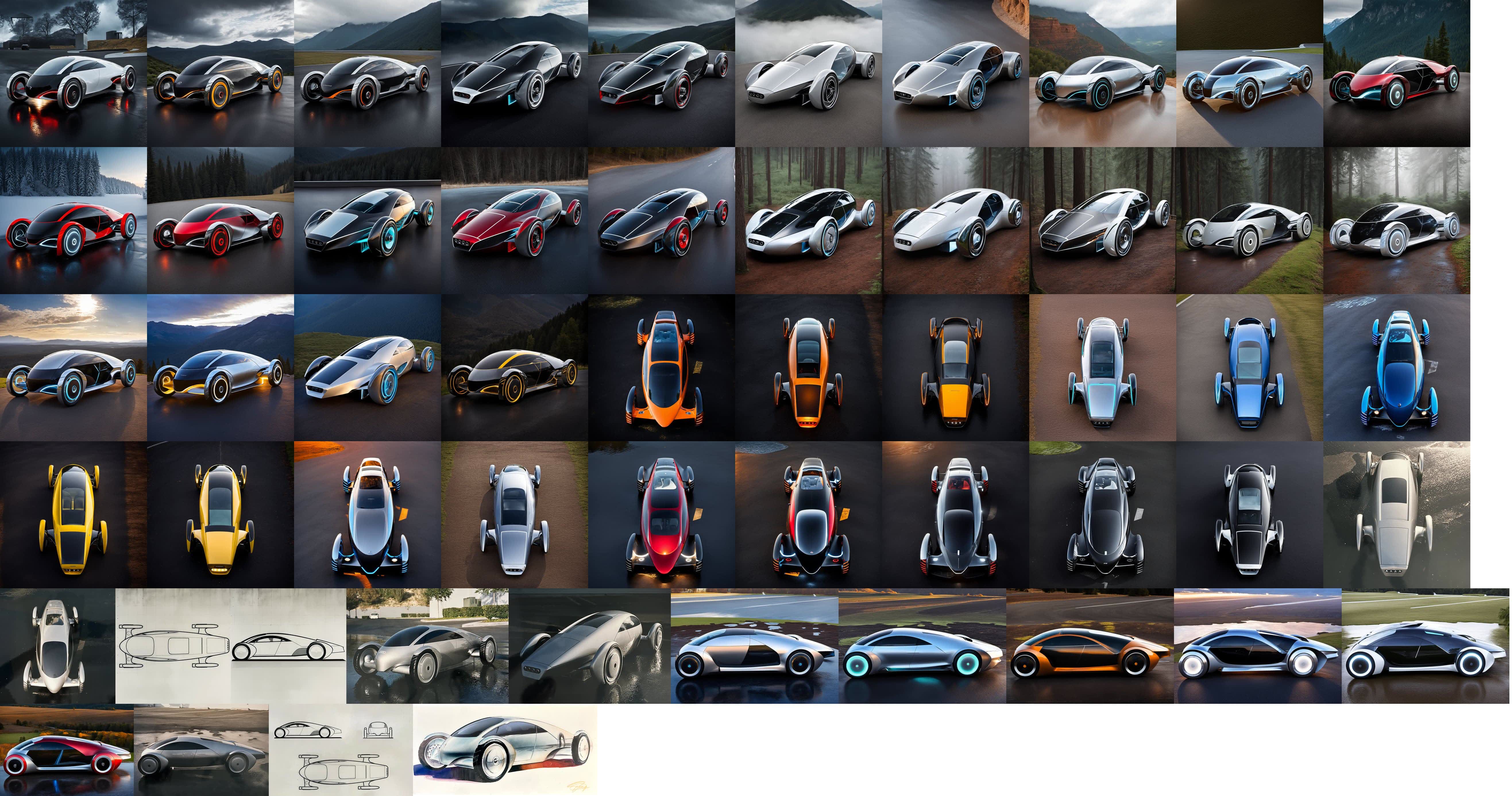}
    \caption{Training images used for multi-subject stylization.}
    \label{fig:car_train}
\end{figure}

\clearpage
\setcitestyle{numbers}
\bibliographystyle{plainnat}
\bibliography{main}

\end{document}